\documentclass[acmtog]{acmart}

 \setcopyright{none} \renewcommand\footnotetextcopyrightpermission[1]{} 

\AtBeginDocument{%
  \providecommand\BibTeX{{%
    \normalfont B\kern-0.5em{\scshape i\kern-0.25em b}\kern-0.8em\TeX}}}

\setcopyright{acmlicensed}
\copyrightyear{2024}
\acmYear{2024}
\acmDOI{XXXXXXX.XXXXXXX}
\usepackage{subfigure}
\usepackage{color,soul}
\usepackage{tabularx}
\usepackage{multirow}
\usepackage{graphicx}
\usepackage{subcaption}
\usepackage{float}
\usepackage{placeins}

\usepackage{tikz}
\usetikzlibrary{fit}
\usetikzlibrary{shapes,arrows}

\newcommand \mbl[1]{%
    \tikz[overlay,remember picture]
        \node (marker-#1-a) at (0,1.2ex) {};%
}
\newcommand \mbr[2]{%
    \tikz[overlay,remember picture]
        \node (marker-#1-b) at (0,0.3ex) {};%
    \tikz[color = #2, overlay,remember picture,thick,inner sep=3pt]
        \node[draw,rounded rectangle,fit=(marker-#1-a.center) (marker-#1-b.center)] {};
}

\begin{document}

\newpage

\title{Semantic Aware Diffusion Inverse Tone Mapping}

\author{Abhishek Goswami}
\affiliation{%
  \institution{University of Warwick}
  % \streetaddress{}
  % \city{}
  \country{UK}}
\email{abhishek.goswami@warwick.ac.uk}

\author{Aru Ranjan Singh}
\affiliation{%
  \institution{University of Warwick}
  % \streetaddress{}
  % \city{}
  \country{UK}}
\email{aru.singh@warwick.ac.uk}

\author{Francesco Banterle}
\affiliation{%
  \institution{ISTI-CNR}
  % \streetaddress{}
  % \city{}
  \country{Italy}}
\email{frabante@gmail.com}

\author{Kurt Debattista}
\affiliation{%
  \institution{University of Warwick}
  % \streetaddress{}
  % \city{}
  \country{UK}}
\email{k.debattista@warwick.ac.uk}

\author{Thomas Bashford-Rogers}
\affiliation{%
  \institution{University of Warwick}
  % \streetaddress{}
  % \city{}
  \country{UK}}
\email{thomas.bashford-rogers@warwick.ac.uk}

\renewcommand{\shortauthors}{DITMO, Goswami et al.}

\sloppy

\begin{abstract}
    The range of real-world scene luminance is larger than the capture capability of many digital camera sensors which leads to details being lost in captured images, most typically in bright regions. Inverse tone mapping attempts to boost these captured Standard Dynamic Range (SDR) images back to High Dynamic Range (HDR) by creating a mapping that linearizes the well exposed values from the SDR image, and provides a luminance boost to the clipped content; however, in most cases, the details in the clipped regions cannot be recovered or estimated. In this paper, we present a novel inverse tone mapping approach for mapping SDR images to HDR that generates lost details in clipped regions through a semantic-aware diffusion based inpainting approach. Our method proposes two major contributions -- first, we propose to use a semantic graph to guide SDR diffusion based inpainting in masked regions in a saturated image. Second, drawing inspiration from traditional HDR imaging and bracketing methods, we propose a principled formulation to lift the SDR inpainted regions to HDR that is compatible with generative inpainting methods. Results show that our method demonstrates superior performance across different datasets on objective metrics, and subjective experiments show that the proposed method matches (and in most cases outperforms) state-of-art inverse tone mapping operators in terms of objective metrics and outperforms them for visual fidelity.
\end{abstract}




\begin{teaserfigure}
  \begin{center}
\setlength{\tabcolsep}{1pt}
  \begin{tabular}{cccc}
  \includegraphics[width=0.24\textwidth]{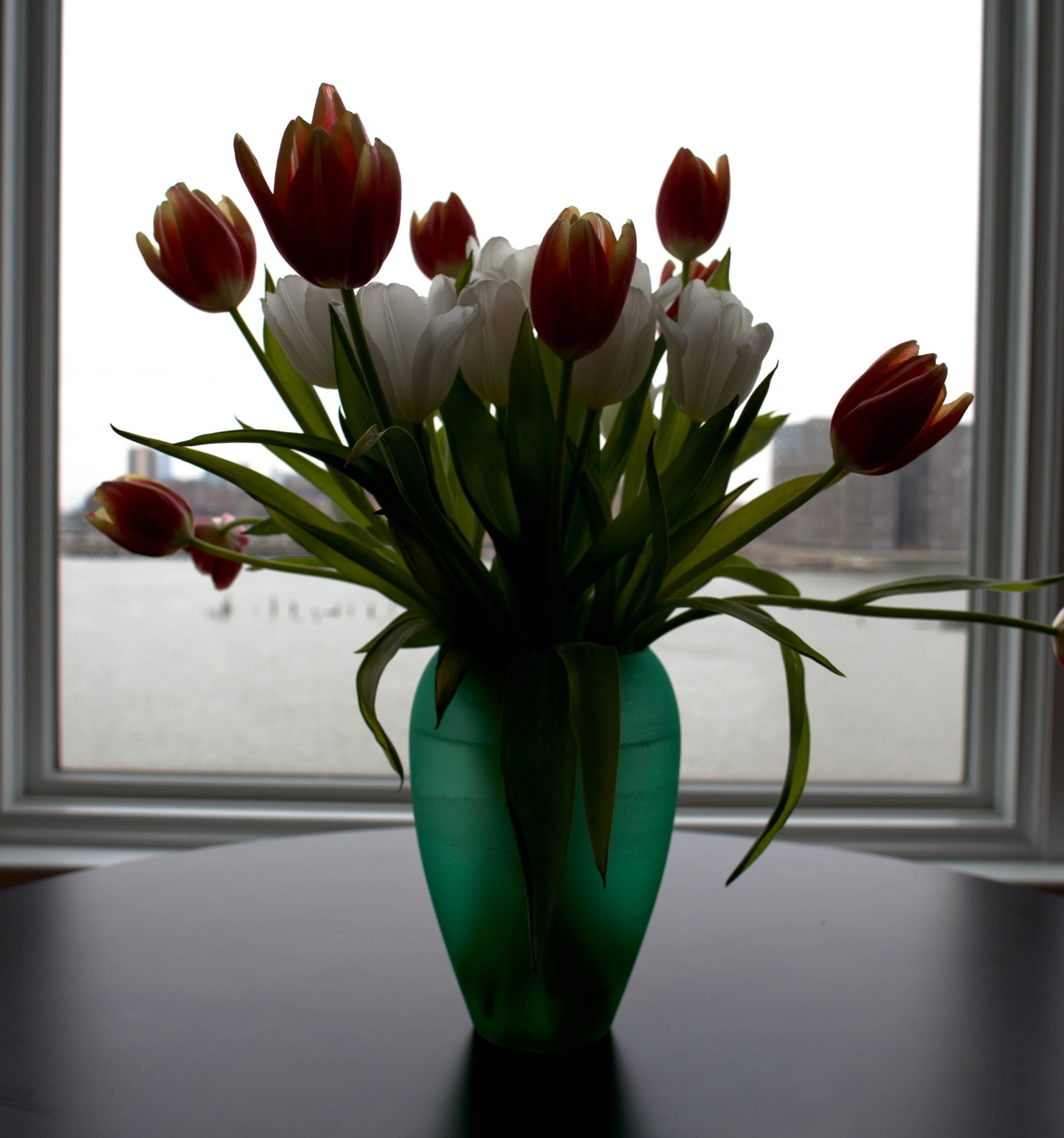} & 
  \includegraphics[width=0.24\textwidth]{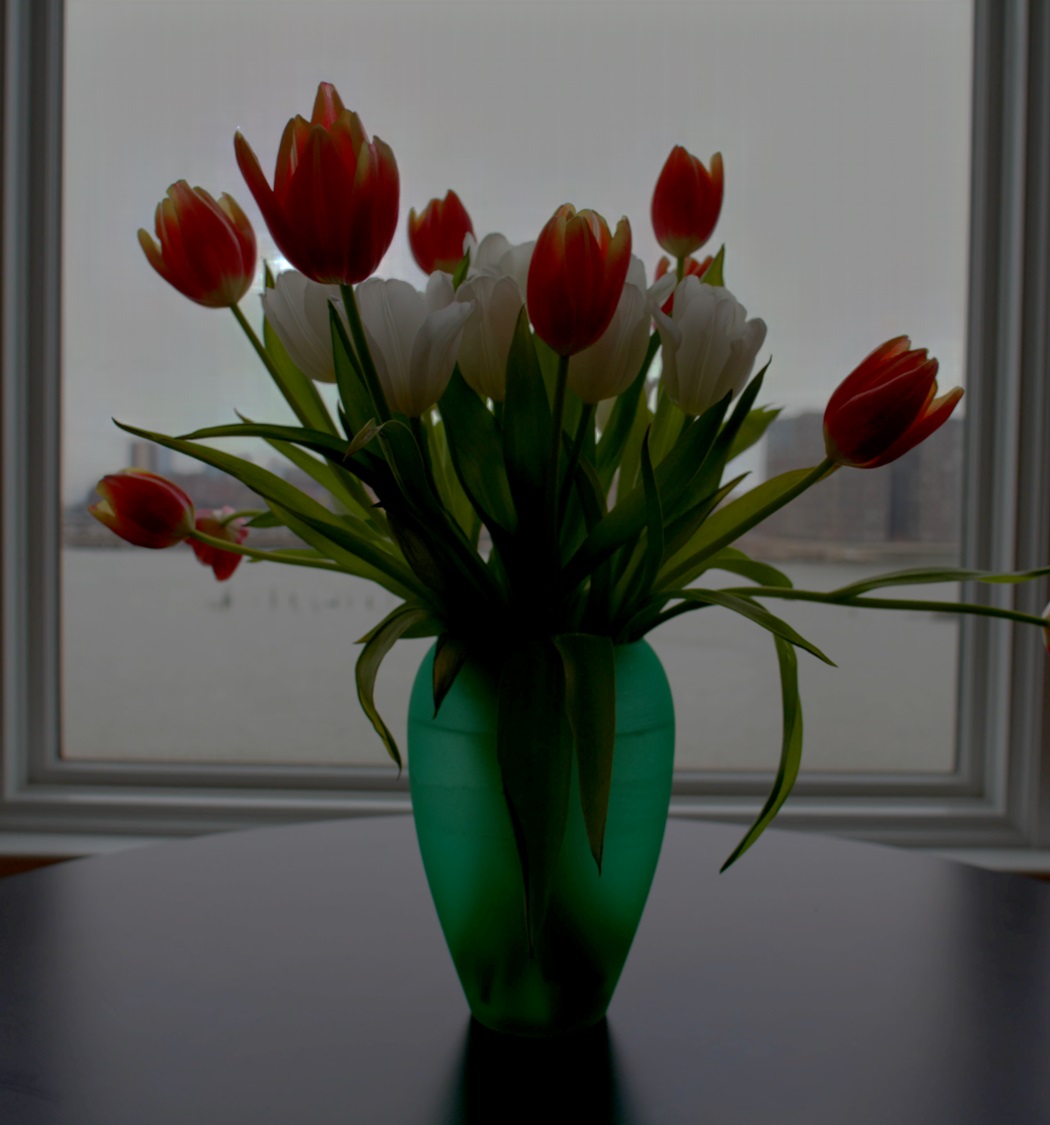} & 
  \includegraphics[width=0.24\textwidth]{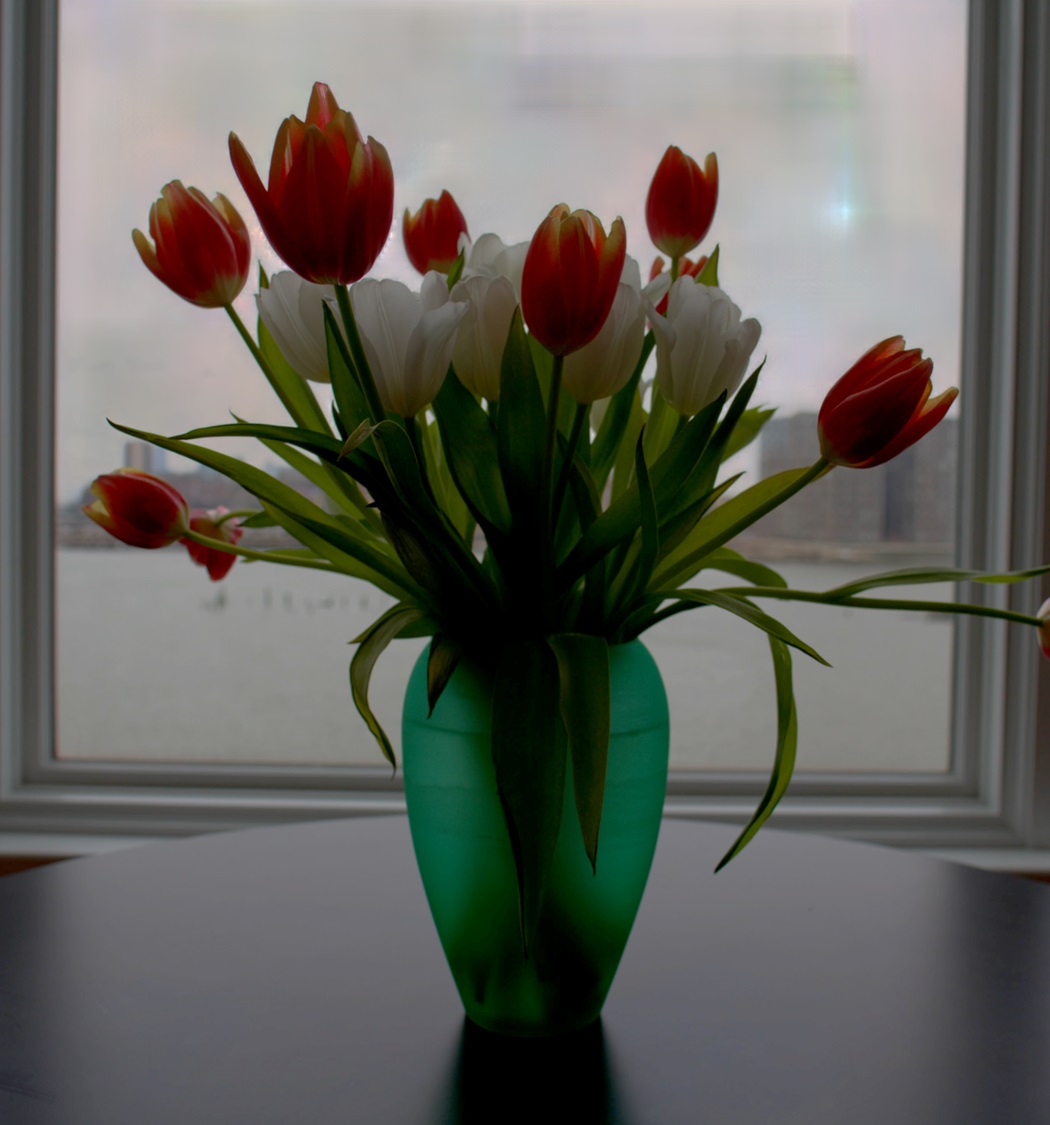} & 
  \includegraphics[width=0.24\textwidth]{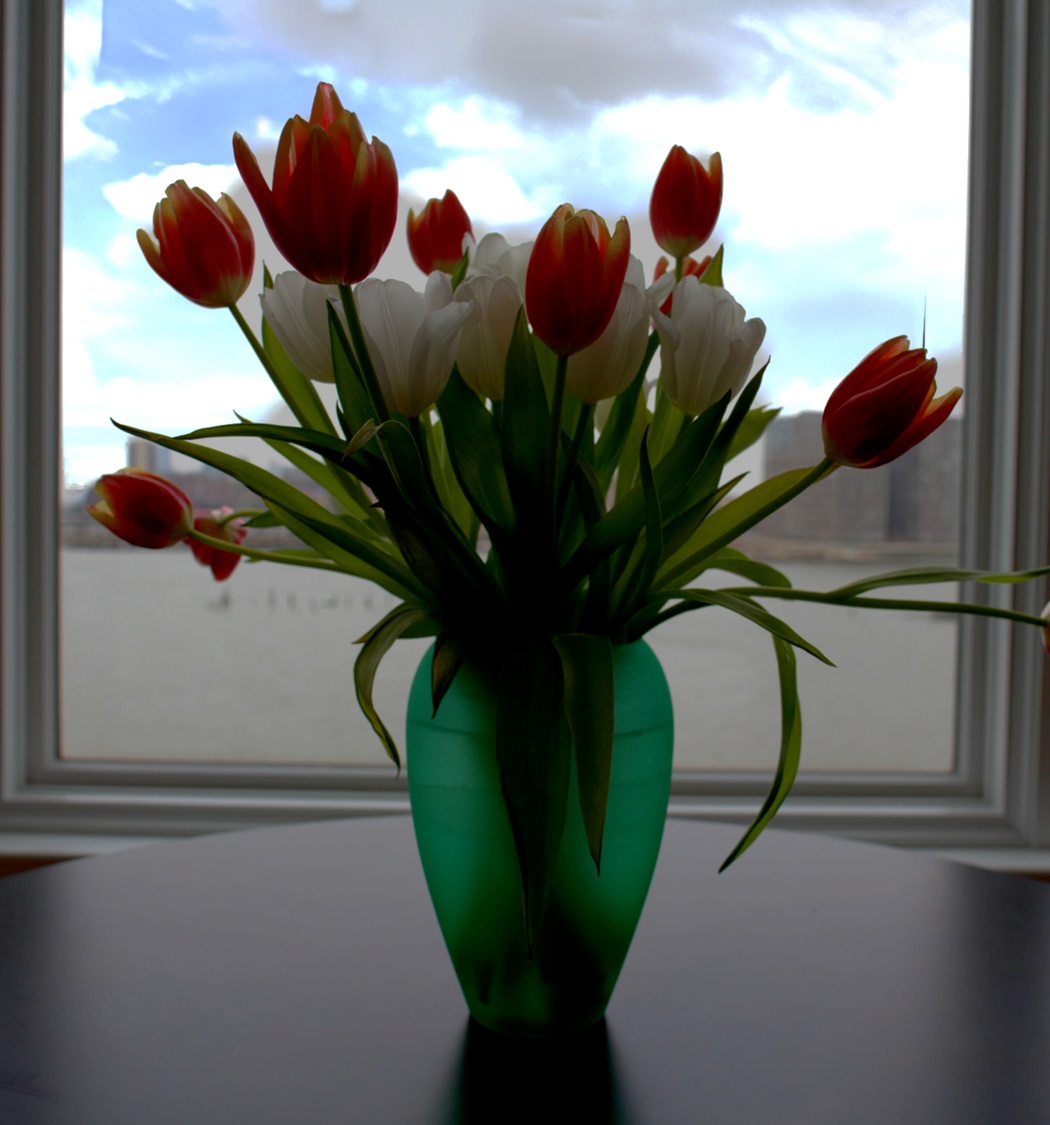} \\
  (a) Input & (b) \cite{Eilertsen+17} & (c) \cite{Santos+20}  & (d) Our DITMO
  \end{tabular}
  \end{center}
  \caption{An example of expansion using our method when applied to a SDR image with large over-exposed areas: (a) The input; (b) The result by Eilertsen et al. \cite{Eilertsen+17} at -2-stop; (c) The result by Santos et al. \cite{Santos+20} at -2-stop; (d) The results using our method. Note that these methods suffer to reconstruct content when over-exposed areas are large; e.g., more than 21\% over-exposed pixels in (a). On the contrary, our method, DITMO, inpaints a plausible sky and clouds.}
  \Description{.}
  \label{fig:teaser}
\end{teaserfigure}

\maketitle

\section{Introduction}

HDR technology has become ubiquitous in our daily life thanks to modern smartphones that can capture multiple exposures and merge them into HDR images or videos \cite{Hasinoff+16}, and display such content in HDR \cite{Seetzen+04}. However, most of the content captured currently is still SDR, and a substantial amount of legacy content dating back decades is also SDR. Many of these images are missing details due to quantization and clipping during the acquisition and storage process.

Inverse tone mapping seeks to reverse this process and generate HDR content from SDR images. Most of these methods attempt to linearize and expand the range of the SDR image by adjusting pixel values in a non-linear manner. This often works well for well exposed pixels, but clipped values at the extremes cannot be recovered, and instead these methods typically shift the luminance in these clipped regions without reconstructing any details. While there are a couple of exceptions, for example, \cite{Santos+20}, most of these methods still do not produce HDR images with plausible real world details in important, and frequently large, clipped regions of the image.

We aim to solve this problem by linearization of the pixels in well exposed regions, and reconstructing these lost details in the clipped regions. This raises two challenges: how can these details be added to the clipped regions and how can we ensure that the process results in a plausible HDR image?

The problem of adding details to SDR images, which is commonly termed inpainting, has been tackled by methods such as Diffusion Models \cite{stableDiffusion}. These use a Deep Neural Network to progressively denoise a random input into an image and are frequently conditioned via text prompts. These methods are trained on a large database of SDR images and naturally produce SDR images. We propose a method, DITMO, to lift these SDR inpainted regions to HDR inspired by the process of merging bracketed exposures. Specifically, we predict the exposure that the inpainted context \emph{would} have been captured at if a real camera was used, and use this to generate the missing exposures of an HDR image. Our method is also agnostic to the specific inpainting model, and as such is compatible with all current and \emph{future} inpainting approaches and we show results using two common inpainting methods.

We also need to ensure consistency between the image and inpainted regions, for example, inpainted sky ideally should be reflected in inpainted water, where both are potentially inpainted at different exposures. This leads to a natural ordering during the inpainting process, and we propose using semantic segmentation and a novel ordered semantic graph to ensure this consistency is maintained.

To summarize, our main contributions are:

\begin{itemize}
    \item An inverse tone mapping method that reconstructs clipped regions and produces HDR images without resolution limitations;
    \item A semantic aware segmentation and ordered semantic graph to produce semantically consistent inpainted results;
    \item Results that are objectively competitive with or improve on the state-of-the-art, and a subjective experiment (N=20) that shows our approach substantially improves over similar methods.
\end{itemize}

\section{Related Work}

This section covers related work in areas close to ours: inverse tone mapping, semantic segmentation for HDR imaging and diffusion models.

Inverse/Reverse Tone Mapping Operators (ITMOs) \cite{Meylan+06, Banterle+06} or Single-Image-HDR Reconstruction (SIHR) methods \cite{Hanji+22} convert images and videos from the SDR domain into the HDR domain. Typically, SDR values are 8-bit images per color channel with a Camera Response Function (CRF) applied to these values.
Researchers have proposed a number of approaches for ITMOs which can be broadly categorized as:
\begin{itemize}
    \item \textbf{Image-Processing-based methods}: that use linear and non-linear filters, Poisson-image-editing, expanding function, etc. to expand the dynamic range in over-exposed areas of the content.
    \item \textbf{Deep-learning-based methods}: that exploit convolutional neural networks (CNNs) to extract features from the content and expand missing details.
\end{itemize}

\subsection{Image-Processing-based Inverse Tone Mapping Operators}
\label{SEC:CLASSIC}
The first ITMOs expanded the dynamic range of over-exposed areas of content using straightforward linear\cite{Meylan+06,Akyuz+07} and non-linear functions \cite{Masia+09} respectively for well-exposed and over-exposed content with only mild compression. These methods can be extended to handle style aesthetics \cite{Bist+17}.
To reduce artifacts in compressed areas, the use of a spatially varying map for guiding the expansion was introduced 
\cite{Banterle+06,Rempel+07,Kovaleski+14}.
Most of these methods may have issues or limitations for certain images, therefore, \cite{Wang+07} introduced a solution based on Photoshop-like tools (e.g., the healing tool) to manually hallucinate SDR images. In another work, \cite{Didyk+08} introduced an interface based on machine-learning that semi-automatically classifies a frame from a video as diffuse, specular, and light-source surfaces. Once surfaces are classified, they are properly expanded.

\begin{figure*}[tp]
    \includegraphics[width=.85\textwidth]{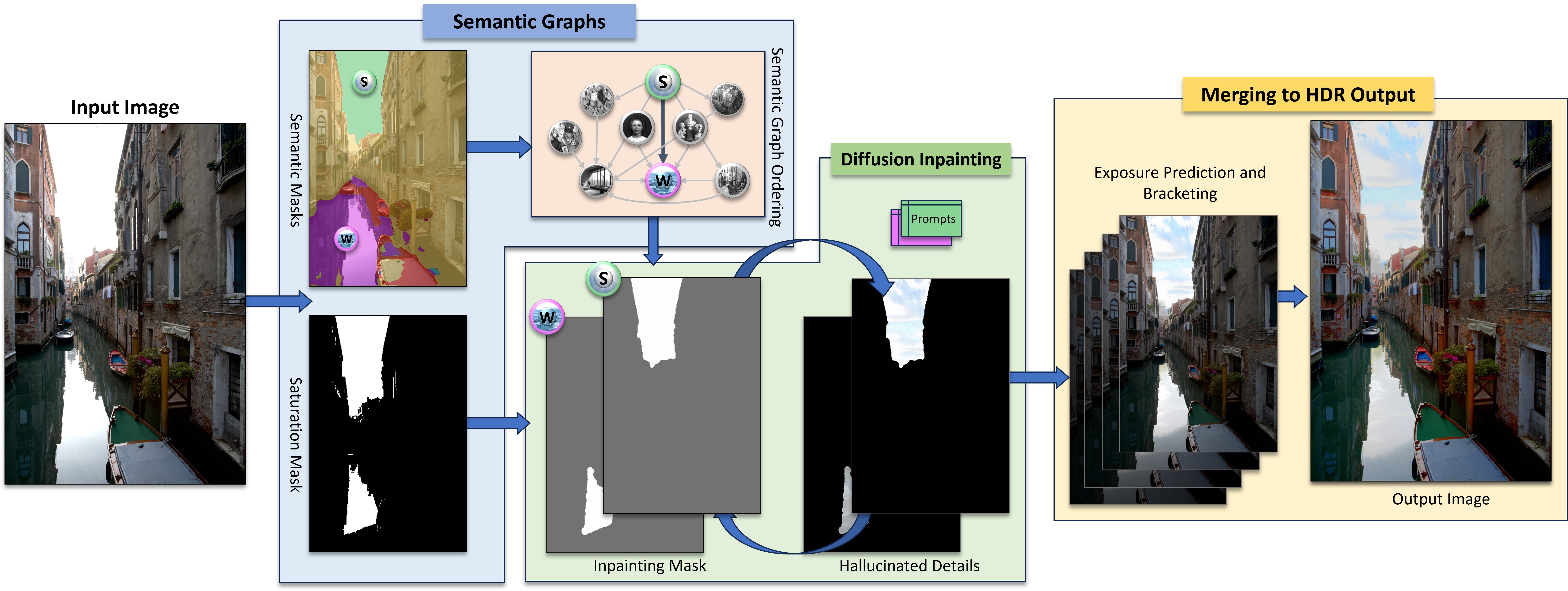}
    \caption{\textit{An overview of DITMO.} 1) \textit{Semantic Graph} computes an ordered directed graph based on semantic instances of the input image (S - Sky, W - Water). The saturation mask, semantic masks and the graph are used to compute the inpainting mask for the 2) \textit{Diffusion inpainting module}. A set of semantic aware prompts are used to hallucinate details in the clipped regions. Predicted exposure for the hallucinated details is used iteratively to produce well-exposed brackets for each semantic instance. The final module 3) \textit{Merging to HDR} fuses the brackets to generate an HDR image.}
    \label{fig:overview}
\end{figure*}

\subsection{Deep-Learning-based Inverse Tone Mapping Operators}
Although previous approaches, Section~\ref{SEC:CLASSIC}, can manage to match the overall luminance levels in over-exposed areas, they struggle in the reconstruction of the missing content; i.e., textures and colors. Only \cite{Wang+07} manages such reconstruction at the cost of extensive manual intervention by the user.

The introduction of deep-learning-based methods \cite{Eilertsen+17,Endo+17} has made the reconstruction process of textures and colors automatic improving the overall fidelity. They introduced two main approaches for deep learning inverse tone mapping: \cite{Eilertsen+17} directly maps from SDR to HDR using a U-Net \cite{Ronneberger+15}, and \cite{Endo+17} creates different SDR images at different exposure times mimicking the capturing of HDR images \cite{Debevec+97} using two U-Nets (over-exposed and under-exposed branches) with a 3D decoder. Then, these SDR images are merged using classic methods \cite{Debevec+97}.

\cite{Marnerides+18} introduced a CNN that extracts features at different scales to directly the dynamic range of SDR images into HDR images. Similarly to \cite{Endo+17},  \cite{Lee+18} used a GAN scheme to estimate under-exposed and over-exposed images starting from a single SDR image; these images are merged into an HDR image. \cite{Jo+22} extended this work to handle arbitrary exposure values.

\cite{Eilertsen+17} work was extended by \cite{Yu+21} by adding attention modules and specializing the method for 360 panoramic images for high quality image-based lighting results. Another approach to the problem was proposed by  \cite{Liu+20}. In their method, they invert each step of the camera pipeline using a network. \cite{Santos+20} proposed a U-Net based method in which the over-exposed regions of the input image are used to modulate, as a mask, the results of weights at different scales driving attention of the computation in such areas. This was able to hallucinate some detail in clipped regions, although this depends on the details available in the input image.

Other recent work has focused on under-exposed areas \cite{Zou+23}, and the use of SDR datasets for expanding the dynamic range \cite{Wang+23, Banterle+24}. Diffusion models have been used in a different HDR context by \cite{Chen+22}. In their work, they proposed a framework for generating 360 HDR panoramic images using prompts and a diffusion model. While sharing the concept of using diffusion for HDR imaging, they generate entirely synthetic HDR images rather than using existing information as in inverse tone mapping. Diffusion models have also been applied to inverse tone mapping by \cite{dalal2023single} who used a UNet along with an autoencoder to attempt to boost details in saturated regions. This is different from our work which seeks to synthesize new details in these regions.

\subsection{Semantic Awareness and Segmentation for HDR}

The use of semantic awareness and deep learning based semantic segmentation has been a broadly investigated topic. Deep learning based semantic segmentation models~\cite{fastfcn+19, yurtkulu+19, xie+21, kirillov+23} rely on detailed annotations to produce plausible masks. The popular FastFCN method ~\cite{fastfcn+19} was trained on 150 fine semantic labels of the ADE20k dataset\cite{zhou+17} and creates fine semantic classes which can be used for application specific enhancement tasks. Despite good annotations, achieving pixel precision for masks in natural scenes is a challenge and hence alpha matting~\cite{Germer+20} or soft segmentation has been vital for image enhancement tasks. 

Semantic segmentation has been used for tone mapping HDR content to SDR. \cite{goswami+20, goswami+22} use semantic features explicitly for HDR tone mapping. They show that each semantic instance has unique mapping target and it is dependent on the spatial arrangement of semantic regions in the image. They further show that a course semantic segmentation of an image can be represented as a graph to extract local neighborhood information better to produce aesthetically pleasing results. 

\subsection{Image Inpainting}

Image inpainting aims to synthetically restore or synthesize portions of an image seamlessly. Generative models such as Generative Adversarial Networks (GANs) have been widely used for the task \cite{ImageInpainting}.
Recent developments in Diffusion model-based image generation methods have shown the ability to generate realistic synthetic information  \cite{dhariwal2021diffusion}. Additionally, prompts from large language models can further guide such conditional models to generate plausible natural outputs \cite{Imagen}, and approaches such as Stable Diffusion Inpainting \cite{stableDiffusion} provide a latent text-to-image diffusion model capable of generating photo-realistic images given any text input and an inpainting mask.

\section{Method}

This section provides details about the semantic segmentation and ordering that drives our inpainting method, and then describes the process that combines the outputs of these steps into an HDR image. 

\subsection{Overview}

An overview of our method, DITMO, is shown in Figure \ref{fig:overview}.
DITMO takes as input an SDR image, possibly with large overexposed and clipped regions.  To reconstruct an HDR image, we first compute the content we want to reconstruct. To do this, we assign each pixel to a semantic class, a labeling associated with the pixel which assigns what class (i.e. grass, water, sky etc) the pixel represents. Then, we mask out the overexposed or clipped regions which require content to be inpainted. These steps are described in Section \ref{sec:semsegment}.

Inpainting follows an ordering specified by an ordered semantic graph, see Section \ref{sec:graph}. Each inpainted region is assigned to an exposure of the resultant HDR image (Section \ref{sec:exposureestimation}), and a series of exposures are generated which are merged into an HDR image. 

\subsection{Semantic Segmentation}
\label{sec:semsegment}

\begin{figure}
    \centering
    \frame{\includegraphics[width=.325\linewidth]{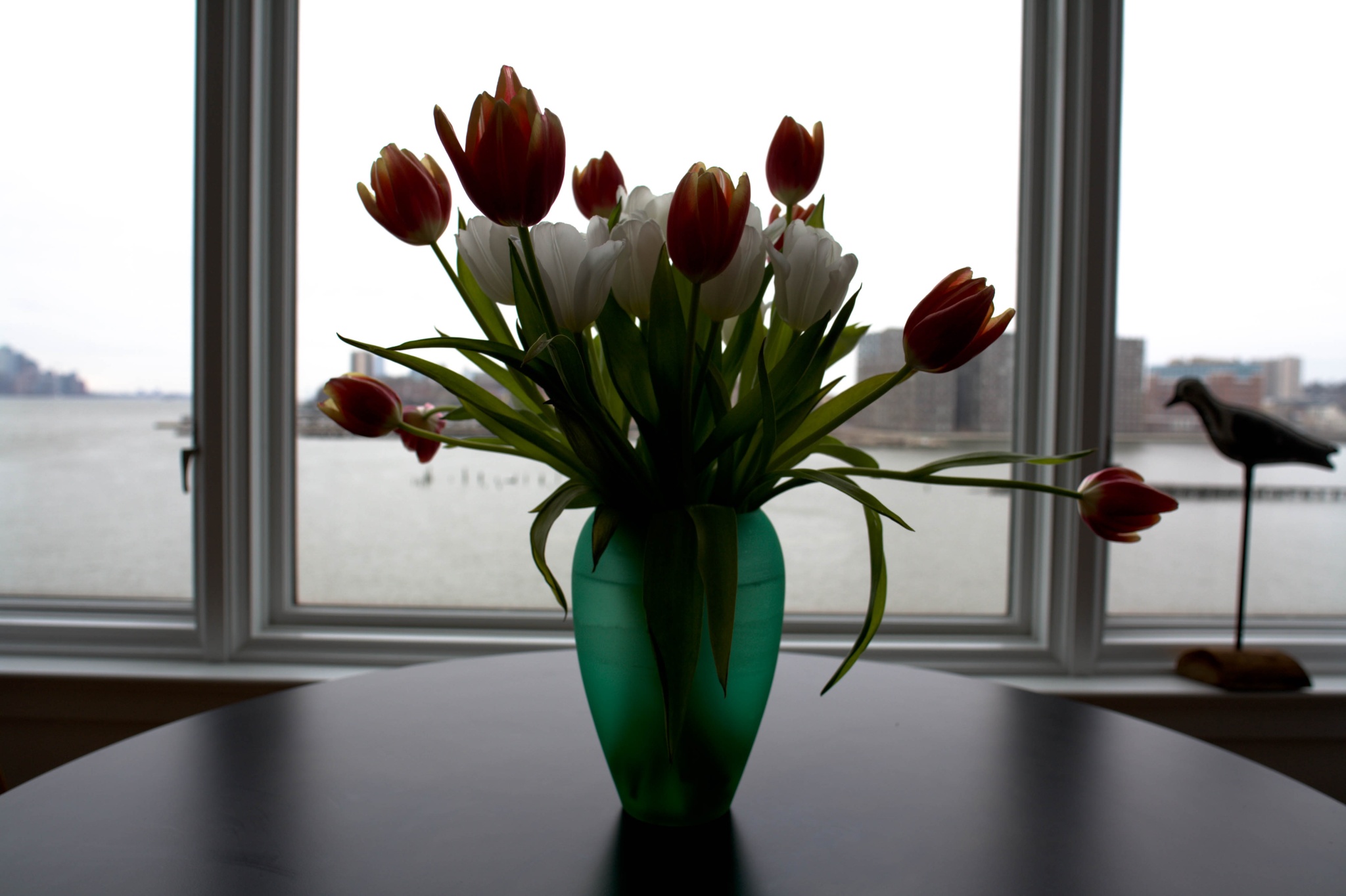}}
    \frame{\includegraphics[width=.325\linewidth]{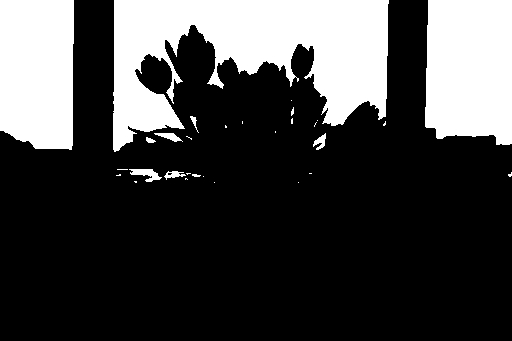}}
    \frame{\includegraphics[width=.325\linewidth]{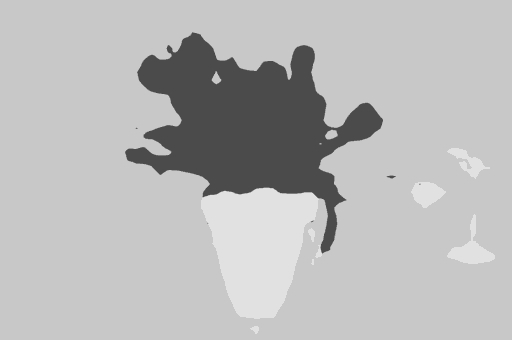}} \\
    \frame{\includegraphics[width=.325\linewidth]{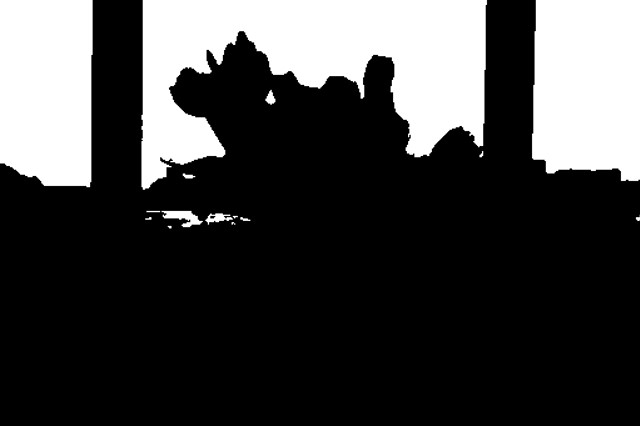}}
    \frame{\includegraphics[width=.325\linewidth]{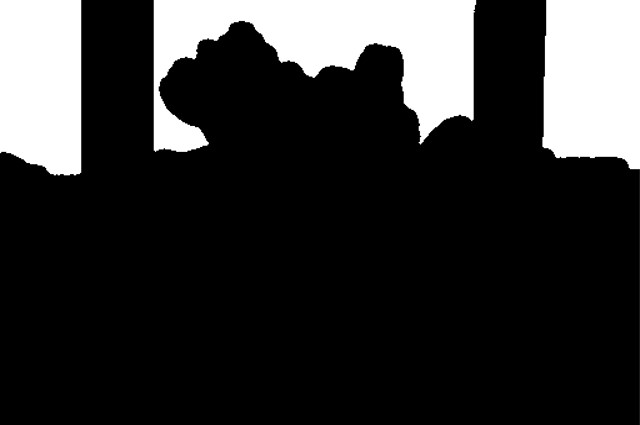}}
    \caption{\textit{Clockwise from top left:} Saturation mask denoting clipped regions, semantic mask denoting indoor scene with view through a window, intermediate mask prior to morphological operations and $M_p$ used for inpainting.}
    \label{fig:mask_refinement}
\end{figure}

We start by generating a `saturation mask' $M_{sat}$ for our SDR image and a set of 9 coarse `semantic masks', $M_{seg_i} \forall i \in [0,8]$. The semantic classes used in this work are -- \textit{sky, ground, vegetation, water, human-subject, non-human subject, cityscape, indoor, others}; however, more can be added if the semantic segmentation algorithm can reliably produce more classes. For each semantic class, we compute a thresholded mask, $M$, determined by the intersection of semantic and saturation masks.
\begin{equation}
    M = M_{sat} \cap M_{seg}
\end{equation}
We observed that the semantic masks are often not pixel precise and may overlap on boundary pixels of neighboring semantic regions. Hence, we apply alpha matting with 2 pixel radius on a lower resolution to improve the pixel precision of our intermediate mask. As a last step, the masks are further processed with morphological erosion and dilation operators to preserve pixel information within semantic regions. Providing information at the edge of the semantic masks helps during inpainting to maintain consistency in the information filled. Let $E$ be a Euclidean space where $M$ is a binary image in $E$. Eq.~\ref{inpaint_mask} shows the use of morphological kernel in the shape of a disk to perform a guided opening, \textit{erosion} followed by \textit{dilation} to obtain the final inpainting mask, $M_p$ with border information to preserve during inpainting.
\begin{equation}
\label{inpaint_mask}
    M_{p} = (M \ominus D) \oplus\hat{D}\text{,}
\end{equation}
\noindent where $\ominus$ denotes morphological erosion for binary mask $M$ by kernel $D$, defined by
\begin{equation}
\label{erosion}
    M \ominus D = \{ z\in E \mid D_z \subseteq M \} \text{,}
\end{equation}
\noindent where $D_z$ is the translation of the kernel by vector $z$, $B_z = \{ b+z \mid b \in B\}, \forall z \in E$. Similarly, $\oplus$ denotes morphological dilation for the eroded mask, $M_e$ by another kernel $\hat{D}$, defined by
\begin{equation}
\label{dilation}
    M_e \oplus \hat{D} = \bigcup_{b \in B} M_{e_b}
\end{equation}
A pair of variable parameters $(\alpha, \beta)$ are used to moderate the morphological operations. $\alpha$ denotes the radius for $D$ whereas $\beta$ denotes the same for $\hat{D}$. The variation of the parameters defines how much information for the mask border influences the inpainting. Figure~\ref{fig:mask_refinement} shows the input, output, and intermediate steps to creating the inpainting mask. We obtain an inpainting mask to fill the clipped regions from the intersection of the saturated mask and semantic mask treated further by morphological operations. The resulting mask, $M_p$ is used as an input to the inpainting pipeline for the SDR image.

\subsection{Semantic Ordering Graph}
\label{sec:graph}

Now that the semantic regions and masks are defined, the ordering for the inpainting step next needs to be specified. As discussed before, this needs to consider the available information in the scene and therefore an encoding of the ordering of inpainting according to the identified semantic classes needs to be identified. To achieve this, we propose a novel data structure: an ordered semantic graph.

This data structure is a Directed Acyclic Graph where vertices correspond to semantic classes and edges correspond to the relationships between classes. The vertices contain a set of prompts associated with the class, and each edge has an associated value of the relative average difference in luminance between the classes. 

We first describe how to build this data structure, and then discuss its use for generating the ordered inpainting.

\subsubsection{Building the ordered semantic graph}

We take a data-driven approach to building the ordered semantic graph. We then take an existing HDR dataset consisting of $T$ images, and for the $t$-th image in the dataset $H_{t}$, we tone map it and compute a per pixel semantic labeling using the same approach as outlined in Section \ref{sec:semsegment}. For the $i$-th class we store this as $w_{i} \in [0,1]$. Then for every image, we initialize an instance of the ordered semantic graph $G_{t}$ as an unconnected set of vertices corresponding to the set of semantic classes. For each vertex, we compute:

\begin{equation}
    V_{i}(t) = \frac{\sum^{A}_{x} w_{i}(x) a(H_{t}(x)) L(H_{t}(x))}{\sum^{A}_{x} w_{i}(x)}\text{,}
\end{equation}

\noindent where $L$ computes the per-pixel luminance, $A$ is the image area, and $a$ is an albedo associated with the class (this can be estimated over the dataset using a method such as \cite{yu2019inverserendernet}). Then, we assign edges to this instance of the ordered semantic graph between every non-zero vertex and assign the edge a direction from the vertex with the highest value $V_{i}$ to the smallest with value $V_{j}$, and the weight assigned to the edge is the relative difference between the two vertices:
\begin{equation}
    e_{ij}(t) = V_{i}(t) - V_{j}(t).
\end{equation}
Now we have an instance of the semantic graph for one HDR image, we can build the final ordered semantic graph $G$ as an average of each graph generated for each HDR image in the dataset:

\begin{equation}
    e_{ij} = \frac{1}{N}\sum^{T}_{t=1}e_{ij}(t).
\end{equation}

\subsubsection{Generating ordered inpainting}

This now gives us a data-driven approach to generate the inpainting order based on luminance observed from a real HDR dataset. We first extract the semantic classes, as discussed in Section \ref{sec:semsegment}, and perform a \emph{max-cut} on the ordered semantic graph over the vertices corresponding to the semantic classes where inpainting is required. Fortunately, due to the limited number of semantic classes present in most images, typically less than five in all of our dataset, this max-cut is usually trivial to perform and is often a simple walk along the edges between valid vertices in $G$.

This results in an ordered set of vertices corresponding to semantic classes for which we can perform the inpainting step. This step is conditioned on the semantic class, and is typically expressed as a human readable `prompt'. One option is to associate a single prompt with each vertex, however this limits the types of images our method could produce. To solve this, we create a set of prompts for each semantic class and associate them with each vertex. Examples of this set of prompts for the \textit{sky} include $\{\text{\textit{``blue sky with clouds''}}, \text{\textit{``cloudy sky''}}, ... , \text{\textit{``clear blue sky''}}\}$. As the ordered set of vertices is traversed, one of the prompts stored at each vertex is sampled at random and used to generate an image. We also include wildcards in the prompts (e.g. for water, one of the prompts could be $\text{\textit{``water reflecting \#''}}$) where $\text{\textit{``\#''}}$ allows for the prompts from previous nodes from the graph traversal to be inserted into the current node, i.e. if $\text{\textit{``clear blue sky''}}$ is selected for a sky, then if a node containing water with a wildcard is selected, the prompt becomes $\text{\textit{``water reflecting clear blue sky''}}$. This illustrates one of the main features of the graph structure as it allows more consistent inpainting across the scene.

We also optionally allow for the prompts from the graph to be overridden by user specified prompts to enable inpainting using a wider range of prompts than are stored in the vertices, for example to add specific content to the clipped region.

\begin{figure}
\setlength{\tabcolsep}{0pt}
    \begin{tabular}{cccc}
\includegraphics[width=0.25\linewidth]{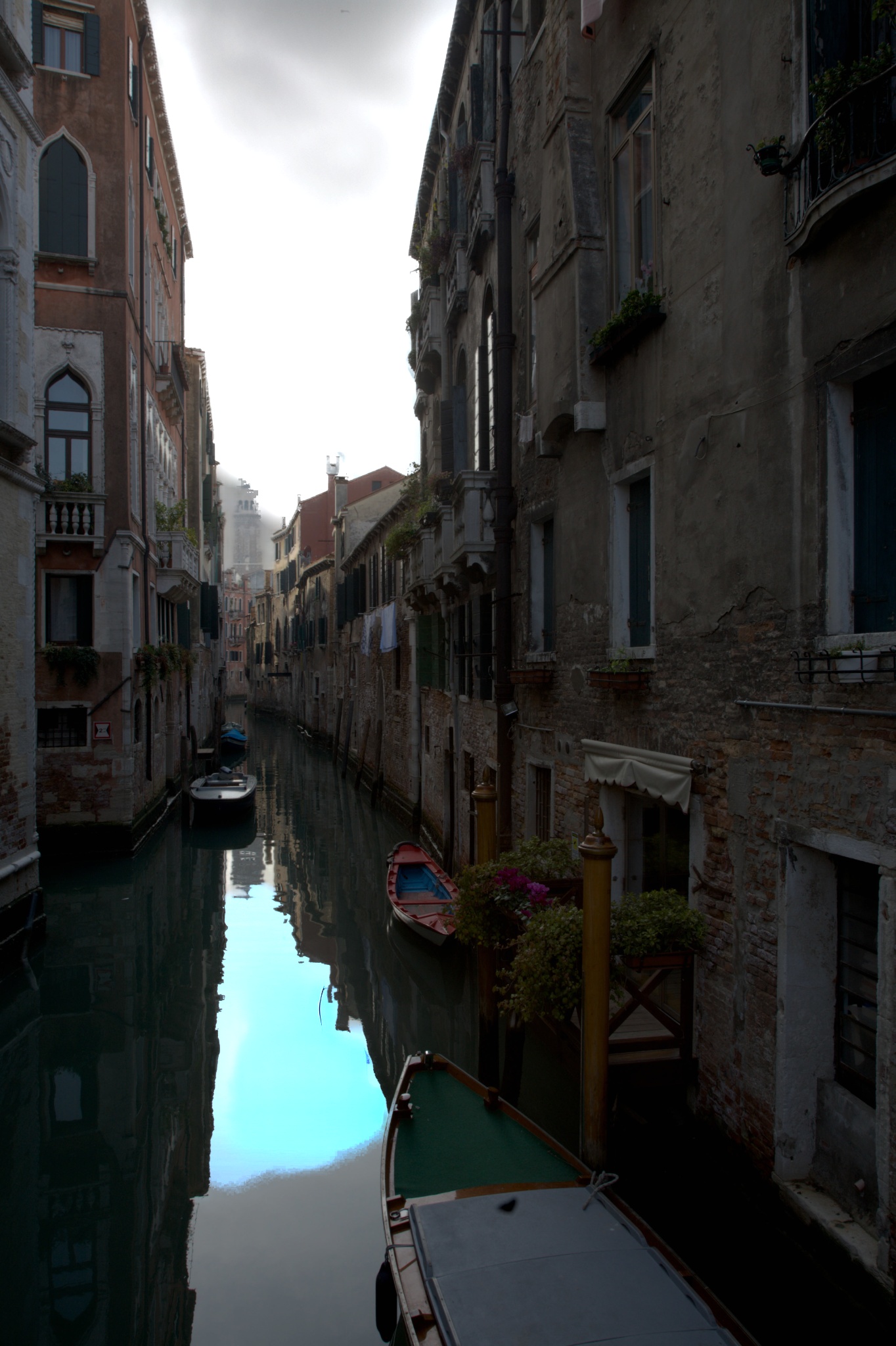} &  \includegraphics[width=0.25\linewidth]{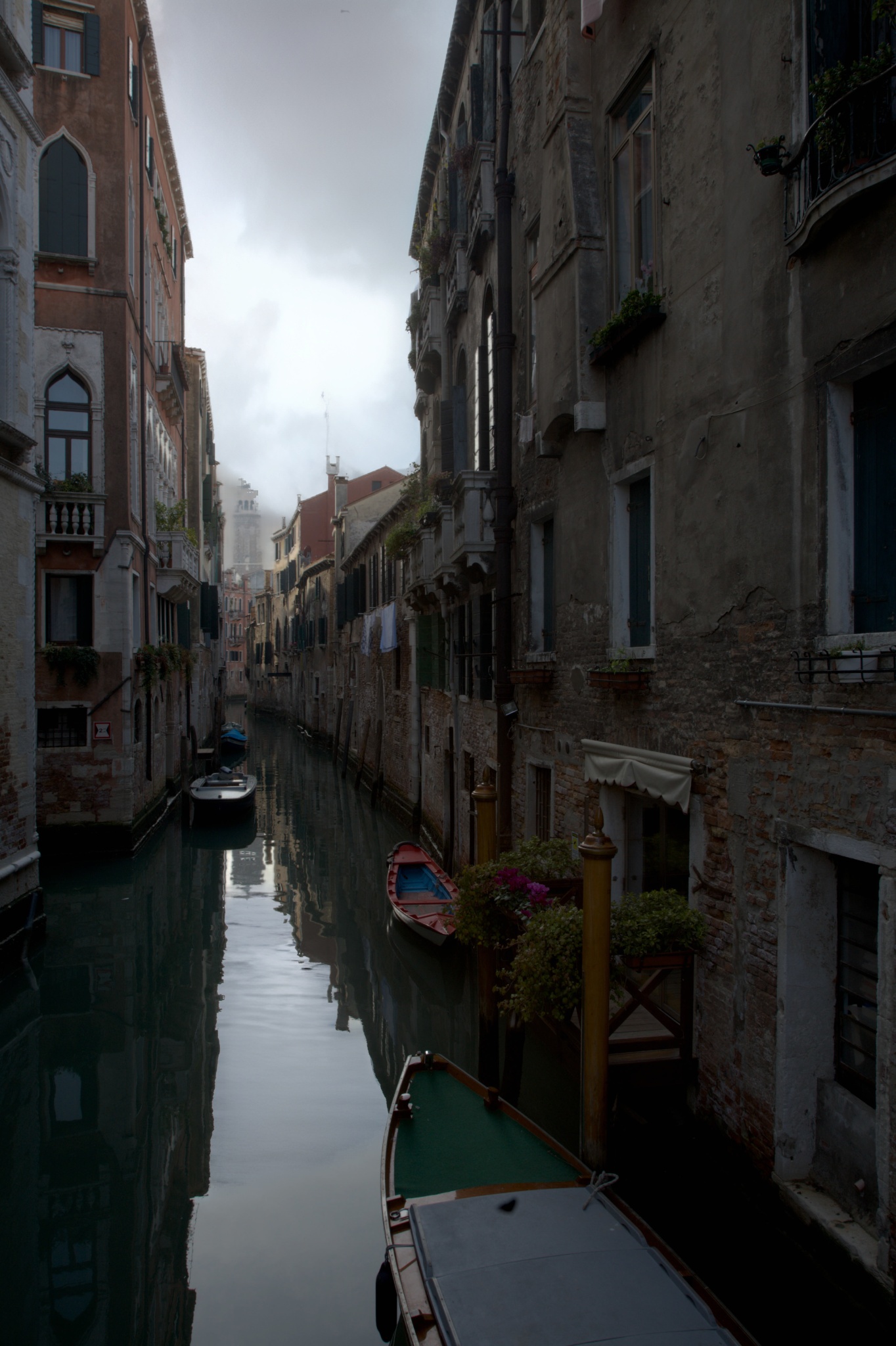} &
\includegraphics[width=0.25\linewidth]{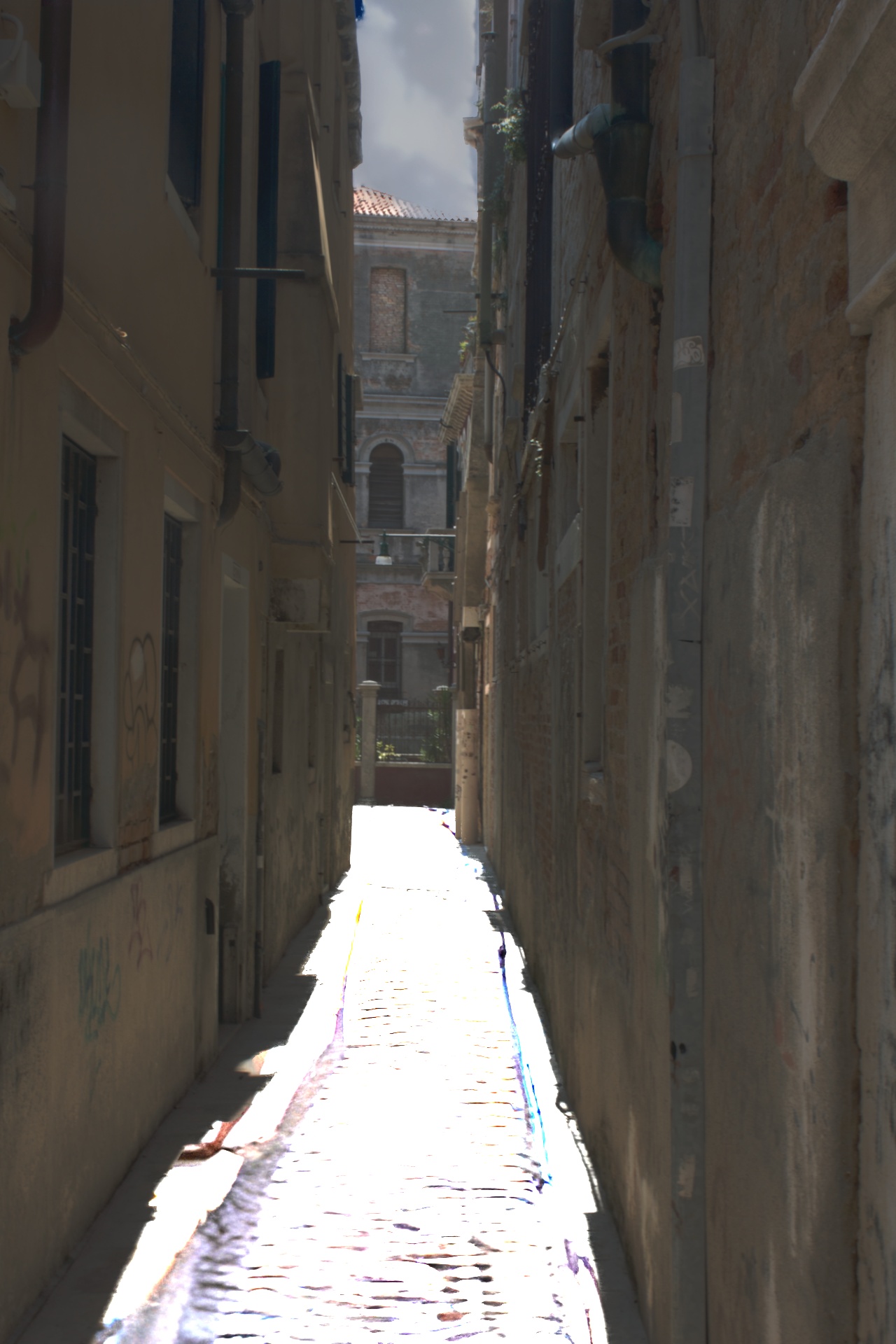} & \includegraphics[width=0.25\linewidth]{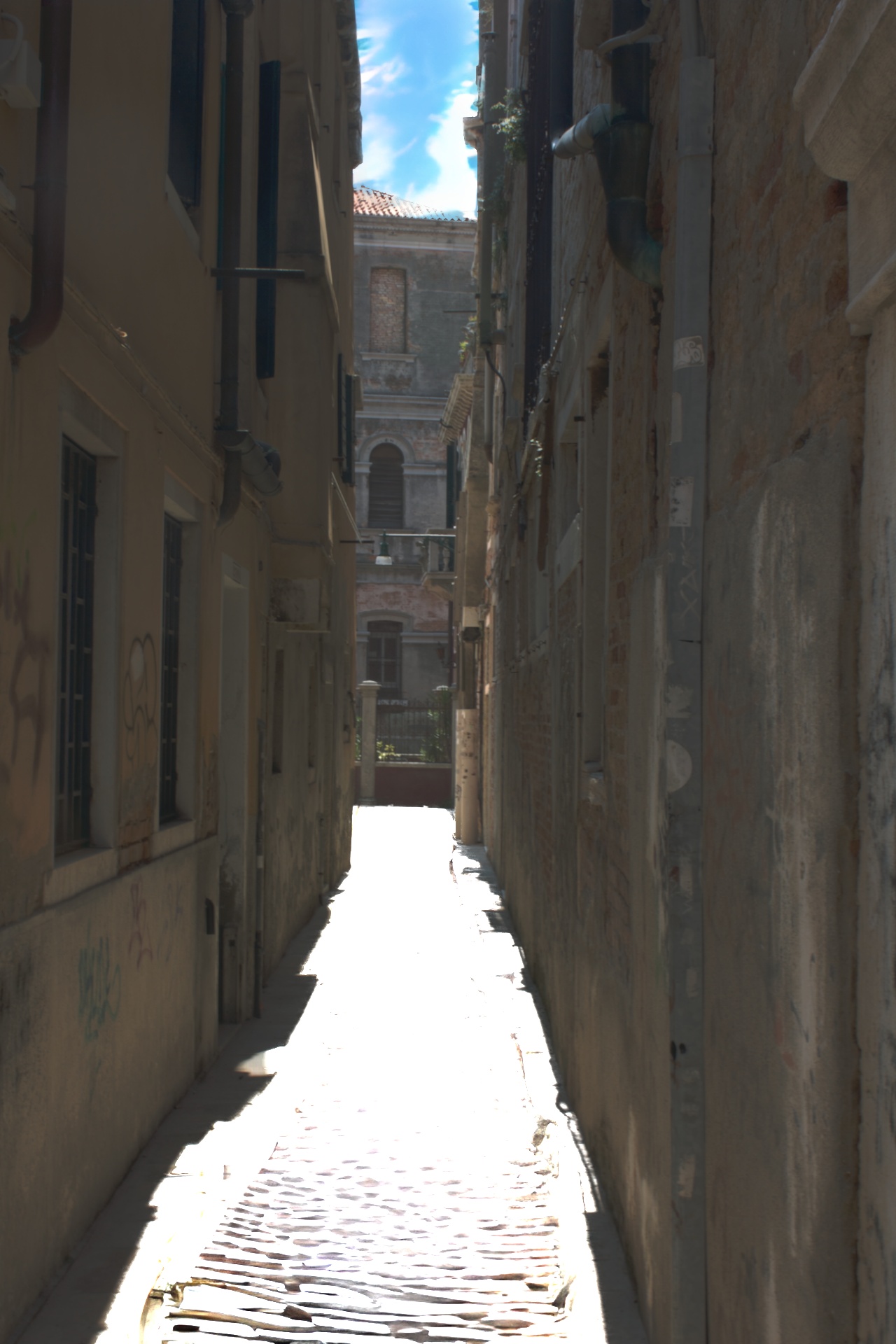} \\
w/o graph & with graph & w/o graph & with graph
\end{tabular}
\caption{The semantic ordering graph helps to ensure consistency of both inpainted content and its resulting luminance (shown at -2 stops).}
\label{fig:sogablation}
\end{figure}

To illustrate the importance of the semantic ordering graph, Figure \ref{fig:sogablation} shows two images inpainted with and without the semantic ordering graph. If the graph is not used, incorrect classes can be inpainted in reflections, for example, the left images show the reflection of a clear sky in the water whereas the sky is cloudy, and the right images illustrate the alleyway reflecting bright light whereas the sky was inpainted with darker clouds leading to inconsistent illumination.

\subsection{Exposure Estimation and Merging}
\label{sec:exposureestimation}

The inpainting process can use any generative model to inpaint the masked regions with the prompt generated from the Semantic Ordering Graph. Once, we obtain this hallucinated detail, $H_i$ for the i'th class in the masked region we need to estimate the exposure in which this region \emph{would} have had to have been captured in order to lead to over exposed pixels in the original image. This is the crucial step in our method which allows us to perform inverse tone mapping without requiring HDR data: we formulate this is a simple optimization process which has the advantage of closely mirroring the physical capture process of using bracketed exposures.

Therefore we need to find the exposure $e_{i}$ that when applied to the new inpainted content, results in clipped values. This is equivalent to finding what value the pixel with the smallest luminance in $H_i$ ($min(H_i) > 0$) would have to be scaled by to output the value 1, assuming the SDR values are normalized between $[0..1]$. This is then formulated as:

\begin{equation}
    2^{-e_{i}} \min(H_i) = 1,
\end{equation}

\noindent which has the trivial solution:

\begin{equation}
    e_{i} = \log_{2}(\min(H_i)).
\end{equation}

This predicts the exposure at which the inpainted content $H_i$ should be placed into. This is achieved by linearizing the input image, scaling it by the predicted exposure $e_{i}$, and then compositing the inpainted region into this exposure. In practice, rather than using the pixel with the smallest value for estimating exposure, we find the pixel corresponding to the 98\% smallest luminance in the inpainted region (the value of 98\% was found by a parameter sweep). This prevents outliers from artificially boosting the predicted exposure.

The linearization process maps the input image from $[0..255]$ to $\mathbb{R}^+$; i.e., re-establishing a linear relationship between input and output. To ensure that the linearization process preserves details in the well exposed and near saturated regions which our method will not modify, we perform linearization using a gamma curve alongside an existing trained U-Net architecture similar to \cite{Eilertsen+17}. Furthermore, a soft semantic mask is used as guidance to add the inpainted contents onto the predicted exposure bracket. Alpha matting with a border of 10 pixel radius is used on the saturation mask computed earlier to generate a pixel precise guide which helps overlay the inpainted details. This ensures that the inpainted region does not overwrite well-exposed pixels from the input image and only modifies those regions beyond the saturation threshold.

Now that we have the correctly exposed content for one semantic class, we can repeat this for other regions and semantic classes. Before merging, these inpainted regions are linearized and propagated into the other predicted exposures to ensure consistent content in each bracket. Finally, each of these brackets are merged using the method proposed by \cite{Debevec+97}.

\subsection{Parameter Selection}
\begin{figure}
    \centering
    \includegraphics[width=\linewidth]{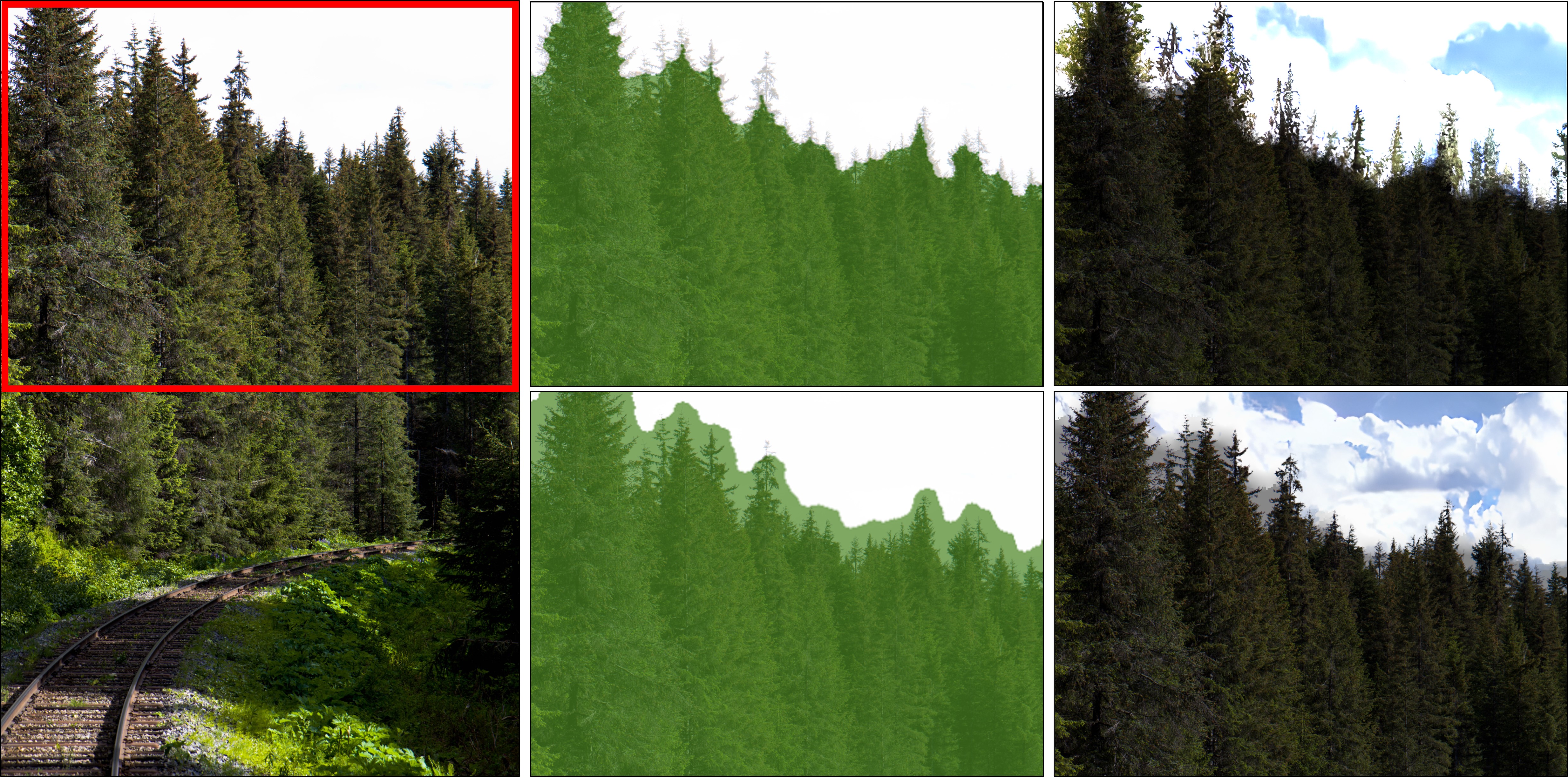}
    \caption{\textit{Regulating $\alpha$ and $\beta$ parameters in morphological opening}. The controlled dilation parameter ensures that information from outside the inpainting mask does not influence the reinvented details.  }
    \label{fig:var_params}
\end{figure}

As described in Section \ref{sec:semsegment} we use two parameters $(\alpha, \beta)$ which denote the radius of the disk structuring elements to guide the morphological opening of the binary masks. These parameters regulate the extent of boundary information that goes into the diffusion-based inpainting. Figure~\ref{fig:var_params} shows how varying parameters influence the inpainted details. The masks generated using $(2,5)$ (top) show how parts of the neighboring \textit{vegetation} information are included and it creates improper details in the inpainted results. The mask generated using $(10,5)$ (bottom) avoids this by carefully focusing the information going into the stable-diffusion module. Empirically, the range is set as $\alpha, \beta \in [1, 10]$.

The choice of pairs are considered as a function of the node relationships in the semantic graph. \textit{Sky--vegetation}, \textit{sky-ground}, \textit{cityscape-vegetation} and such node pairs which share a high frequency boundary require $\alpha > \beta$ by a significant margin. For low frequency edges the pair can be adjusted and a small margin yields good results.

\section{Results}
We present quantitative results via traditional metrics to show that DITMO is comparable to state-of-the-art and provide qualitative and a participant-based experiment to highlight how DITMO can lead to better perceptual outcomes.
For these results, we employed FastFCN, trained on the ADE20K dataset for semantic labels \cite{fastfcn+19}, and two diffusion models for inpainting within the DITMO framework: a text-to-image stable diffusion model \cite{stableDiffusion}, and ControlNet \cite{zhang2023adding}. We use these two models to illustrate the generalizability of DITMO, and choose these methods as they are recent prompt driven diffusion models

\subsection{Quantitative comparisons}
We compare DITMO with the following state-of-the-art methods: \cite{Eilertsen+17} (EIL),  \cite{Santos+20} (SAN),  \cite{Liu+20} (LIU), \cite{Marnerides+18} (MAR), LaNet \cite{Yu+21} (YU). We present results for two datasets, the Fairchild dataset \cite{Fairchild+07} used for comparisons with ground-truth and a selection of images containing clipped regions from the MIT-5k dataset \cite{Bychkovsky+11} used for no reference comparisons.

\begin{table*}
\vspace{-2.5mm}
\begin{tabular}{|l|rr|rrrr|}
\hline
                    & \multicolumn{2}{c|}{MIT5K}&\multicolumn{4}{c|}{FairChild}\\
\hline
                     & \textbf{NoR-VDP++ ($\uparrow$)} 
                     & \textbf{PU21-Pique ($\downarrow$)}
                     & \textbf{NoR-VDP++ ($\uparrow$)}      & \textbf{PU21-PSNR ($\uparrow$)}        & \textbf{HDRVDP3 ($\uparrow$)} & \textbf{PU21-Pique ($\downarrow$)}    \\
\hline
\textbf{SAN}    & 50.331    &    12.397       & 54.087                & 28.609               & \textcolor{red}{6.495}  &    12.762           \\
\textbf{YU}  & 51.038         & 16.651       & 54.175            & 27.459               & \textcolor{blue}{6.479}         & 16.822       \\
\textbf{MAR} & 51.836         &  {12.248}  & 50.657               & 22.740               & 5.642        & {10.826}        \\
\textbf{EIL}   & 50.606         & 14.329        & {54.842}               & 28.303               & 6.459        & 14.227    \\
\textbf{LIU}    & \textcolor{red}{54.578}     & 20.419       & 53.406              & \textcolor{red}{34.718}               & 6.003        & 22.674        \\
\textbf{DITMO (SD)}   & {53.226}  & \textcolor{blue}{8.640}      & \textcolor{blue}{57.051} & {29.989} & 6.175  & \textcolor{red}{8.183} \\

\textbf{DITMO (CN)}   & \textcolor{blue}{53.756}  & \textcolor{red}{8.290}     & \textcolor{red}{58.378} &  \textcolor{blue}{31.321} & 6.080  & 
\textcolor{blue}{8.554} \\
\hline
\textbf{GlowGAN*}    &     52.034     &    11.591   &      \underline{58.088}         &      22.238          &   \underline{6.312}      &    10.143  \\
\textbf{DITMO (CN)*}     & \underline{54.337} & \underline{8.789}  & 57.230 & \underline{27.980} & 6.306  & \underline{7.801}\\
\hline
\end{tabular}

\caption{Comparison between state-of-the-art iTMOs and variants of our method. The top and second best performance are highlighted in red and blue. GlowGAN* results were computed on a subset of the images conforming to their set classes and computed at a lower resolution; results have been provided by the authors. For fairness, DITMO (CN)* uses the same images and resolution as GlowGAN*. The best performance in the subset is underlined.}
\label{tab:results:ref}
\vspace{-2.5mm}
\end{table*}

The reference-based metrics we used only on the Fairchild dataset are PU21-PSNR \cite{Hanji+22} and HDR-VDP3 \cite{Mantiuk+23}. The no reference metrics we used are PU21-Pique \cite{Hanji+22} based on Pique \cite{Venkatanath+15}, and NoR-VDPNet++ \cite{Banterle+23}, a no reference version of HDR-VDP2.2.1.

Results for both datasets are shown in Table \ref{tab:results:ref} including DITMO using \cite{stableDiffusion} (referred to as DITMO (SD)) and \cite{zhang2023adding} (referred to as DITMO (CN)). This shows that our method performs best or is close to best for the no reference metrics, and is competitive for the full reference metrics. Although our method performs well in PU21-PSNR, these full reference metrics are unfavorable to our method as the inpainting is unlikely to produce content which exactly matches the ground truth, however, it does synthesize plausible and high frequency content in clipped regions. We also analyzed the dynamic range of the reconstructed HDR images, please see the supplementary document for values. This shows that the dynamic range of our method is comparable across scenes to other methods, while also generating more details in clipped regions.

We also compare against GlowGAN \cite{Wang+23} as this method has similarities to our work. This comparison was performed for a subset of the Fairchild images and MIT5K images and was generated by the authors of GlowGAN; these images were applied to a subset because their current method supports a limited number of cases (i.e., lightning, sunsets, fireplaces, landscape, windows, etc.) that need to be manually selected. Additionally, these images are computed at a lower resolution than our method (due to memory limitations) and are indicated by the * in Table \ref{tab:results:ref}. We repeat our metrics on the same subset of images used for GlowGAN, and lower our resolution to the same one GlowGAN uses for fairness. This shows that our method is competitive with, or outperforms, GlowGAN for most metrics. It is also worth noting that the computation time for inference for our method is around 45 seconds per image on Nvidia RTX5000, whereas the reported computation time on an Nvidia A40 is 14 minutes for GlowGAN meaning that inference with our method is an order of magnitude faster than GlowGAN.

Note that the aim of DITMO is not to surpass state-of-the-art in these objective metrics (in particular for comparisons with ground truth), rather it is adding information and details that make the images more visually compelling. The quantitative results in this subsection demonstrate that the overall quality of the images produced by DITMO is comparable to or exceeds the state-of-the-art methods.

\subsection{Experiment}
Since objective metrics may not be an effective criterion to judge hallucinated material, due to their broadly comparative properties, we ran a subjective experiment (N = 20) in order to identify whether the proposed method improved perceived quality. 

\textbf{Design:} A rating design was used to quantify the quality of the images due to its straightforward nature. Four iTMOs (\cite{Eilertsen+17}, \cite{Marnerides+18}, \cite{Santos+20}, and DITMO) were selected due to their performance in previous related works, their popularity and ease of use; furthermore, we wanted to keep the number of methods low to reduce fatigue in the task. The dependent variable used was the perceived quality on a scale from 1 to 10. The question asked of the participants was: ``Rate the image (1-10) in terms of visual quality''. Images were randomly sorted and presented on an HDR display. Each participant was initially shown 2 scenes for training purposes. The were then shown 11 scenes for which they were asked to rate the individual images. 

\textbf{Materials:} The images were generated for 13 scenes from the MIT-FIVEK dataset for the four iTMOs. A 55-inch SIM2 HDR display with a peak brightness of 6,000 nits was used as the display device. A customized software package was used for displaying the images and collecting the data. Images were display referred and are shown in the supplementary material.

\textbf{Participants:} 20 people from the local university volunteered for the experiment (M=11, F=9, age($\mu$) = 30). 

\textbf{Results:} The non-parametric statistical test Friedman (a non-parametric equivalent of a repeated measures ANOVA) was used to analyze the results and identify overall significance between methods. The results showed a significant difference among the methods $\chi^2$(3) = 23.964.  Kendall’s co-efficient of Concordance as $W$ = 0.395, p $<0.01$. Pairwise comparisons were subsequently conducted in order to directly compare between the methods.  Table \ref{tab:results:stats} shows the results with groupings indicating methods that did demonstrate significant difference among themselves. There is a significant difference between our method (DITMO) and the rest of the methods, p $<$ 0.01, but no further significant differences between the other methods. The figures in the brackets show the mean rank for each method across all the images. 

\begin{table}[tp]
\vspace{-2.5mm}
    \centering
    
    \begin{tabular}{llll}
           \mbl{c12} DITMO (1.35) \mbr{c12}{red} & \mbl{c13} MAR (2.55) & EIL (3.00) & SAN (3.10)  \mbr{c13}{blue} \\
    \end{tabular}

    \caption{\small{Overall order of the experimental results based on statistical significance. Colored groupings demonstrate no significant changes using pairwise comparisons at p < 0.05. Numbers in brackets show mean rank.}}
    \label{tab:results:stats}
    \vspace{-2.5mm}
\end{table}

\subsection{Qualitative comparisons}
\begin{figure}
\vspace{-2.5mm}
    \centering
    \includegraphics[width=\linewidth]{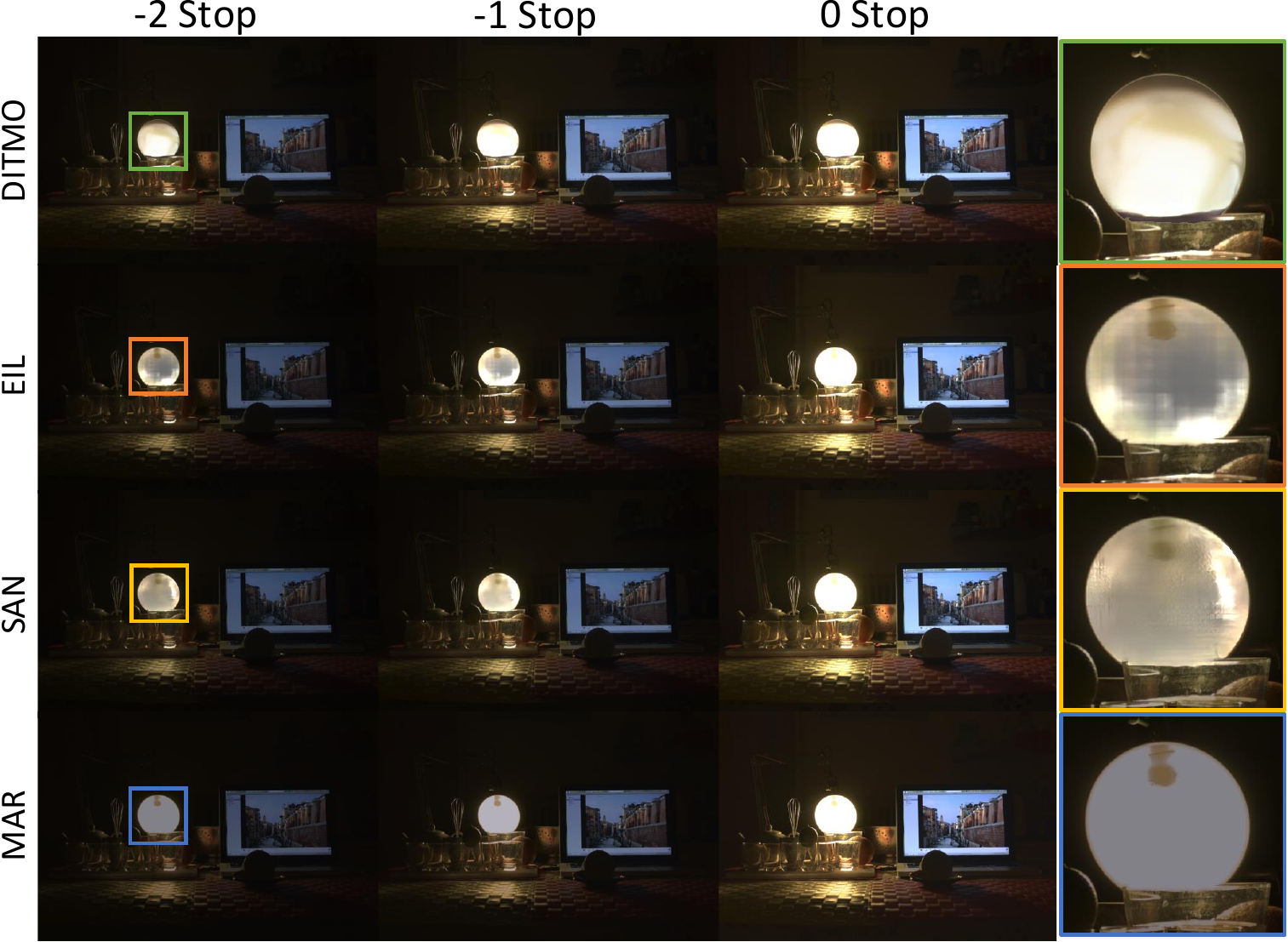}
    \caption{In this comparison, DITMO adds the bulb to the overexposed lampshade in the desk lamp leading to more plausible results than other methods.}
    \label{fig:results:qual:lightbulb}
\vspace{-2.5mm}
\end{figure}

\begin{figure}[H]
    \centering
    \includegraphics[width=.9\linewidth]{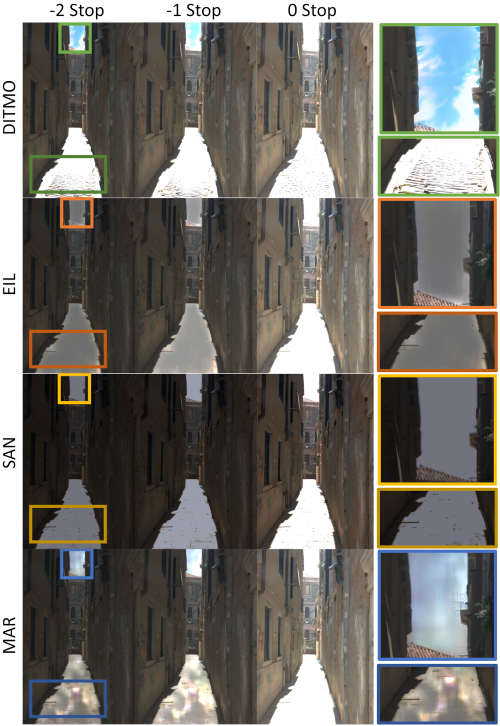}
    \caption{In this comparison DITMO generates cobblestones in the over-exposed floor area and the other methods are not able to generate details in these images.}
    \label{fig:results:qual:cobbles}
\end{figure}

\begin{figure}[H]
    \centering
    \includegraphics[width=.9\linewidth]{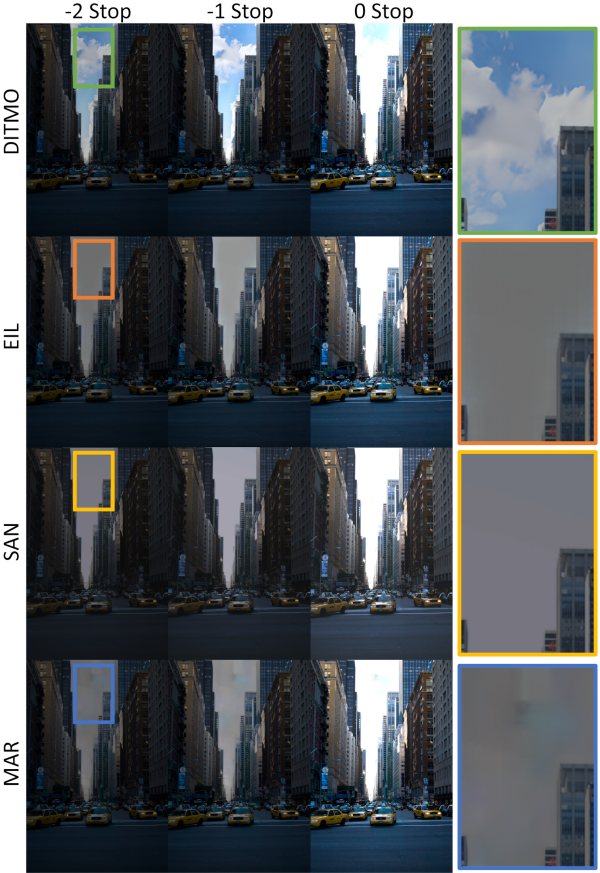}
    \caption{In this city scene, DITMO generates a partially cloudy sky in the over-exposed sky area whereas the other methods do not reconstruct this.}
    \label{fig:results:qual:ny}
\end{figure}

In this section, we show and analyze some of the DITMO outputs. The figures in this subsection are presented using three exposures. We first compare with other leading methods. Figure \ref{fig:results:qual:lightbulb} shows how DITMO reconstructs the filament in the light in an area which was overexposed in the original (see inset). Similarly, Figure \ref{fig:results:qual:cobbles} demonstrates how DITMO hallucinates cobblestones on the ground which was over exposed (see inset); other methods produce some reconstruction but it is fairly limited comparatively. Finally, Figure \ref{fig:results:qual:ny} shows reconstruction of the sky with cloudy details. Figure \ref{fig:Glow} shows results comparing our method to GlowGAN and illustrates that our method inpaints more details in clipped regions.

\begin{figure*}[htbp]
\centering
\setlength{\tabcolsep}{0pt}
\begin{tabular}{cccccccc}
\multicolumn{2}{c}{Input Image} & \multicolumn{2}{c}{-2} & \multicolumn{2}{c}{-1} & \multicolumn{2}{c}{+1} \\
\multicolumn{2}{c}{\includegraphics[width=.234\textwidth]{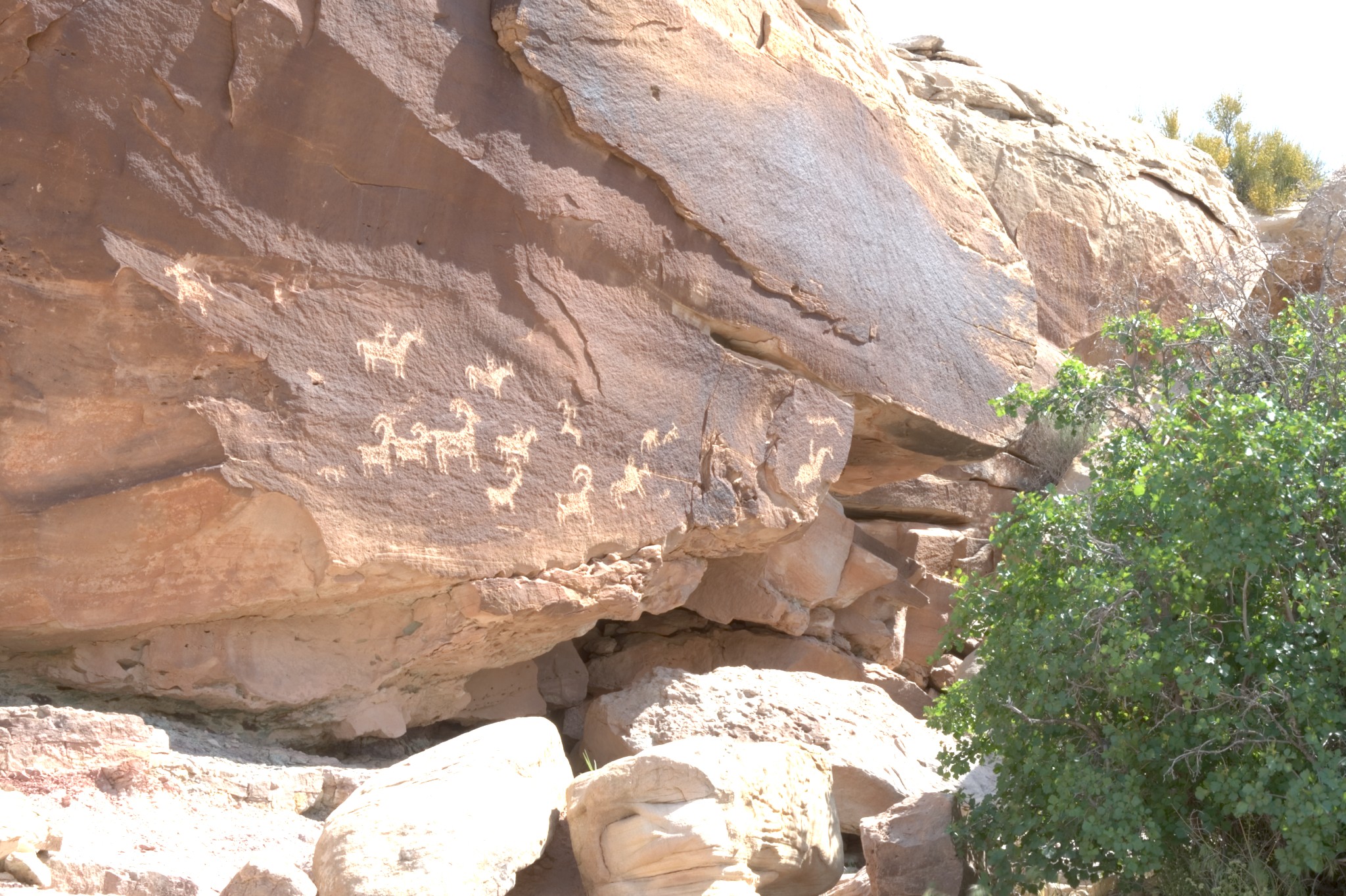}} & \multicolumn{2}{c}{\includegraphics[width=.234\textwidth]{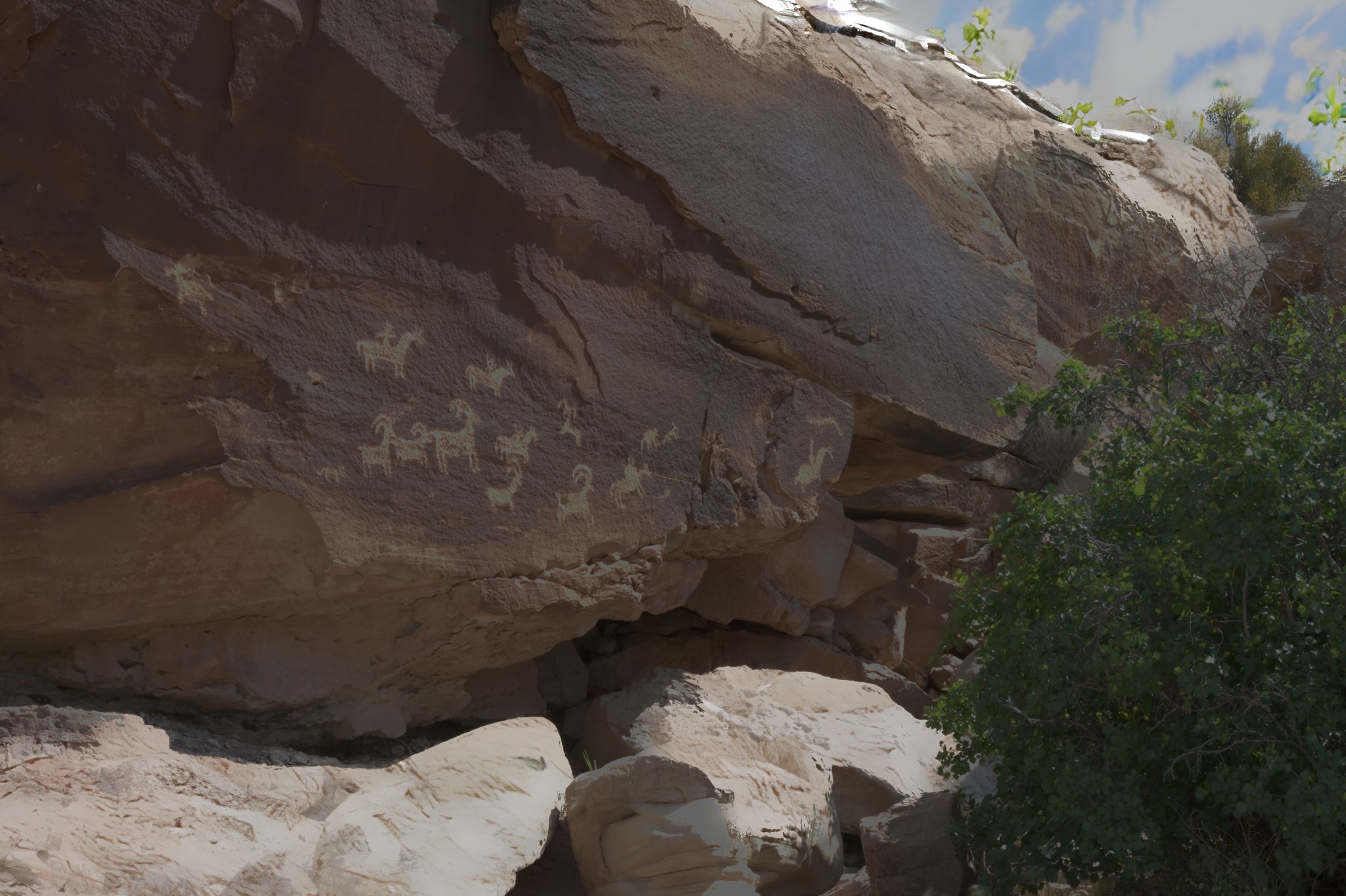}} & \multicolumn{2}{c}{\includegraphics[width=.234\textwidth]{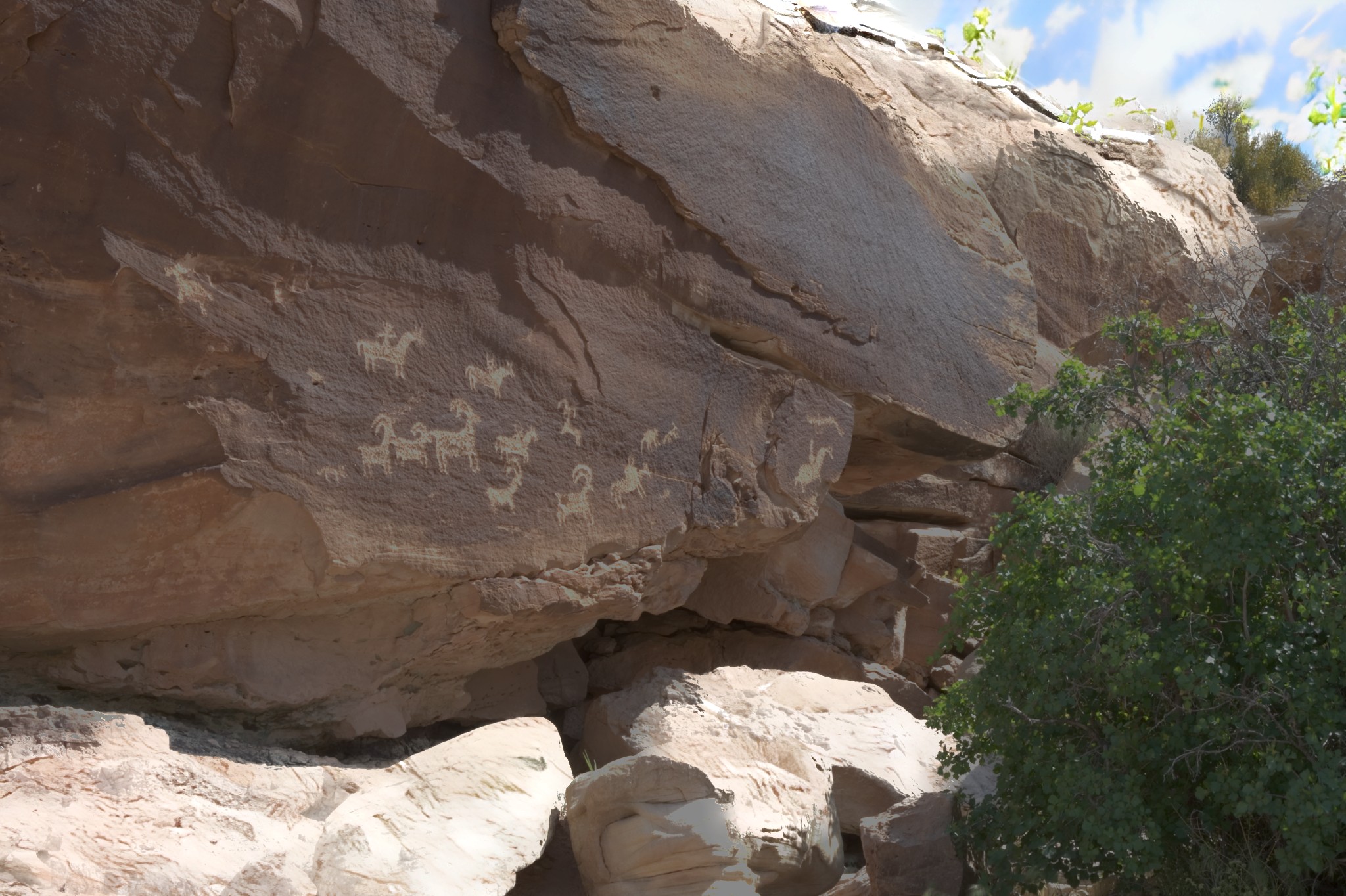}} & \multicolumn{2}{c}{\includegraphics[width=.234\textwidth]{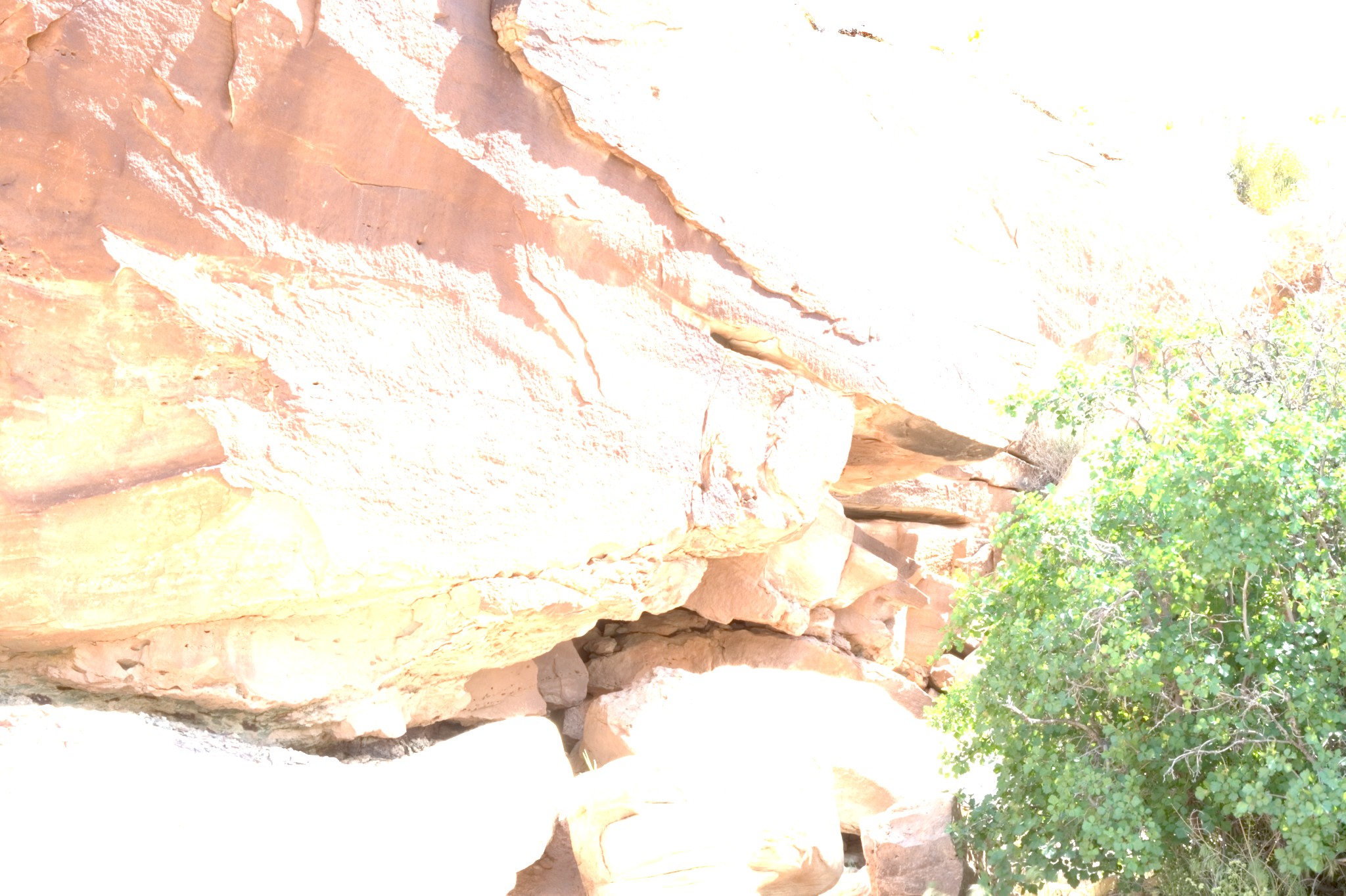}} \\
\multicolumn{2}{c}{\includegraphics[width=.234\textwidth]{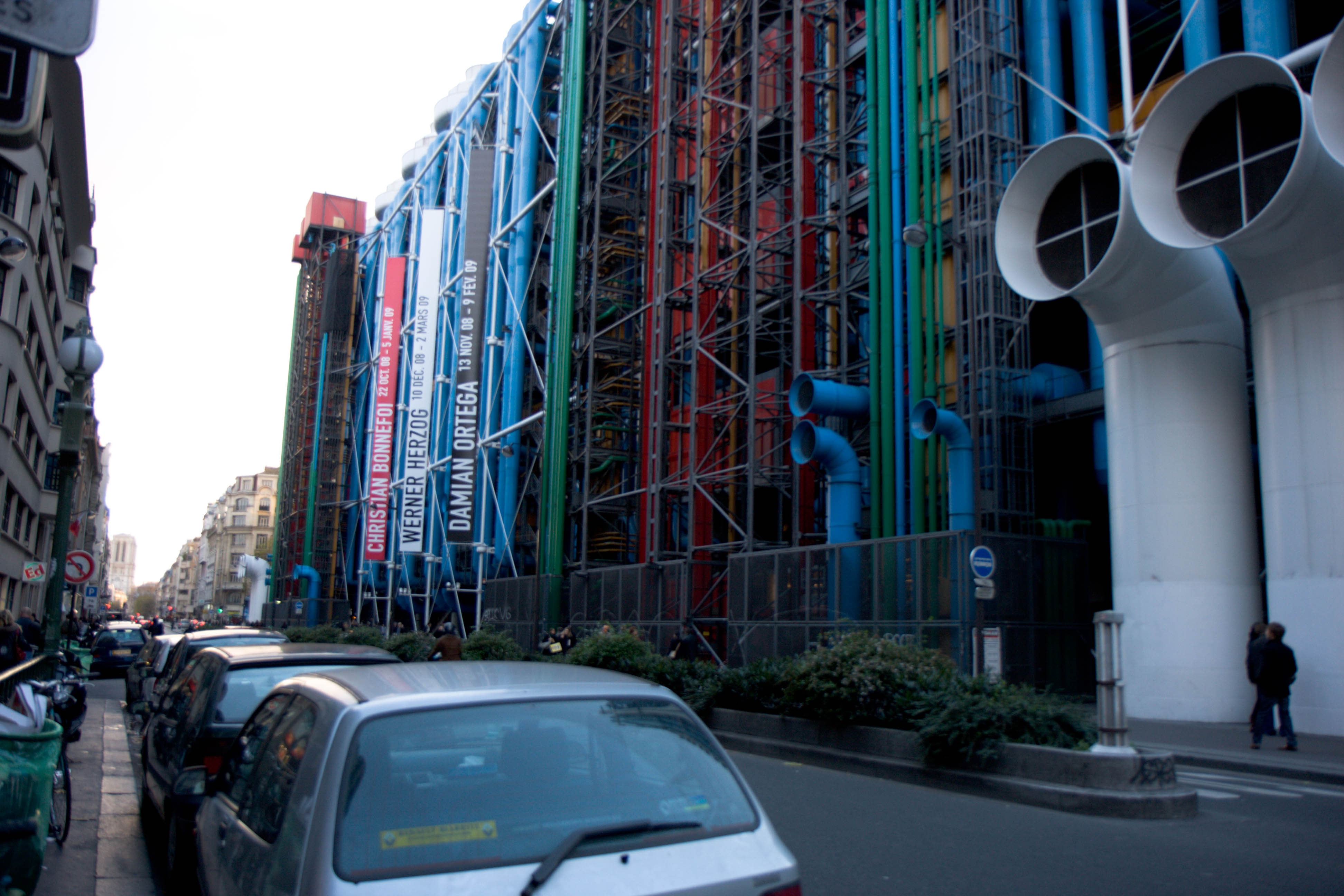}} & \multicolumn{2}{c}{\includegraphics[width=.234\textwidth]{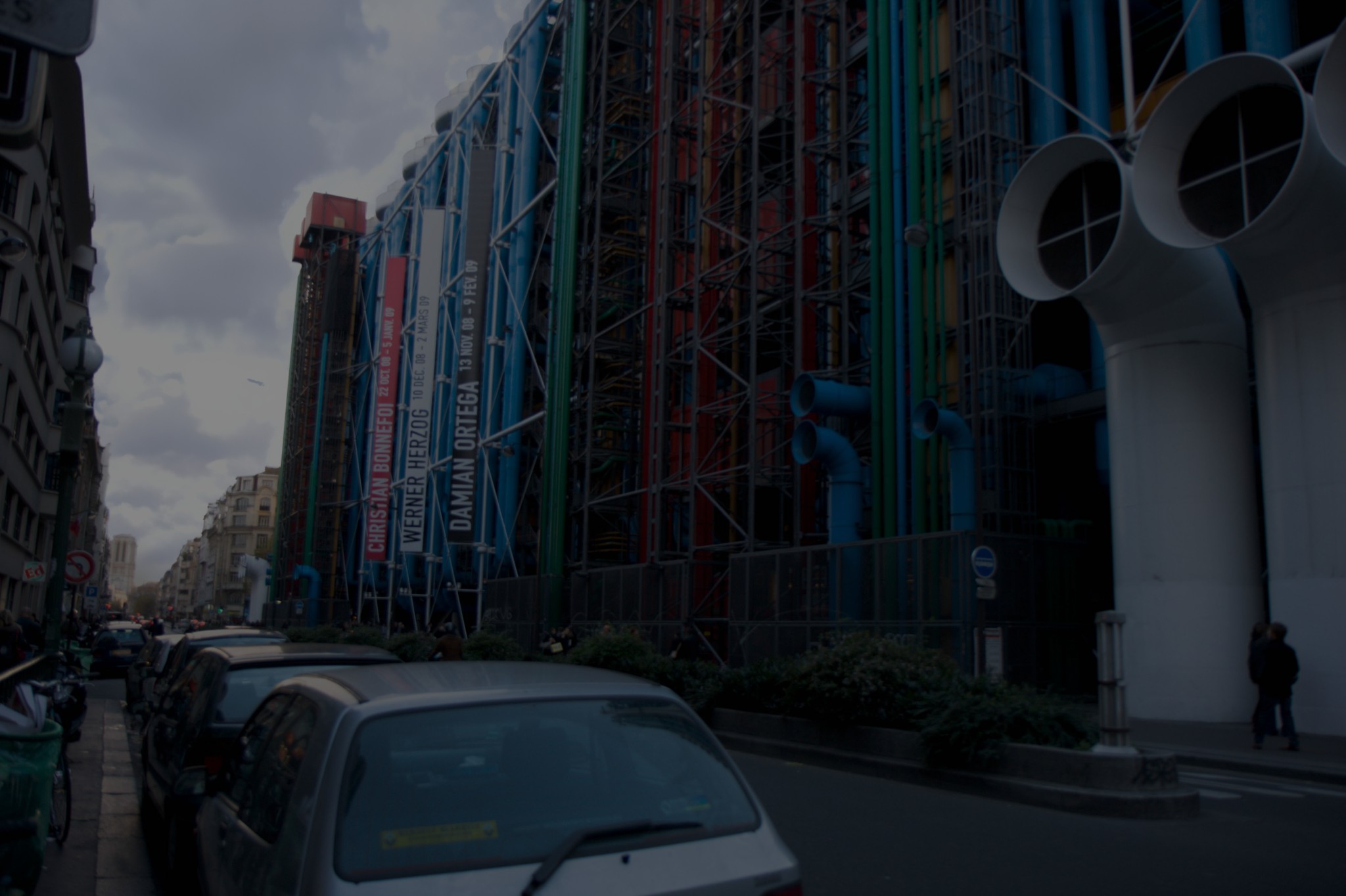}} & \multicolumn{2}{c}{\includegraphics[width=.234\textwidth]{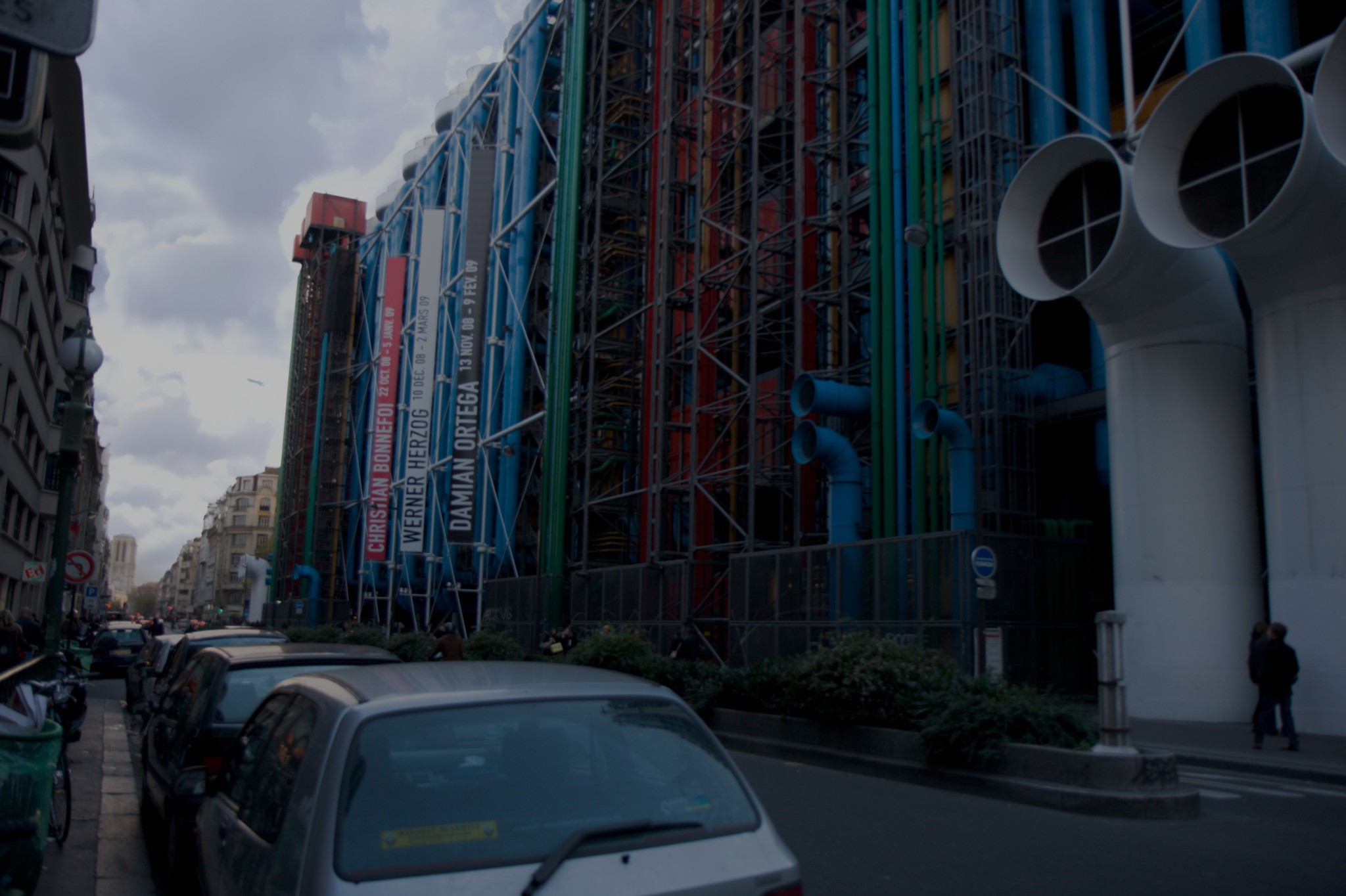}} & \multicolumn{2}{c}{\includegraphics[width=.234\textwidth]{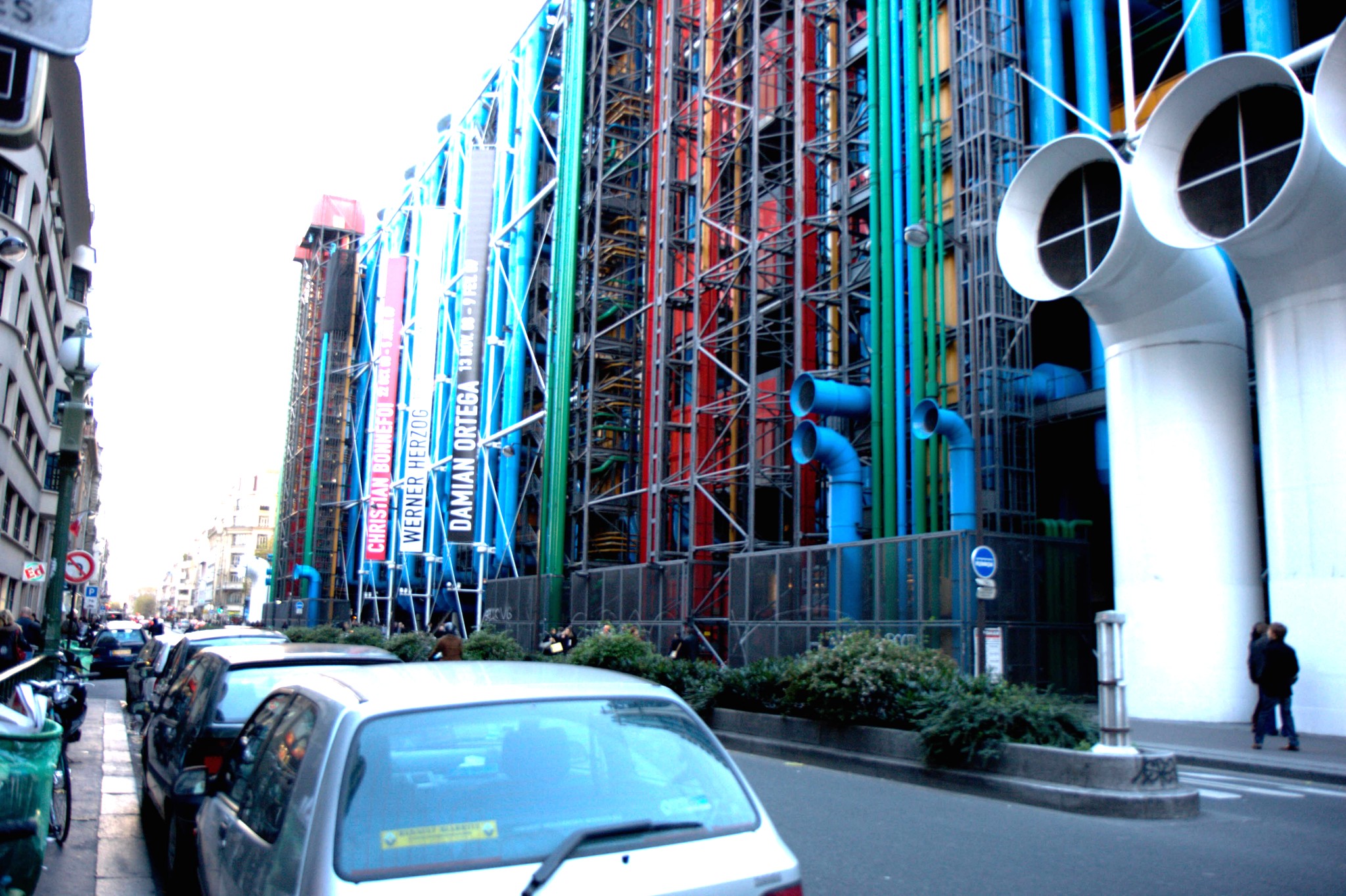}} \\
Input Image & -2 & -1 & +1 & Input Image & -2 & -1 & +1 \\
\includegraphics[width=.1171\textwidth]{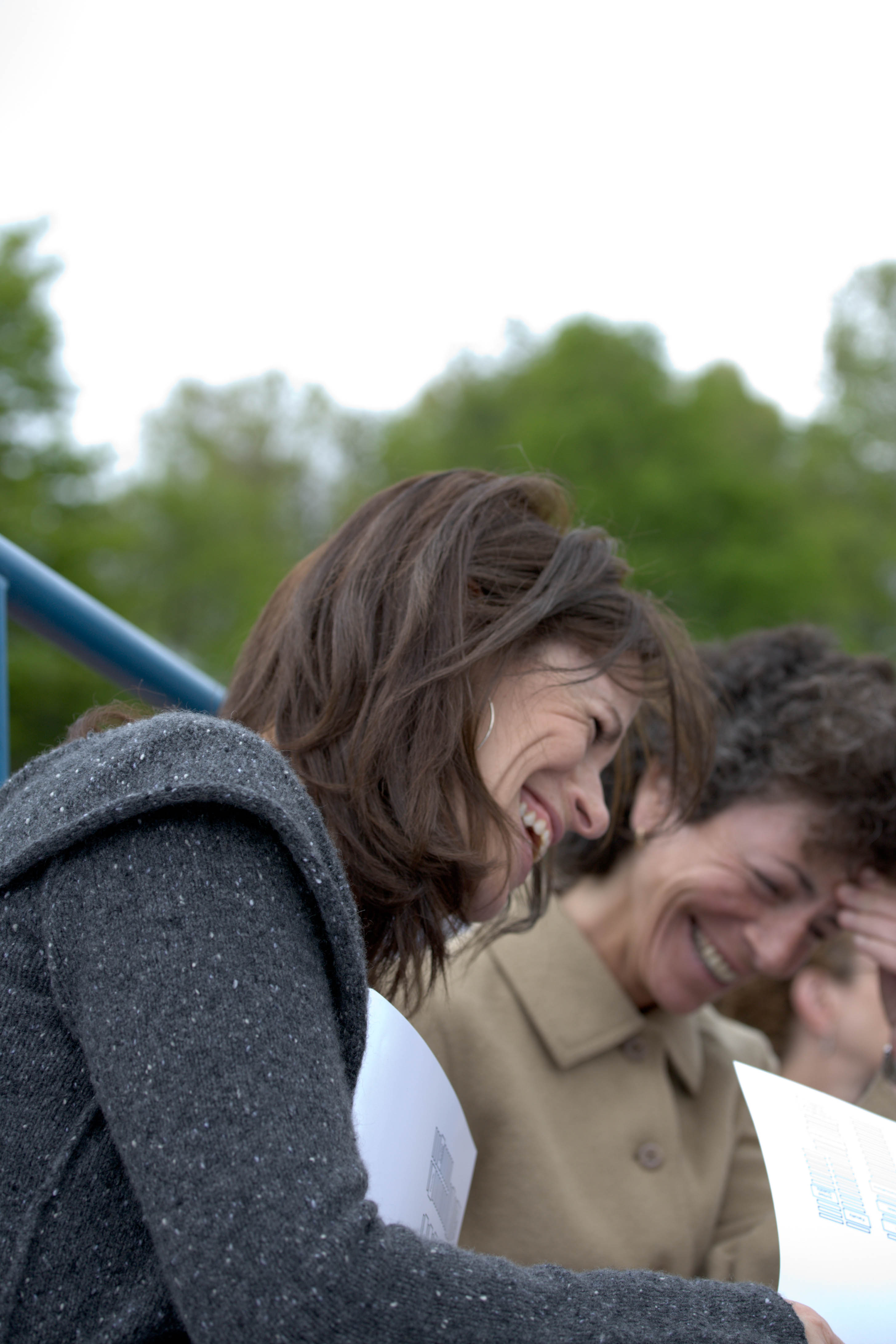} &
\includegraphics[width=.1171\textwidth]{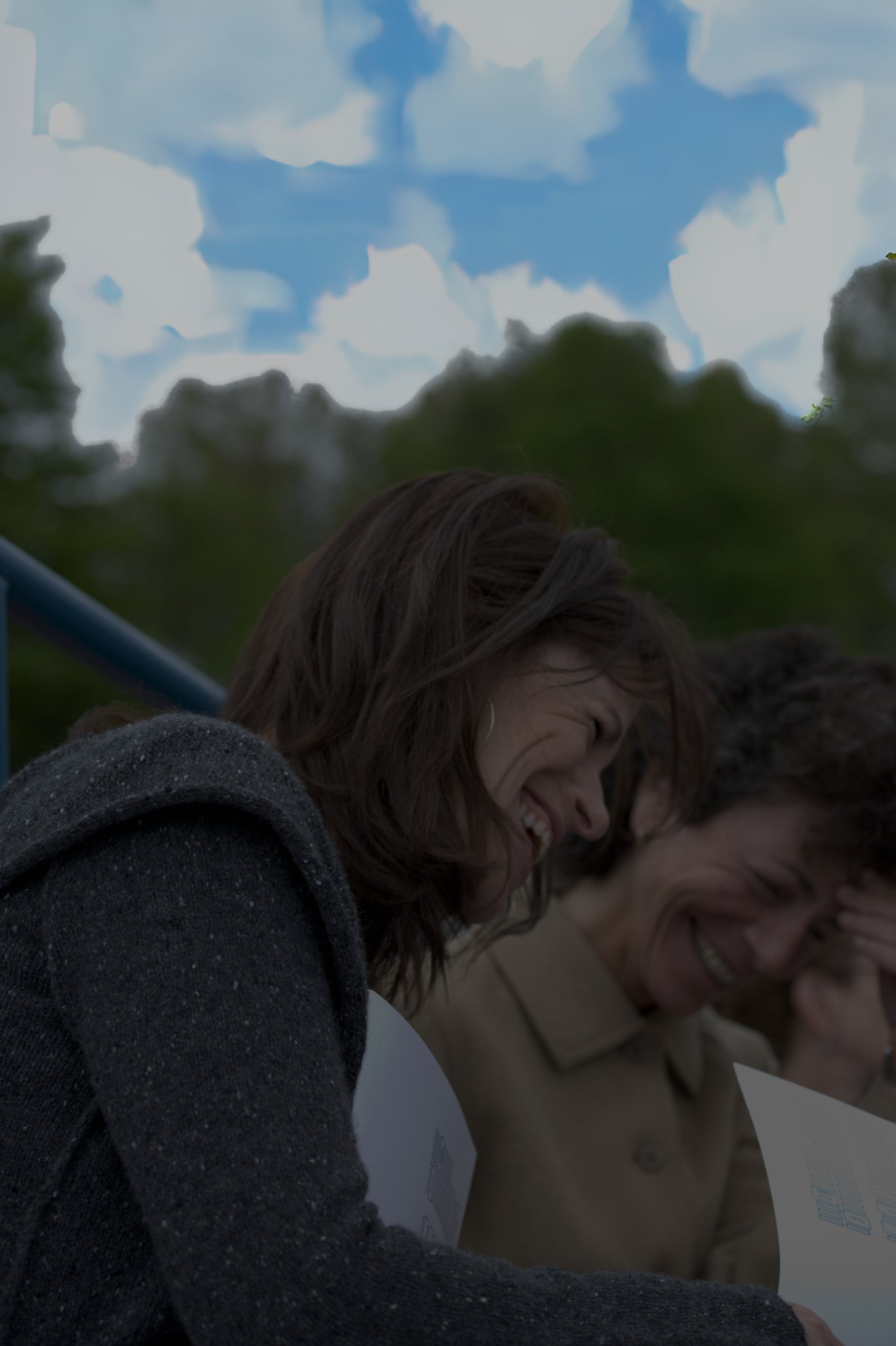} & 
\includegraphics[width=.1171\textwidth]{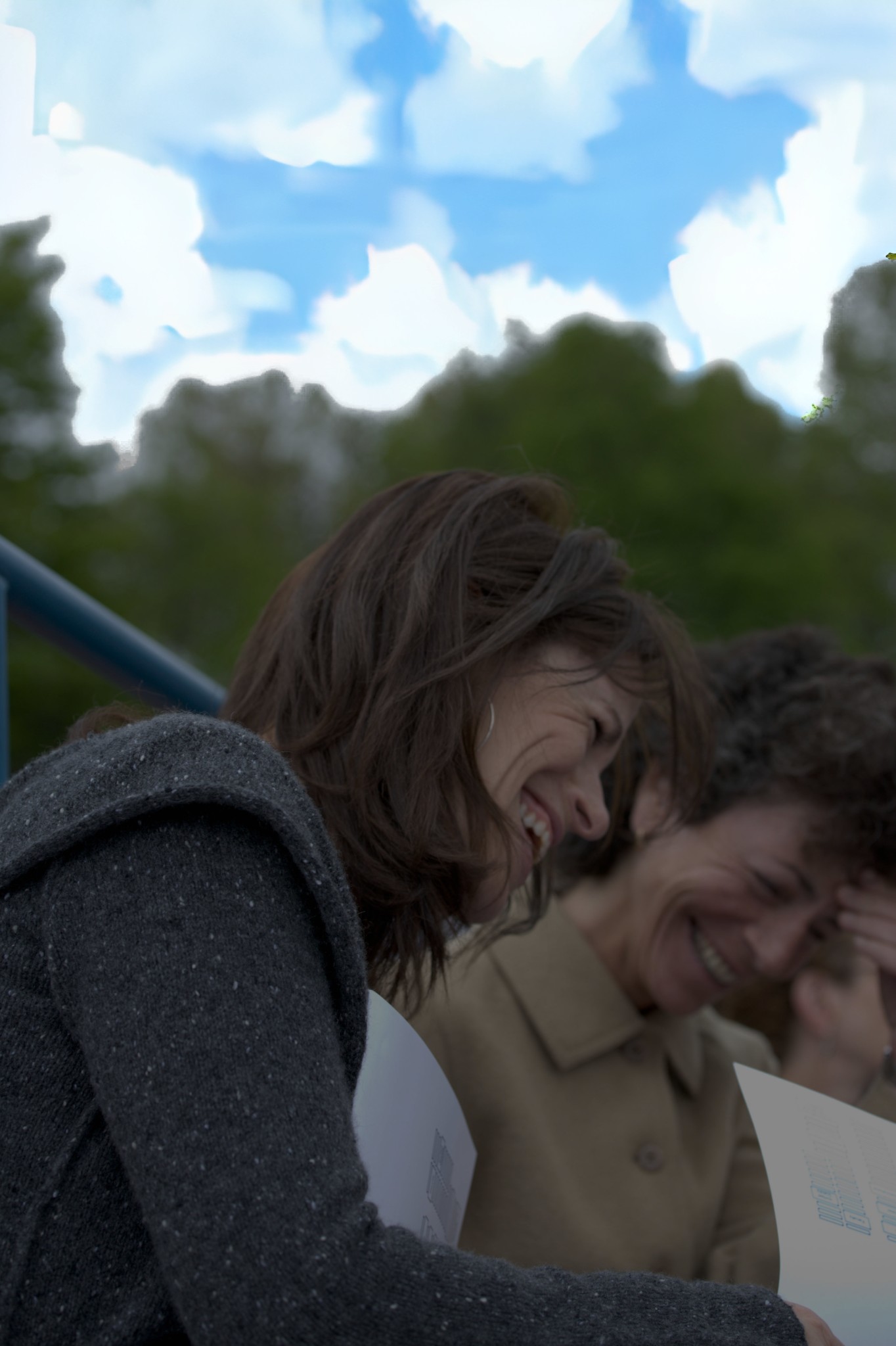} & 
\includegraphics[width=.1171\textwidth]{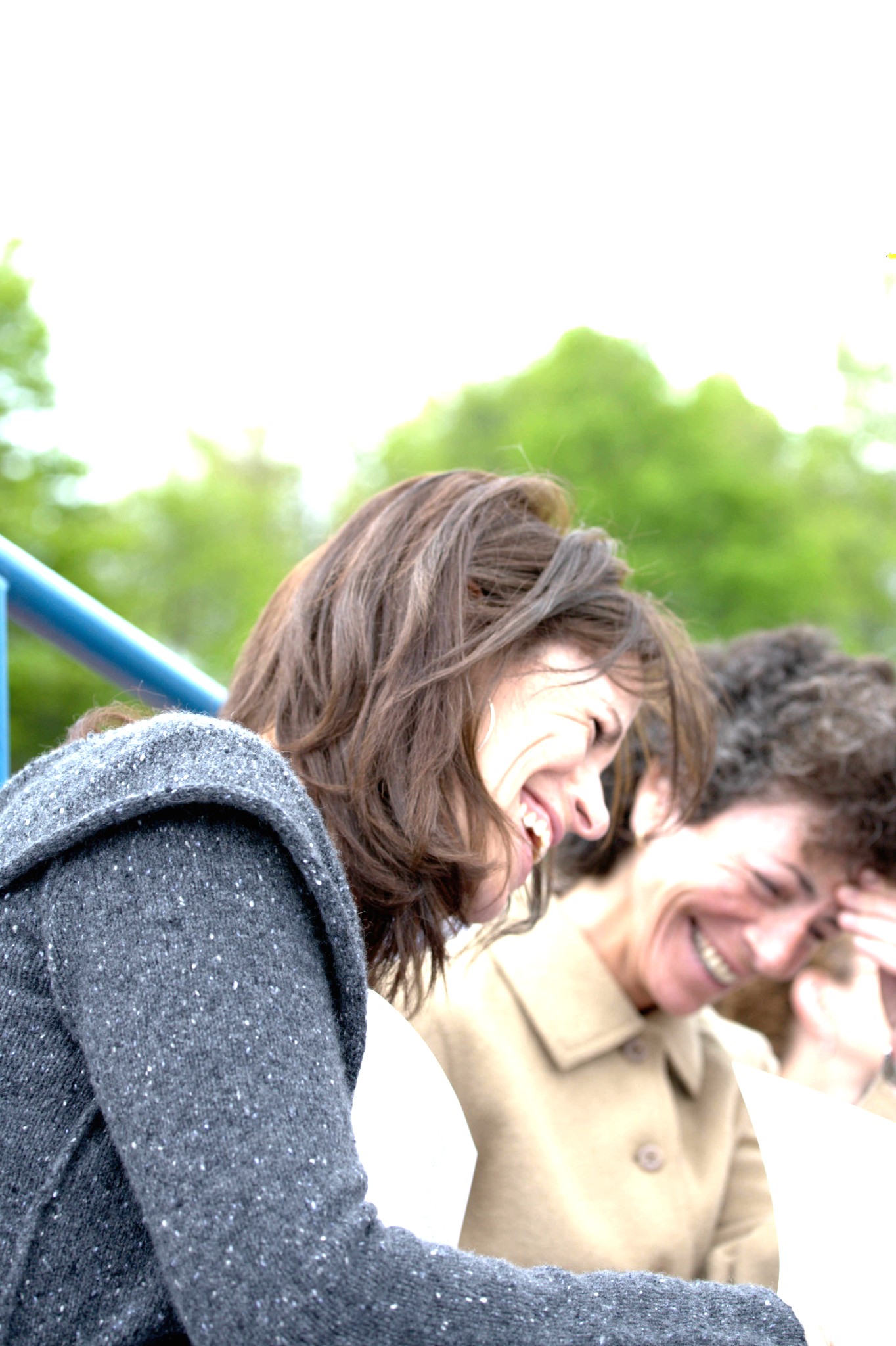} & \includegraphics[width=.1171\textwidth]{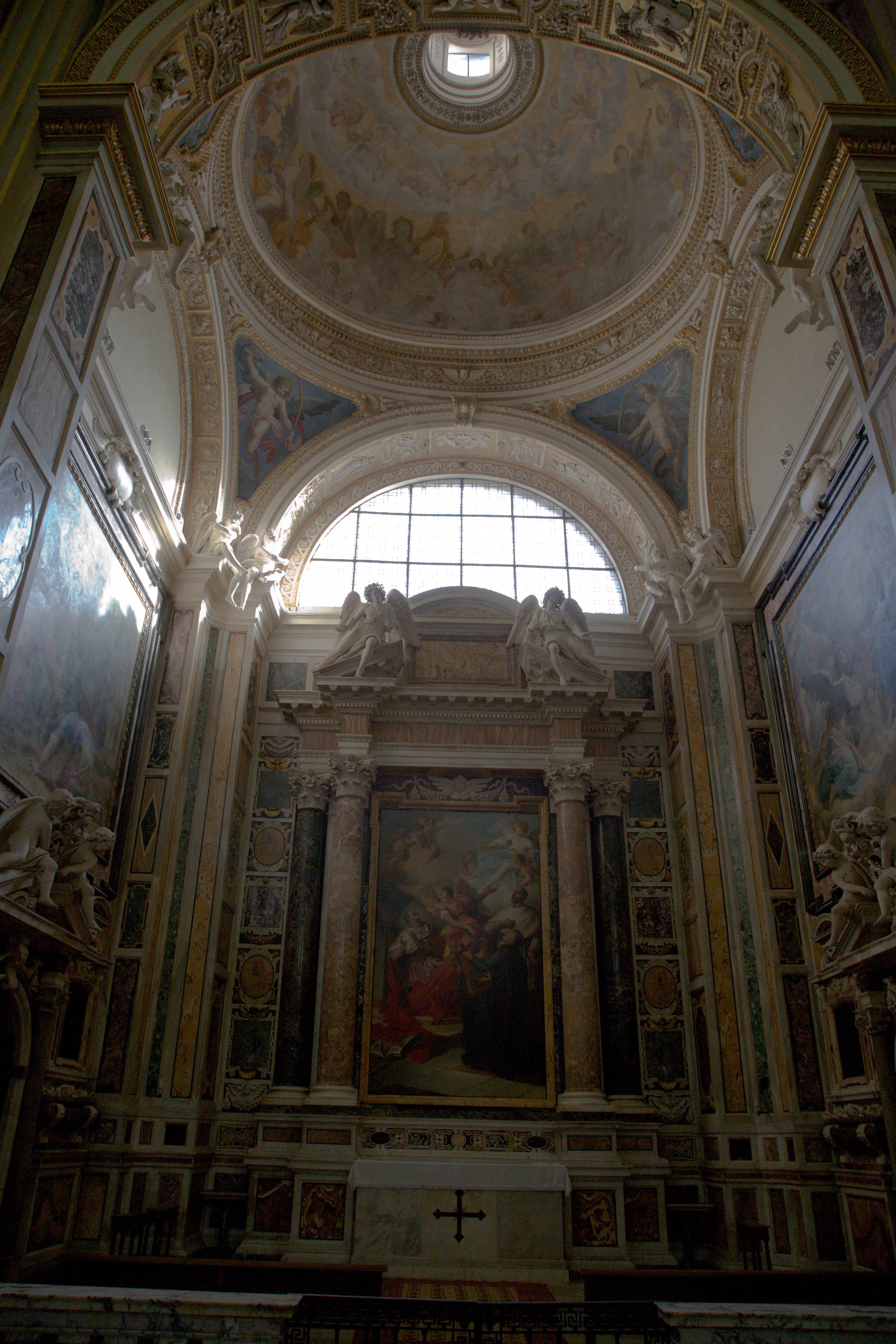} &
\includegraphics[width=.1171\textwidth]{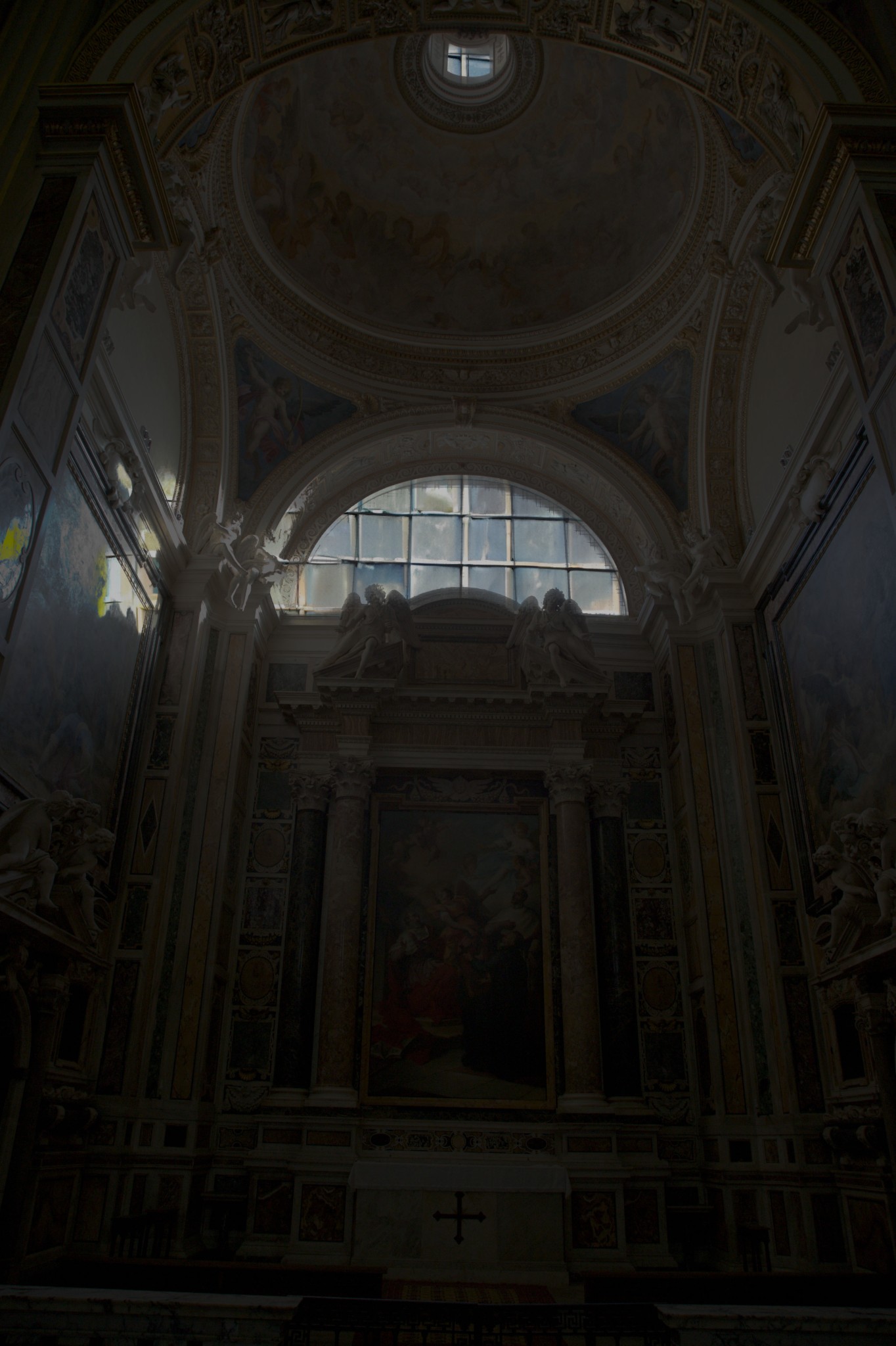} & 
\includegraphics[width=.1171\textwidth]{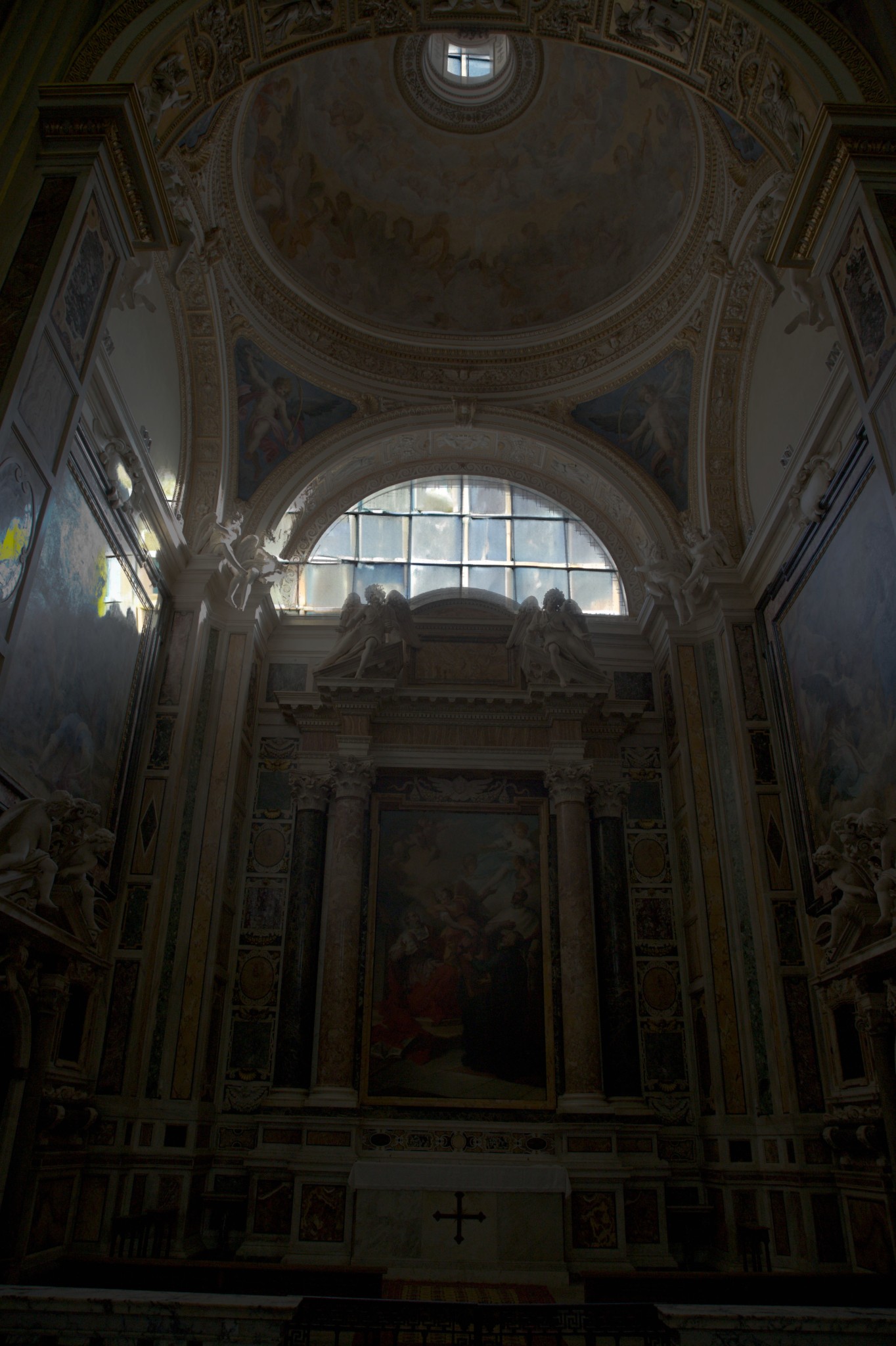} & 
\includegraphics[width=.1171\textwidth]{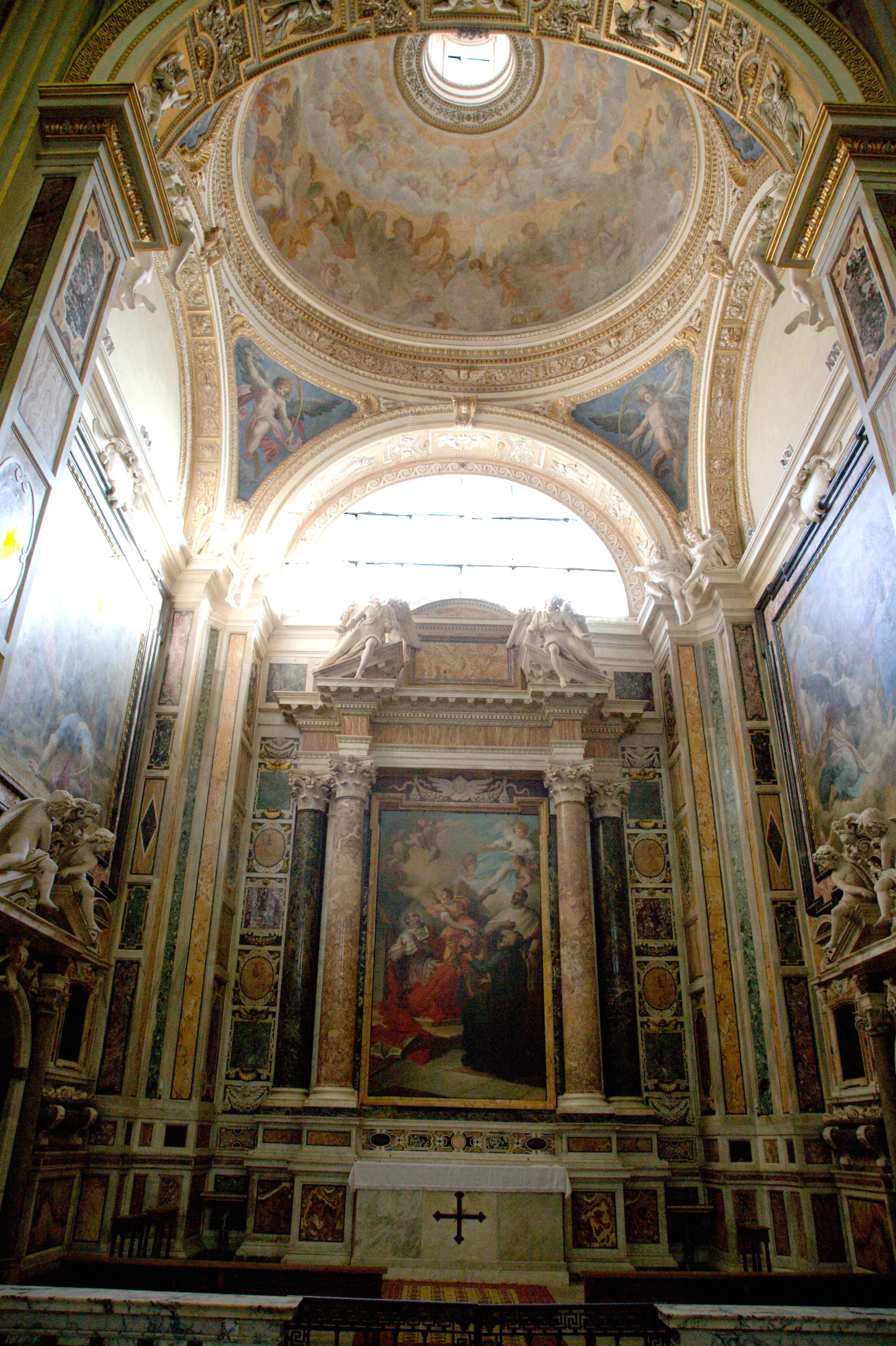}
\end{tabular}
\caption{Results using DITMO with inpainting from the SD model \cite{stableDiffusion} over a range of images.}
\label{fig:ResSD}
\end{figure*}

\begin{figure*}[htbp]
\centering
\setlength{\tabcolsep}{0pt}
\begin{tabular}{cccccccc}
\multicolumn{2}{c}{Input Image} & \multicolumn{2}{c}{-2} & \multicolumn{2}{c}{-1} & \multicolumn{2}{c}{+1} \\
\multicolumn{2}{c}{\includegraphics[width=.234\textwidth]{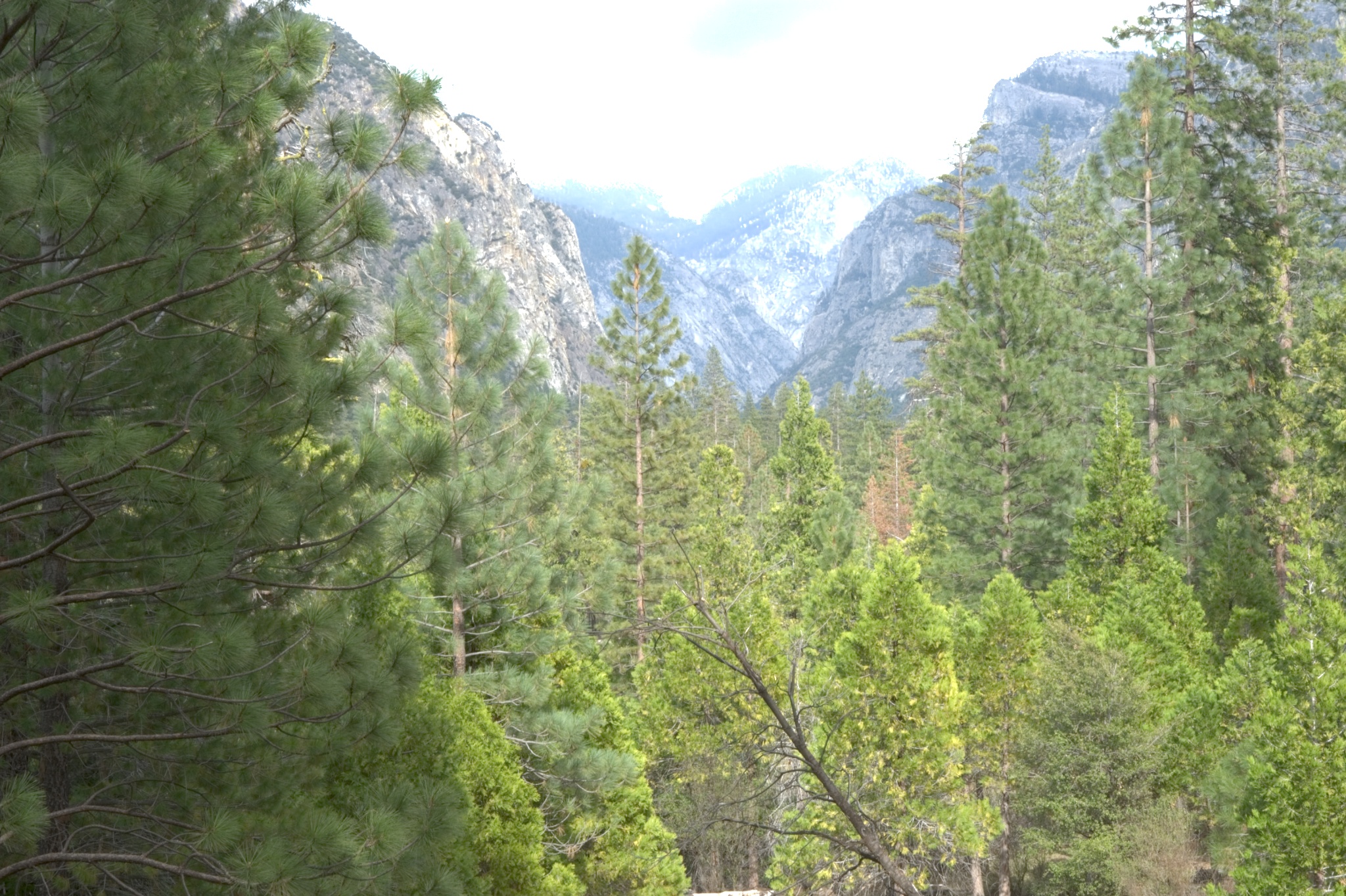}} & 
\multicolumn{2}{c}{\includegraphics[width=.234\textwidth]{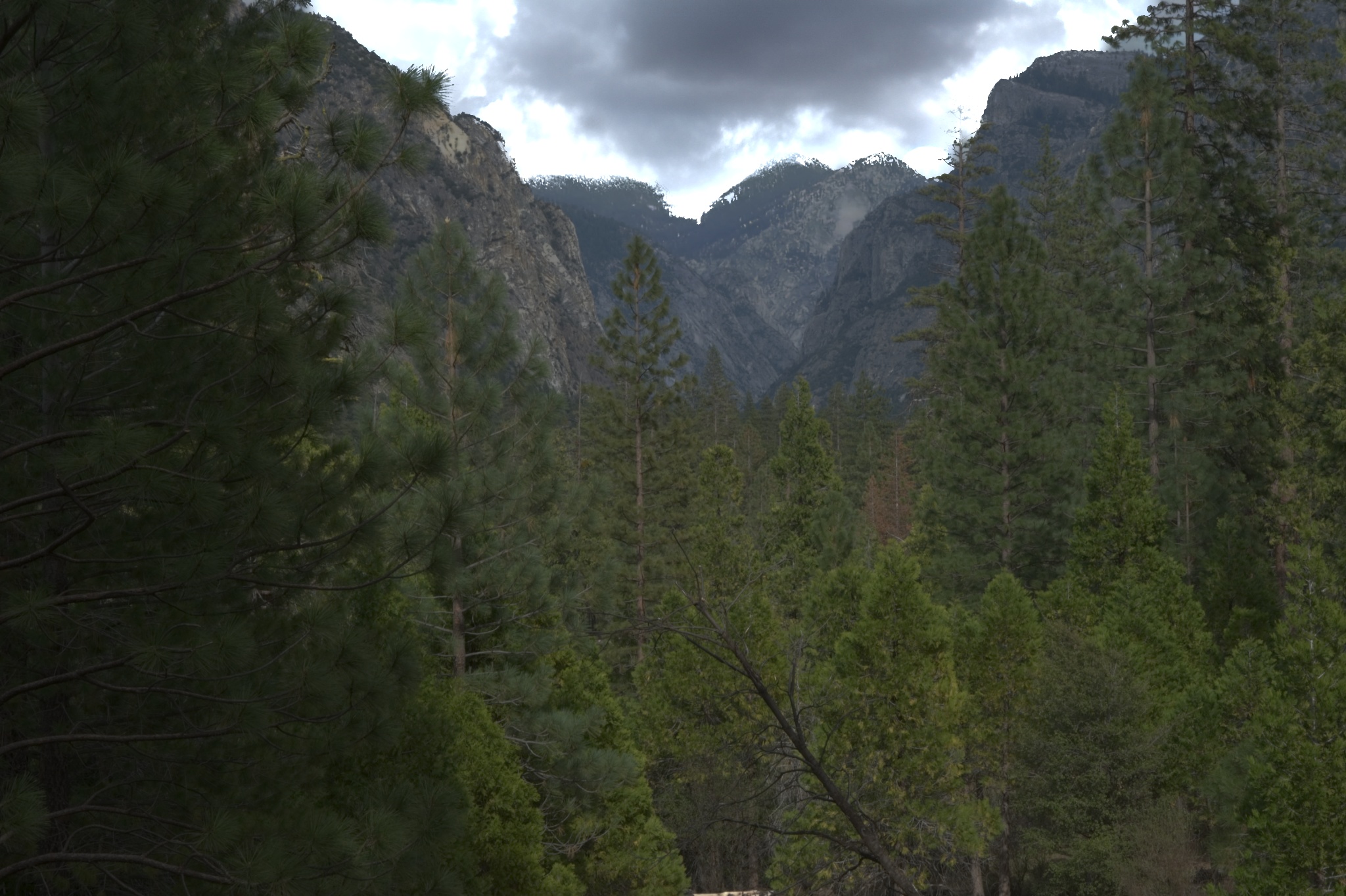}} &
\multicolumn{2}{c}{\includegraphics[width=.234\textwidth]{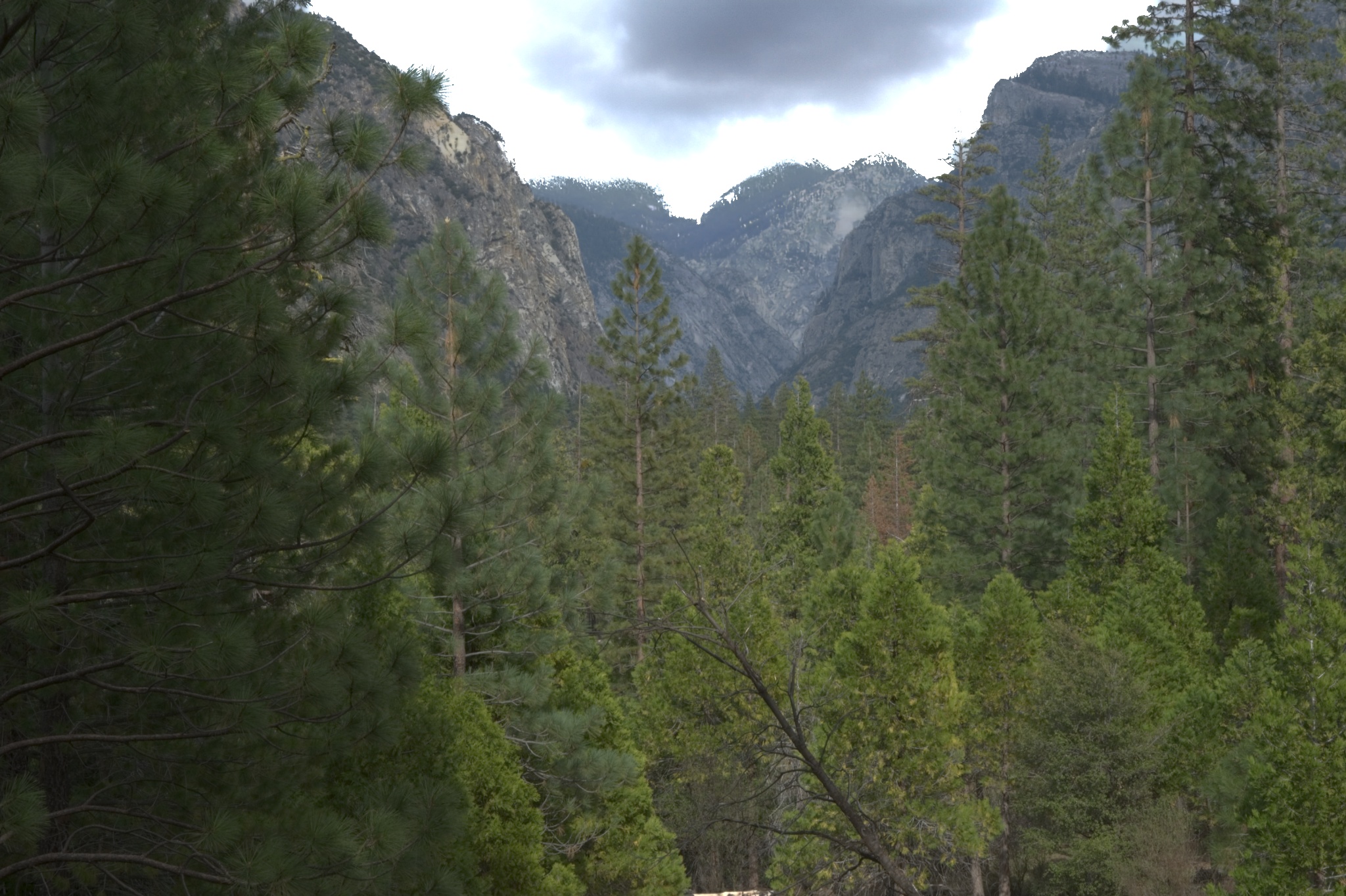}} &
\multicolumn{2}{c}{\includegraphics[width=.234\textwidth]{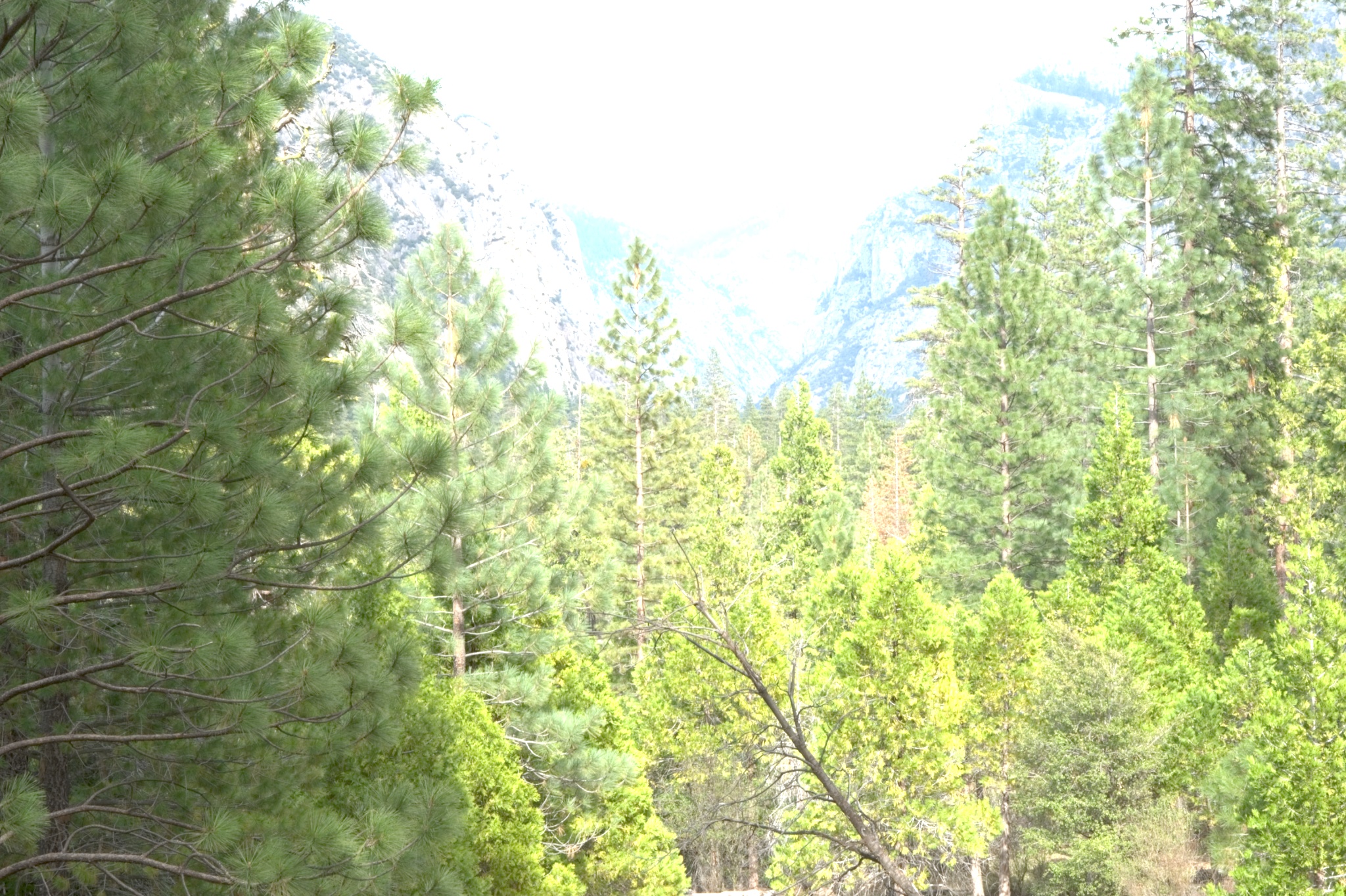}} \\
%         &
\multicolumn{2}{c}{\includegraphics[width=.234\linewidth]{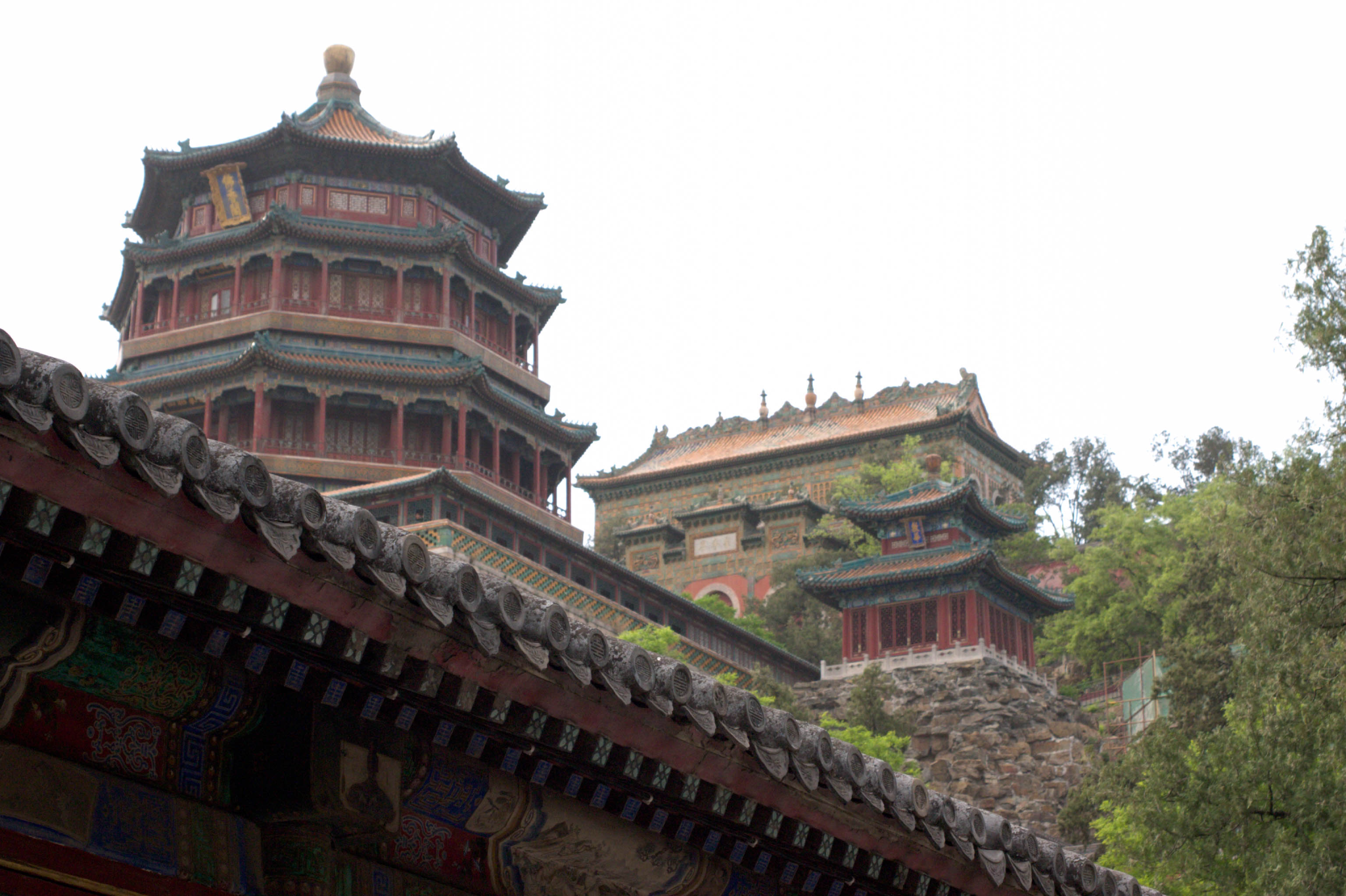}} &
\multicolumn{2}{c}{\includegraphics[width=.234\linewidth]{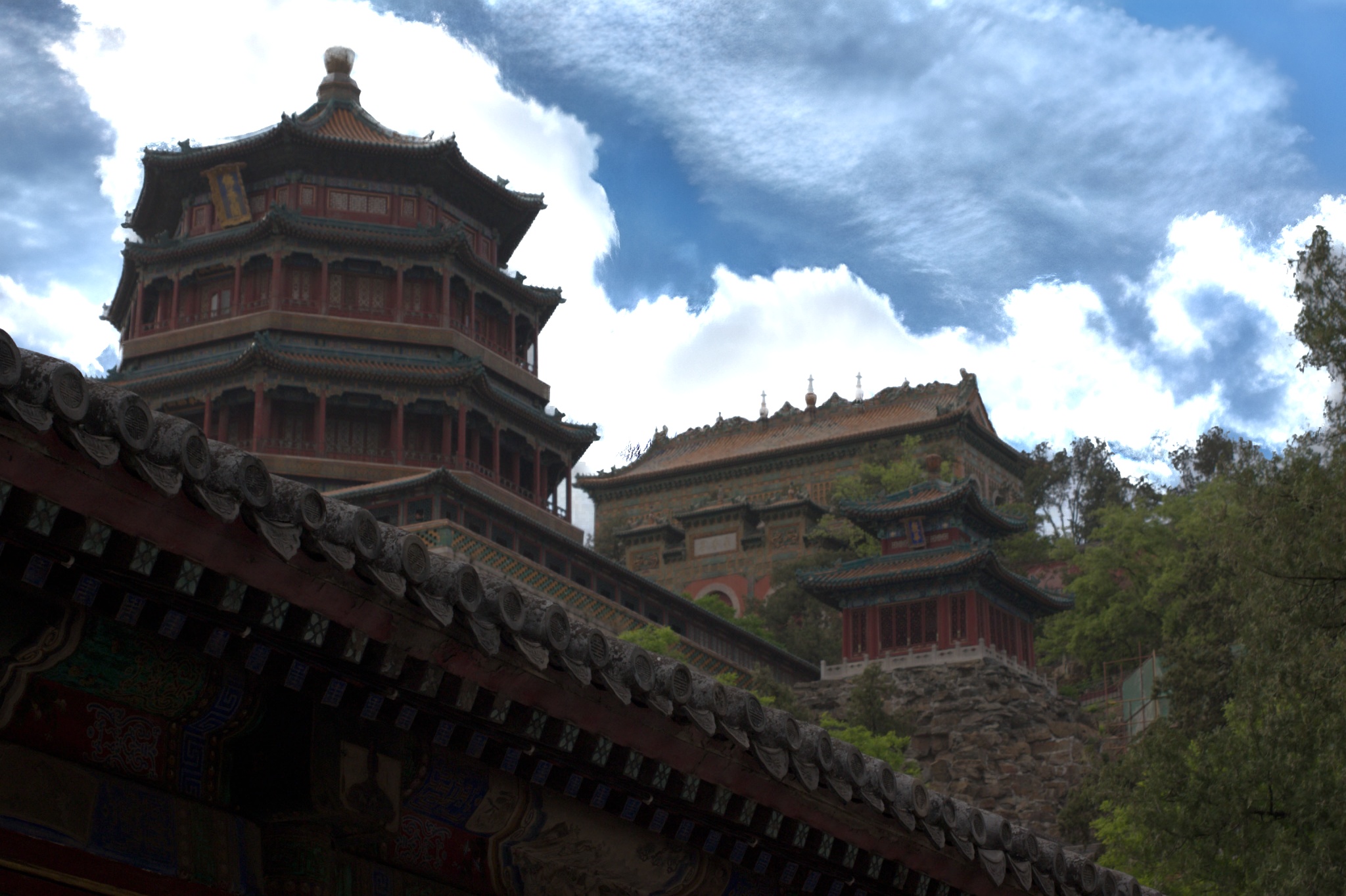}} &
\multicolumn{2}{c}{\includegraphics[width=.234\linewidth]{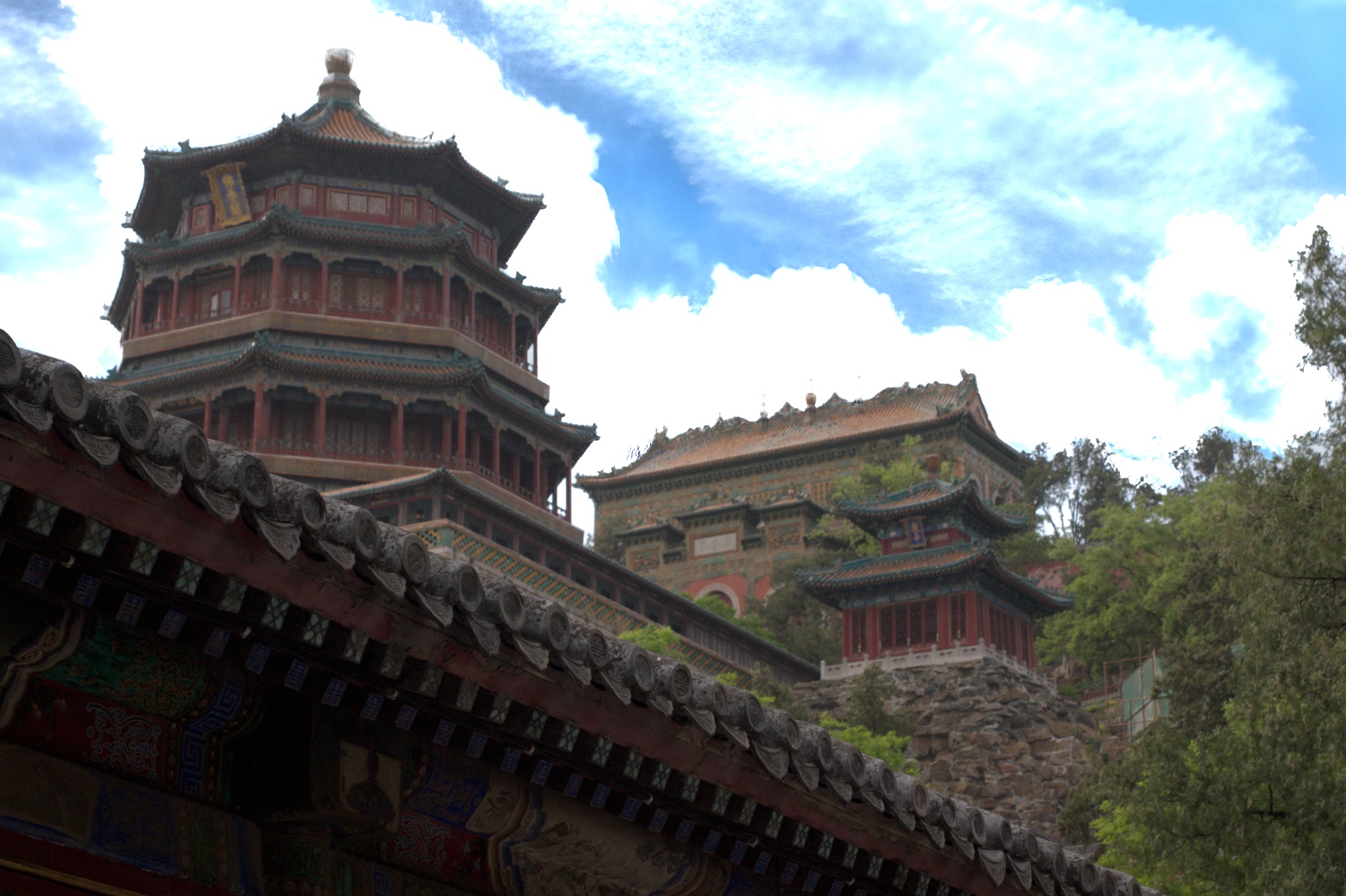}} &
\multicolumn{2}{c}{\includegraphics[width=.234\linewidth]{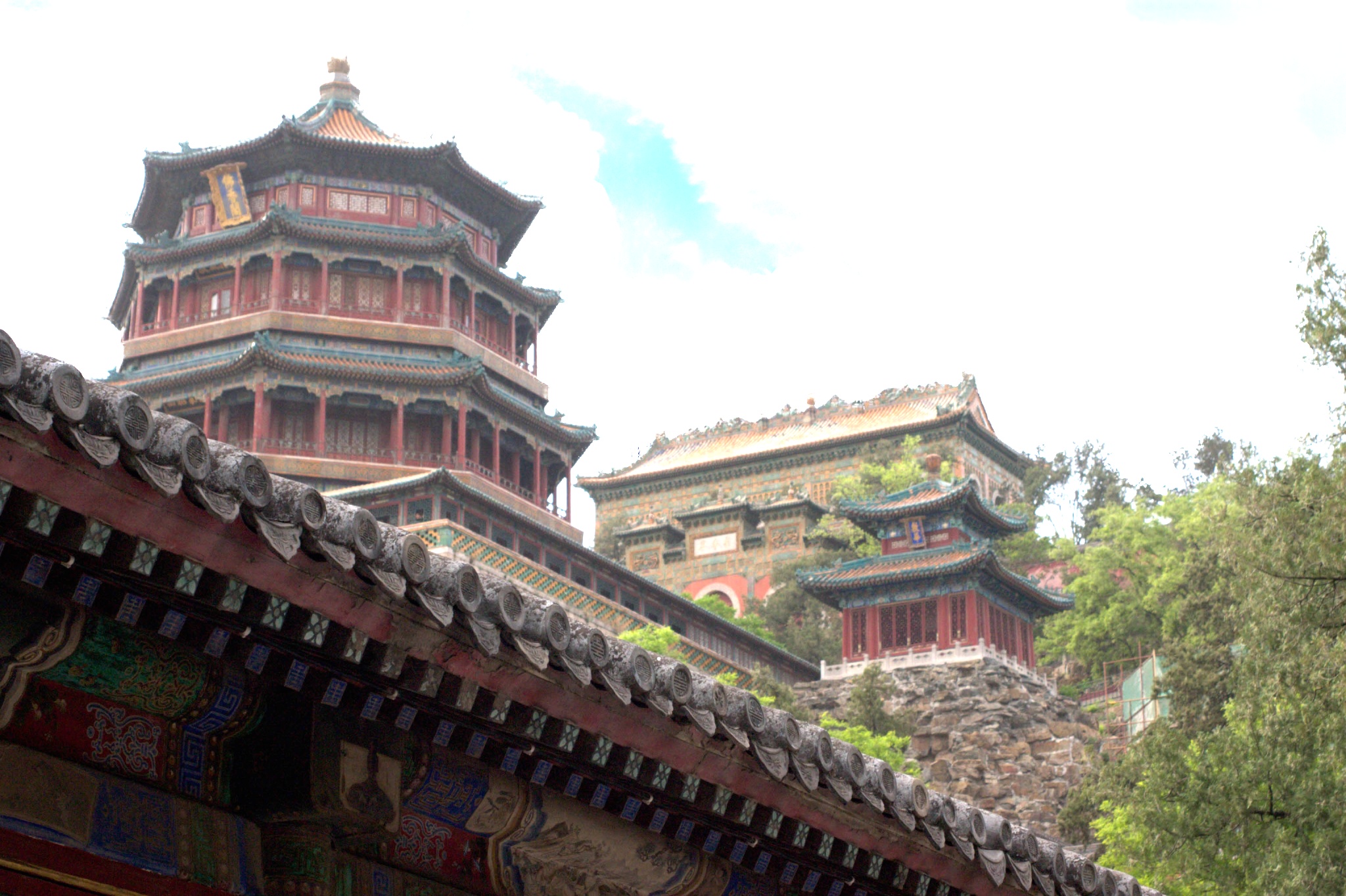}} \\  
Input Image & -2 & -1 & +1 & Input Image & -2 & -1 & +1 \\
\includegraphics[width=.1171\textwidth]{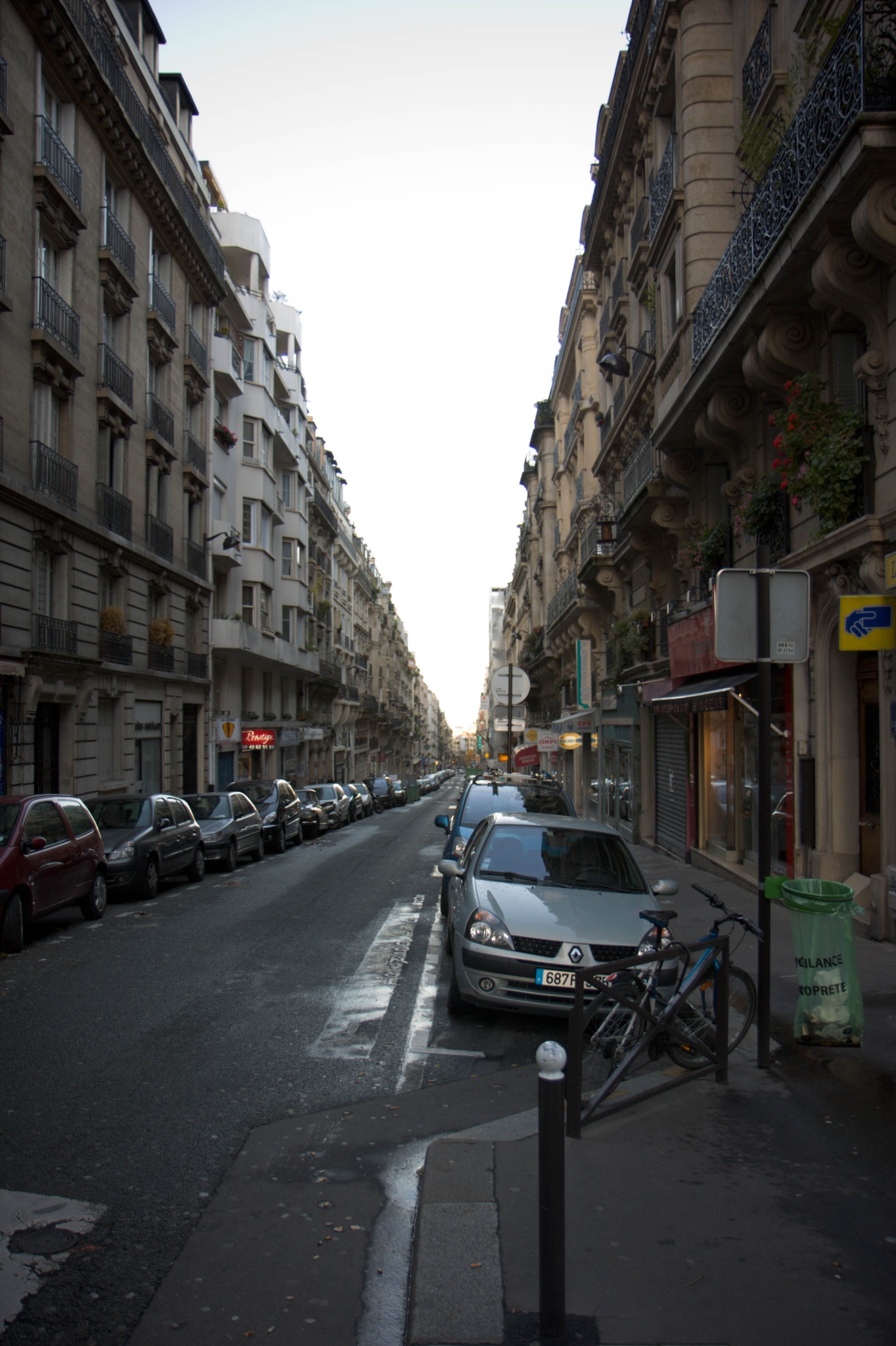} & \includegraphics[width=.1171\textwidth]{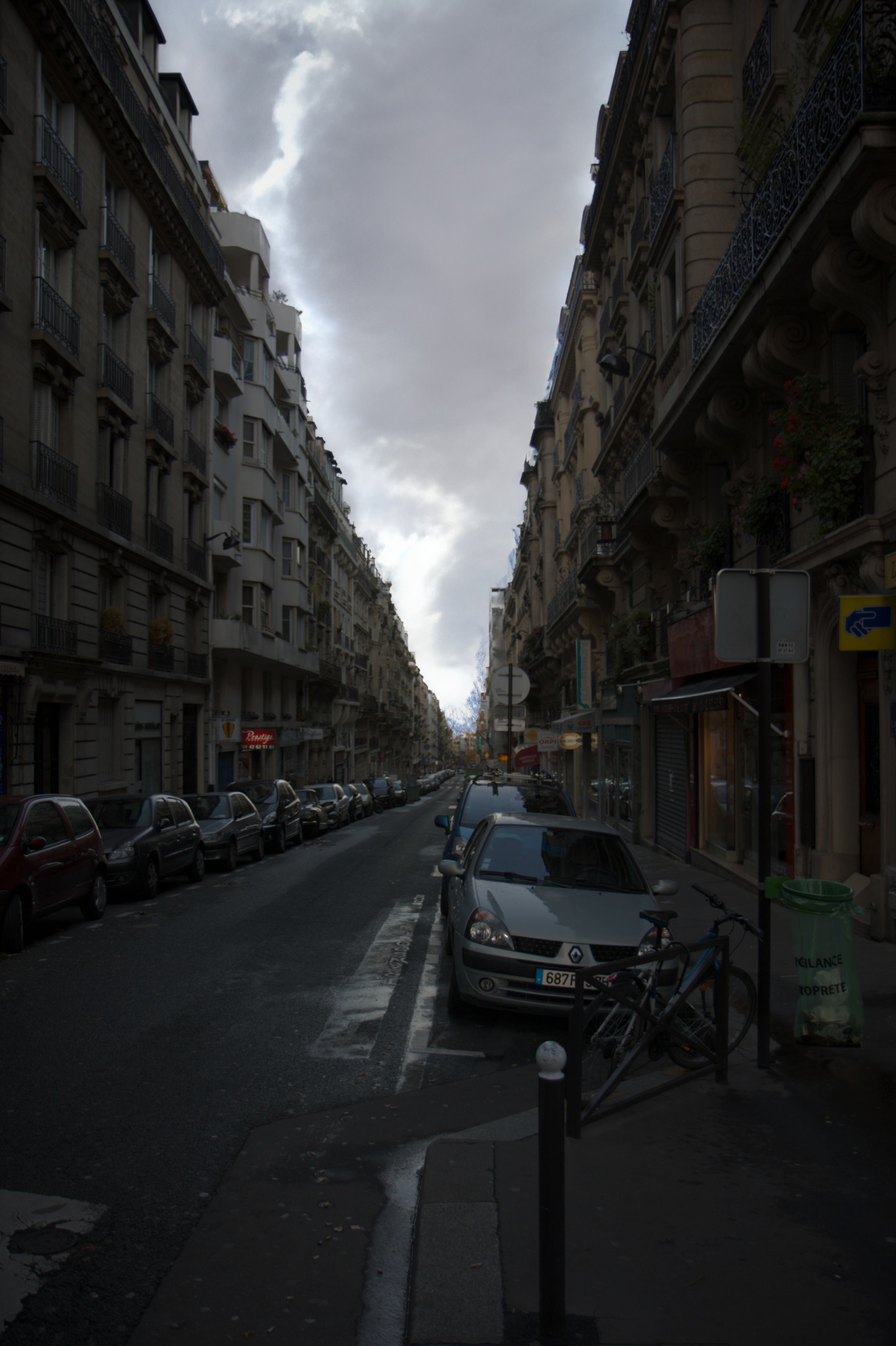} & \includegraphics[width=.1171\textwidth]{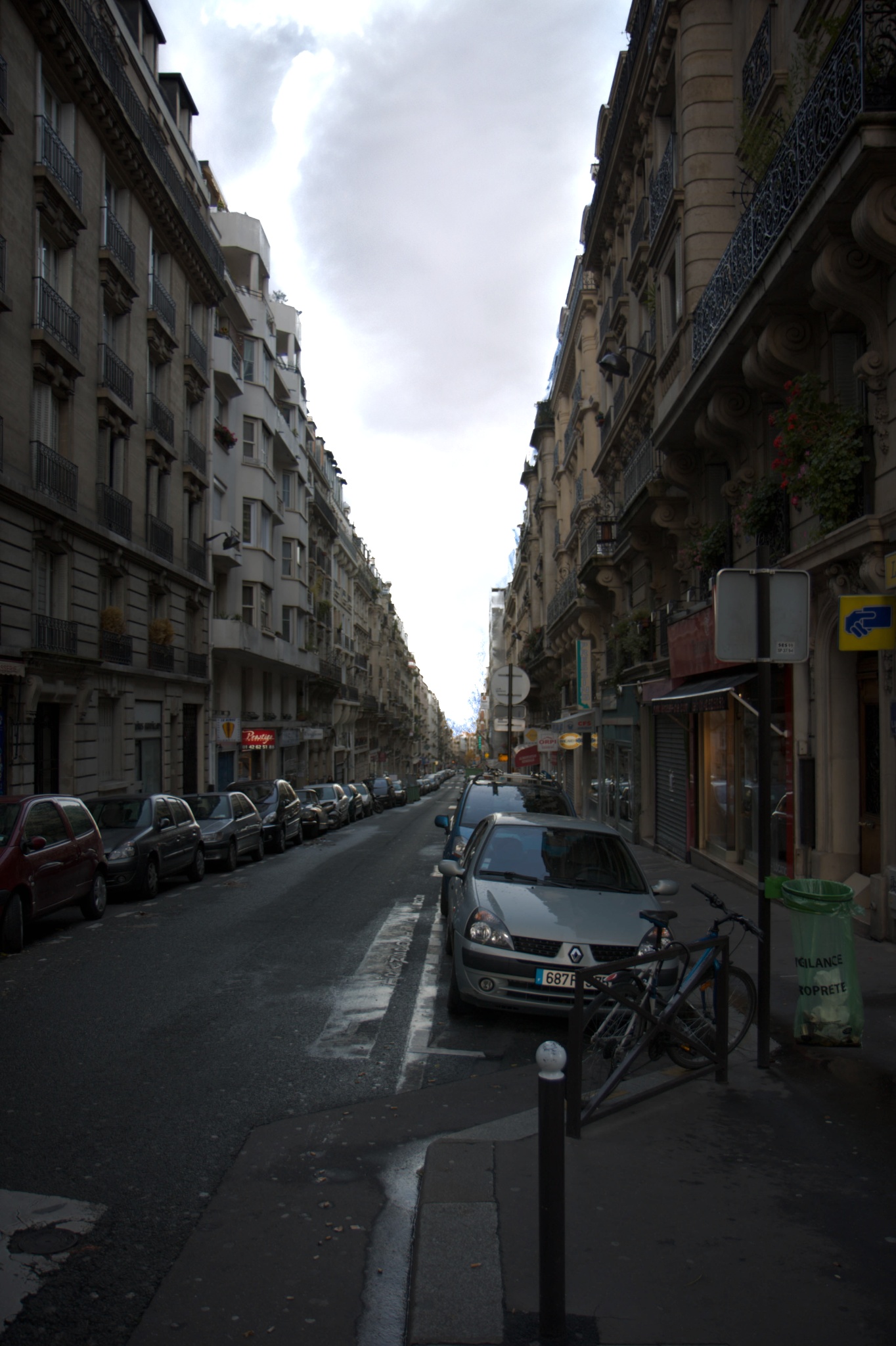} & \includegraphics[width=.1171\textwidth]{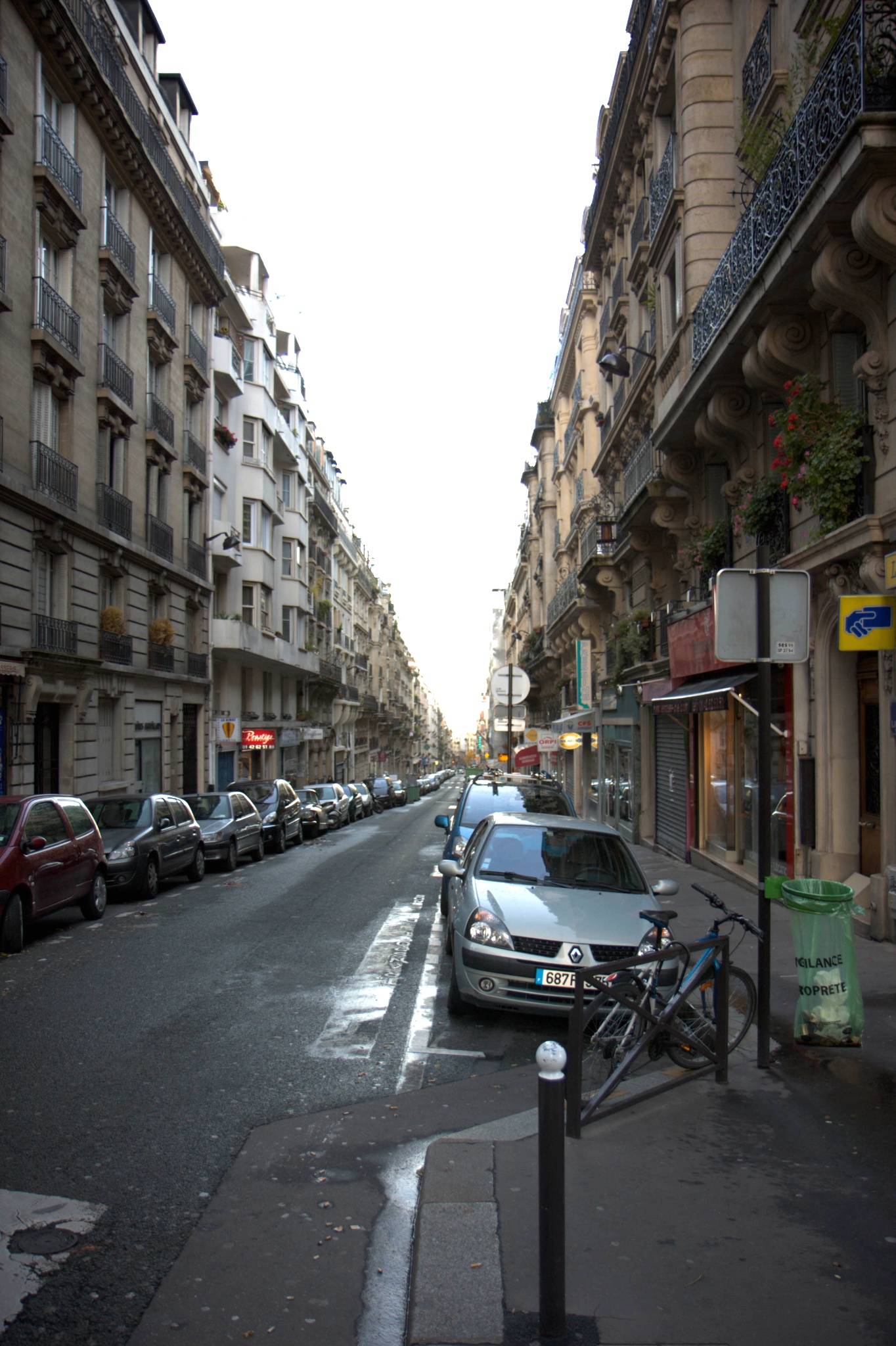} & \includegraphics[width=.1171\textwidth]{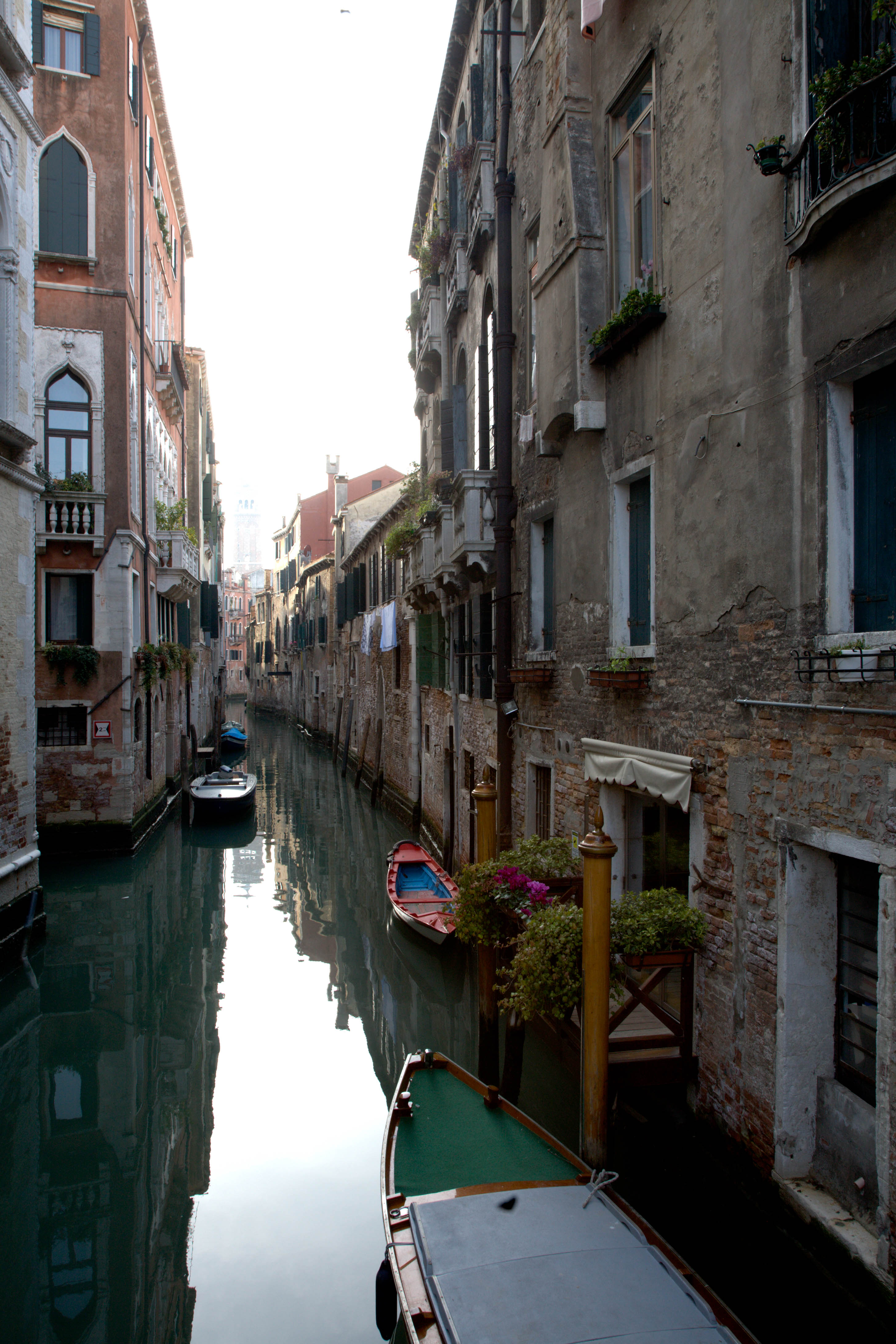} & 
\includegraphics[width=.1171\textwidth]{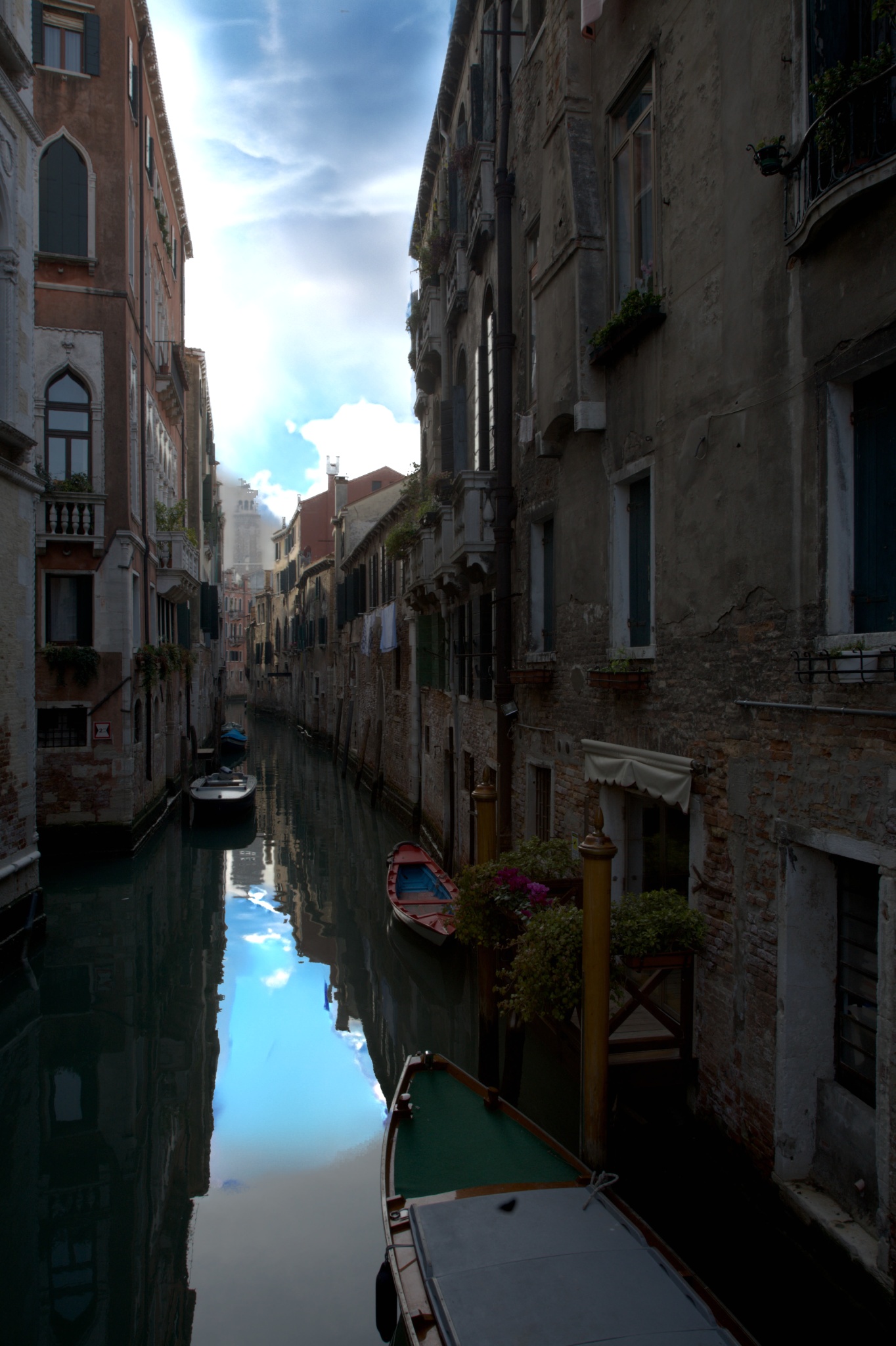} & \includegraphics[width=.1171\textwidth]{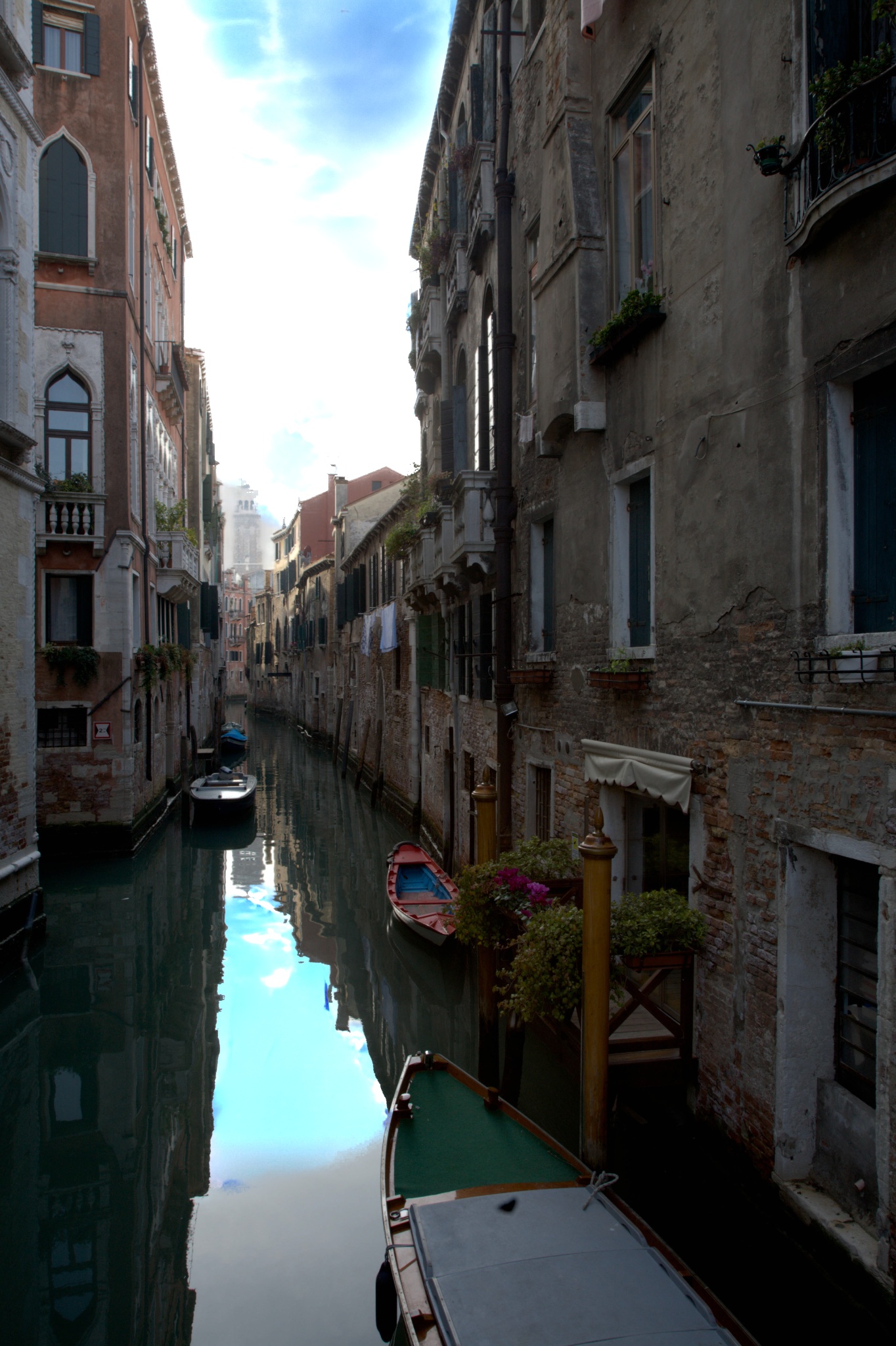} & 
\includegraphics[width=.1171\textwidth]{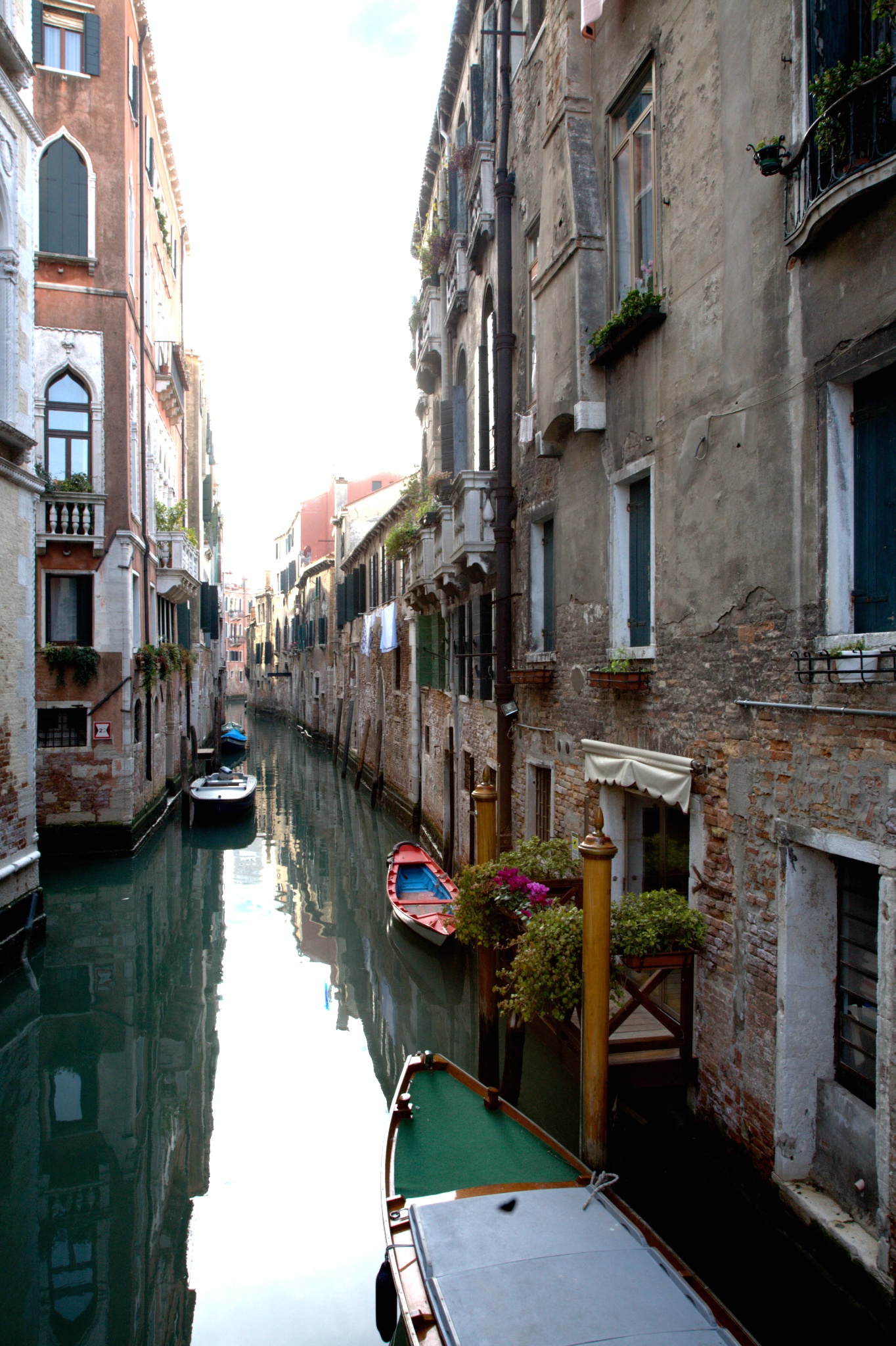}
\end{tabular}
\caption{Results using DITMO with inpainting from the CN model \cite{zhang2023adding} showing details in multiple exposures.}
\label{fig:ResCN}
\end{figure*}

Figures \ref{fig:ResSD} and \ref{fig:ResCN} show further results for DITMO using both \cite{stableDiffusion}, and \cite{zhang2023adding}. The input image is shown first, followed by three exposures. Our method is good at generating missing details in various areas, such as clouds, sky, and interior details in these scenes using both inpainting methods.

\subsection{Limitations}

Our work has some limitations, such as the segmentation model defining the number of classes. If a model can only predict a small number of classes, or the per pixel labels are not correct, then the inpainting results may be less accurate (please see supplementary material for an example). Improved segmentation models can improve this. Furthermore, this relies on a dictionary of prompts, and these need to be expressive enough to generate meaningful content in the clipped regions. 

\section{Conclusion}
In this work, we have introduced a semantic aware diffusion inpainting based ITMO. It is an end-to-end process that takes a single exposure SDR image and hallucinates details in the clipped regions using a diffusion inpainting guided by an ordered semantic graph representing the semantic instances in the image. The main advantages of our work are that we can reconstruct from input images with large over-exposed areas where state-of-the-art methods exhibit severe limitations in such cases and the method produces subjective results that substantially outperform the state-of-the-art based on the outcome of our experiment. For future work, we plan to explore video diffusion \cite{Blattmann+23} to extend the application of our approach to HDR videos, and to investigate improved approaches for synthesizing the prompts used for inpainting. %In doing so, the semantic graph has to be extended to the temporal domain to maintain coherence, which may pose challenges given connectivity may change over frames. %In doing so, there are several challenges especially for long scenes such as a movie for maintaing video coherence.

\clearpage
\newpage
\bibliographystyle{ACM-Reference-Format}
\bibliography{Refs}


\begin{thebibliography}{47}

%%% the Web address.

\ifx \showCODEN    \undefined \def \showCODEN     #1{\unskip}     \fi
\ifx \showDOI      \undefined \def \showDOI       #1{#1}\fi
\ifx \showISBNx    \undefined \def \showISBNx     #1{\unskip}     \fi
\ifx \showISBNxiii \undefined \def \showISBNxiii  #1{\unskip}     \fi
\ifx \showISSN     \undefined \def \showISSN      #1{\unskip}     \fi
\ifx \showLCCN     \undefined \def \showLCCN      #1{\unskip}     \fi
\ifx \shownote     \undefined \def \shownote      #1{#1}          \fi
\ifx \showarticletitle \undefined \def \showarticletitle #1{#1}   \fi
\ifx \showURL      \undefined \def \showURL       {\relax}        \fi
% invisible to TeX
\providecommand\bibfield[2]{#2}
\providecommand\bibinfo[2]{#2}
\providecommand\natexlab[1]{#1}
\providecommand\showeprint[2][]{arXiv:#2}

\bibitem[Aky{\"{u}}z and Reinhard(2007)]%
        {Akyuz+07}
\bibfield{author}{\bibinfo{person}{Ahmet~Oguz Aky{\"{u}}z} {and} \bibinfo{person}{Erik Reinhard}.} \bibinfo{year}{2007}\natexlab{}.
\newblock \showarticletitle{Noise reduction in high dynamic range imaging}.
\newblock \bibinfo{journal}{\emph{J. Vis. Commun. Image Represent.}} \bibinfo{volume}{18}, \bibinfo{number}{5} (\bibinfo{year}{2007}), \bibinfo{pages}{366--376}.
\newblock
\urldef\tempurl%
\url{https://doi.org/10.1016/j.jvcir.2007.04.001}
\showDOI{\tempurl}


\bibitem[Banterle et~al\mbox{.}(2023)]%
        {Banterle+23}
\bibfield{author}{\bibinfo{person}{Francesco Banterle}, \bibinfo{person}{Alessandro Artusi}, \bibinfo{person}{Alejandro Moreo}, \bibinfo{person}{Fabio Carrara}, {and} \bibinfo{person}{Paolo Cignoni}.} \bibinfo{year}{2023}\natexlab{}.
\newblock \showarticletitle{NoR-VDPNet++: Real-Time No-Reference Image Quality Metrics}.
\newblock \bibinfo{journal}{\emph{IEEE Access}}  \bibinfo{volume}{11} (\bibinfo{year}{2023}), \bibinfo{pages}{34544--34553}.
\newblock
\urldef\tempurl%
\url{https://doi.org/10.1109/ACCESS.2023.3263496}
\showDOI{\tempurl}


\bibitem[Banterle et~al\mbox{.}(2006)]%
        {Banterle+06}
\bibfield{author}{\bibinfo{person}{Francesco Banterle}, \bibinfo{person}{Patrick Ledda}, \bibinfo{person}{Kurt Debattista}, {and} \bibinfo{person}{Alan Chalmers}.} \bibinfo{year}{2006}\natexlab{}.
\newblock \showarticletitle{Inverse Tone Mapping}. In \bibinfo{booktitle}{\emph{GRAPHITE '06}} (Kuala Lumpur, Malaysia). \bibinfo{publisher}{ACM}, \bibinfo{address}{New York, NY, USA}, \bibinfo{pages}{349–356}.
\newblock
\showISBNx{1595935649}
\urldef\tempurl%
\url{https://doi.org/10.1145/1174429.1174489}
\showDOI{\tempurl}


\bibitem[Banterle et~al\mbox{.}(2024)]%
        {Banterle+24}
\bibfield{author}{\bibinfo{person}{Francesco Banterle}, \bibinfo{person}{Demetris Marnerides}, \bibinfo{person}{Thomas Bashford-rogers}, {and} \bibinfo{person}{Kurt Debattista}.} \bibinfo{year}{2024}\natexlab{}.
\newblock \showarticletitle{Self-supervised High Dynamic Range Imaging: What Can Be Learned from a Single 8-bit Video?}
\newblock \bibinfo{journal}{\emph{ACM Trans. Graph.}} \bibinfo{volume}{43}, \bibinfo{number}{2}, Article \bibinfo{articleno}{24} (\bibinfo{date}{mar} \bibinfo{year}{2024}), \bibinfo{numpages}{16}~pages.
\newblock
\showISSN{0730-0301}
\urldef\tempurl%
\url{https://doi.org/10.1145/3648570}
\showDOI{\tempurl}


\bibitem[Bist et~al\mbox{.}(2017)]%
        {Bist+17}
\bibfield{author}{\bibinfo{person}{Cambodge Bist}, \bibinfo{person}{R{\'{e}}mi Cozot}, \bibinfo{person}{G{\'{e}}rard Madec}, {and} \bibinfo{person}{Xavier Ducloux}.} \bibinfo{year}{2017}\natexlab{}.
\newblock \showarticletitle{Tone expansion using lighting style aesthetics}.
\newblock \bibinfo{journal}{\emph{Comput. Graph.}}  \bibinfo{volume}{62} (\bibinfo{year}{2017}), \bibinfo{pages}{77--86}.
\newblock
\urldef\tempurl%
\url{https://doi.org/10.1016/j.cag.2016.12.006}
\showDOI{\tempurl}


\bibitem[Blattmann et~al\mbox{.}(2023)]%
        {Blattmann+23}
\bibfield{author}{\bibinfo{person}{Andreas Blattmann}, \bibinfo{person}{Tim Dockhorn}, \bibinfo{person}{Sumith Kulal}, \bibinfo{person}{Daniel Mendelevitch}, \bibinfo{person}{Maciej Kilian}, \bibinfo{person}{Dominik Lorenz}, \bibinfo{person}{Yam Levi}, \bibinfo{person}{Zion English}, \bibinfo{person}{Vikram Voleti}, \bibinfo{person}{Adam Letts}, \bibinfo{person}{Varun Jampani}, {and} \bibinfo{person}{Robin Rombach}.} \bibinfo{year}{2023}\natexlab{}.
\newblock \bibinfo{title}{Stable Video Diffusion: Scaling Latent Video Diffusion Models to Large Datasets}.
\newblock
\newblock
\showeprint[arxiv]{2311.15127}~[cs.CV]


\bibitem[Bychkovsky et~al\mbox{.}(2011)]%
        {Bychkovsky+11}
\bibfield{author}{\bibinfo{person}{Vladimir Bychkovsky}, \bibinfo{person}{Sylvain Paris}, \bibinfo{person}{Eric Chan}, {and} \bibinfo{person}{Fr{\'e}do Durand}.} \bibinfo{year}{2011}\natexlab{}.
\newblock \showarticletitle{Learning Photographic Global Tonal Adjustment with a Database of Input / Output Image Pairs}. In \bibinfo{booktitle}{\emph{The Twenty-Fourth IEEE Conference on Computer Vision and Pattern Recognition}}. \bibinfo{publisher}{IEEE}.
\newblock


\bibitem[Chen et~al\mbox{.}(2022)]%
        {Chen+22}
\bibfield{author}{\bibinfo{person}{Zhaoxi Chen}, \bibinfo{person}{Guangcong Wang}, {and} \bibinfo{person}{Ziwei Liu}.} \bibinfo{year}{2022}\natexlab{}.
\newblock \showarticletitle{Text2Light: Zero-Shot Text-Driven HDR Panorama Generation}.
\newblock \bibinfo{journal}{\emph{ACM Trans. Graph.}} \bibinfo{volume}{41}, \bibinfo{number}{6}, Article \bibinfo{articleno}{195} (\bibinfo{date}{nov} \bibinfo{year}{2022}), \bibinfo{numpages}{16}~pages.
\newblock
\showISSN{0730-0301}
\urldef\tempurl%
\url{https://doi.org/10.1145/3550454.3555447}
\showDOI{\tempurl}


\bibitem[Dalal et~al\mbox{.}(2023)]%
        {dalal2023single}
\bibfield{author}{\bibinfo{person}{Dwip Dalal}, \bibinfo{person}{Gautam Vashishtha}, \bibinfo{person}{Prajwal Singh}, {and} \bibinfo{person}{Shanmuganathan Raman}.} \bibinfo{year}{2023}\natexlab{}.
\newblock \showarticletitle{Single image ldr to hdr conversion using conditional diffusion}. In \bibinfo{booktitle}{\emph{2023 IEEE International Conference on Image Processing (ICIP)}}. IEEE, \bibinfo{pages}{3533--3537}.
\newblock


\bibitem[Debevec and Malik(1997)]%
        {Debevec+97}
\bibfield{author}{\bibinfo{person}{Paul~E. Debevec} {and} \bibinfo{person}{Jitendra Malik}.} \bibinfo{year}{1997}\natexlab{}.
\newblock \showarticletitle{Recovering High Dynamic Range Radiance Maps from Photographs}. In \bibinfo{booktitle}{\emph{Proceedings of the 24th Annual Conference on Computer Graphics and Interactive Techniques}} \emph{(\bibinfo{series}{SIGGRAPH '97})}. \bibinfo{publisher}{ACM Press/Addison-Wesley Publishing Co.}, \bibinfo{address}{USA}, \bibinfo{pages}{369–378}.
\newblock
\showISBNx{0897918967}
\urldef\tempurl%
\url{https://doi.org/10.1145/258734.258884}
\showDOI{\tempurl}


\bibitem[Dhariwal and Nichol(2021)]%
        {dhariwal2021diffusion}
\bibfield{author}{\bibinfo{person}{Prafulla Dhariwal} {and} \bibinfo{person}{Alexander Nichol}.} \bibinfo{year}{2021}\natexlab{}.
\newblock \showarticletitle{Diffusion models beat gans on image synthesis}.
\newblock \bibinfo{journal}{\emph{Advances in neural information processing systems}}  \bibinfo{volume}{34} (\bibinfo{year}{2021}), \bibinfo{pages}{8780--8794}.
\newblock


\bibitem[Didyk et~al\mbox{.}(2008)]%
        {Didyk+08}
\bibfield{author}{\bibinfo{person}{Piotr Didyk}, \bibinfo{person}{Rafa{\l} Mantiuk}, \bibinfo{person}{Matthias Hein}, {and} \bibinfo{person}{Hans-Peter Seidel}.} \bibinfo{year}{2008}\natexlab{}.
\newblock \showarticletitle{Enhancement of Bright Video Features for {HDR} Displays}.
\newblock \bibinfo{journal}{\emph{Computer Graphics Forum}} \bibinfo{volume}{27}, \bibinfo{number}{4} (\bibinfo{year}{2008}), \bibinfo{pages}{1265--1274}.
\newblock


\bibitem[Eilertsen et~al\mbox{.}(2017)]%
        {Eilertsen+17}
\bibfield{author}{\bibinfo{person}{Gabriel Eilertsen}, \bibinfo{person}{Joel Kronander}, \bibinfo{person}{Gyorgy Denes}, \bibinfo{person}{Rafa\l~K. Mantiuk}, {and} \bibinfo{person}{Jonas Unger}.} \bibinfo{year}{2017}\natexlab{}.
\newblock \showarticletitle{{HDR} image reconstruction from a single exposure using deep CNNs}.
\newblock \bibinfo{journal}{\emph{{ACM} Trans. Graph.}} \bibinfo{volume}{36}, \bibinfo{number}{6} (\bibinfo{year}{2017}), \bibinfo{pages}{178:1--178:15}.
\newblock
\urldef\tempurl%
\url{https://doi.org/10.1145/3130800.3130816}
\showDOI{\tempurl}


\bibitem[Elharrouss et~al\mbox{.}(2020)]%
        {ImageInpainting}
\bibfield{author}{\bibinfo{person}{Omar Elharrouss}, \bibinfo{person}{Noor Almaadeed}, \bibinfo{person}{Somaya Al-Maadeed}, {and} \bibinfo{person}{Younes Akbari}.} \bibinfo{year}{2020}\natexlab{}.
\newblock \showarticletitle{Image inpainting: A review}.
\newblock \bibinfo{journal}{\emph{Neural Processing Letters}}  \bibinfo{volume}{51} (\bibinfo{year}{2020}), \bibinfo{pages}{2007--2028}.
\newblock


\bibitem[Endo et~al\mbox{.}(2017)]%
        {Endo+17}
\bibfield{author}{\bibinfo{person}{Yuki Endo}, \bibinfo{person}{Yoshihiro Kanamori}, {and} \bibinfo{person}{Jun Mitani}.} \bibinfo{year}{2017}\natexlab{}.
\newblock \showarticletitle{Deep Reverse Tone Mapping}.
\newblock \bibinfo{journal}{\emph{ACM Trans. Graph.}} \bibinfo{volume}{36}, \bibinfo{number}{6}, Article \bibinfo{articleno}{177} (\bibinfo{year}{2017}), \bibinfo{numpages}{10}~pages.
\newblock
\showISSN{0730-0301}
\urldef\tempurl%
\url{https://doi.org/10.1145/3130800.3130834}
\showDOI{\tempurl}


\bibitem[Fairchild(2007)]%
        {Fairchild+07}
\bibfield{author}{\bibinfo{person}{Mark~D Fairchild}.} \bibinfo{year}{2007}\natexlab{}.
\newblock \showarticletitle{The HDR photographic survey}. In \bibinfo{booktitle}{\emph{Color and imaging conference}}, Vol.~\bibinfo{volume}{15}. Society of Imaging Science and Technology, \bibinfo{pages}{233--238}.
\newblock


\bibitem[Germer et~al\mbox{.}(2020)]%
        {Germer+20}
\bibfield{author}{\bibinfo{person}{Thomas Germer}, \bibinfo{person}{Tobias Uelwer}, \bibinfo{person}{Stefan Conrad}, {and} \bibinfo{person}{Stefan Harmeling}.} \bibinfo{year}{2020}\natexlab{}.
\newblock \showarticletitle{{PyMatting: A Python Library for Alpha Matting}}.
\newblock \bibinfo{journal}{\emph{Journal of Open Source Software}} \bibinfo{volume}{5}, \bibinfo{number}{54} (\bibinfo{year}{2020}), \bibinfo{pages}{2481}.
\newblock
\urldef\tempurl%
\url{https://doi.org/10.21105/joss.02481}
\showDOI{\tempurl}


\bibitem[Goswami et~al\mbox{.}(2022)]%
        {goswami+22}
\bibfield{author}{\bibinfo{person}{Abhishek Goswami}, \bibinfo{person}{Erwan Bernard}, \bibinfo{person}{Wolf Hauser}, \bibinfo{person}{Frederic Dufaux}, {and} \bibinfo{person}{Rafal Mantiuk}.} \bibinfo{year}{2022}\natexlab{}.
\newblock \showarticletitle{{G-SemTMO: Tone Mapping with a Trainable Semantic Graph}}.
\newblock \bibinfo{journal}{\emph{arXiv preprint arXiv:2208.14113}} (\bibinfo{year}{2022}).
\newblock


\bibitem[Goswami et~al\mbox{.}(2020)]%
        {goswami+20}
\bibfield{author}{\bibinfo{person}{Abhishek Goswami}, \bibinfo{person}{Mathis Petrovich}, \bibinfo{person}{Wolf Hauser}, {and} \bibinfo{person}{Fr{\'e}d{\'e}ric Dufaux}.} \bibinfo{year}{2020}\natexlab{}.
\newblock \showarticletitle{{Tone mapping operators: Progressing towards semantic-awareness}}. In \bibinfo{booktitle}{\emph{2020 IEEE International Conference on Multimedia \& Expo Workshops (ICMEW)}}. IEEE, \bibinfo{pages}{1--6}.
\newblock


\bibitem[Hanji et~al\mbox{.}(2022)]%
        {Hanji+22}
\bibfield{author}{\bibinfo{person}{Param Hanji}, \bibinfo{person}{Rafal Mantiuk}, \bibinfo{person}{Gabriel Eilertsen}, \bibinfo{person}{Saghi Hajisharif}, {and} \bibinfo{person}{Jonas Unger}.} \bibinfo{year}{2022}\natexlab{}.
\newblock \showarticletitle{Comparison of Single Image HDR Reconstruction Methods — the Caveats of Quality Assessment}. In \bibinfo{booktitle}{\emph{ACM SIGGRAPH 2022 Conference Proceedings}} (Vancouver, BC, Canada) \emph{(\bibinfo{series}{SIGGRAPH '22})}. \bibinfo{publisher}{Association for Computing Machinery}, \bibinfo{address}{New York, NY, USA}, Article \bibinfo{articleno}{1}, \bibinfo{numpages}{8}~pages.
\newblock
\showISBNx{9781450393379}
\urldef\tempurl%
\url{https://doi.org/10.1145/3528233.3530729}
\showDOI{\tempurl}


\bibitem[Hasinoff et~al\mbox{.}(2016)]%
        {Hasinoff+16}
\bibfield{author}{\bibinfo{person}{Samuel~W. Hasinoff}, \bibinfo{person}{Dillon Sharlet}, \bibinfo{person}{Ryan Geiss}, \bibinfo{person}{Andrew Adams}, \bibinfo{person}{Jonathan~T. Barron}, \bibinfo{person}{Florian Kainz}, \bibinfo{person}{Jiawen Chen}, {and} \bibinfo{person}{Marc Levoy}.} \bibinfo{year}{2016}\natexlab{}.
\newblock \showarticletitle{Burst Photography for High Dynamic Range and Low-Light Imaging on Mobile Cameras}.
\newblock \bibinfo{journal}{\emph{ACM Trans. Graph.}} \bibinfo{volume}{35}, \bibinfo{number}{6}, Article \bibinfo{articleno}{192} (\bibinfo{date}{nov} \bibinfo{year}{2016}), \bibinfo{numpages}{12}~pages.
\newblock
\showISSN{0730-0301}
\urldef\tempurl%
\url{https://doi.org/10.1145/2980179.2980254}
\showDOI{\tempurl}


\bibitem[Jo et~al\mbox{.}(2022)]%
        {Jo+22}
\bibfield{author}{\bibinfo{person}{So~Yeon Jo}, \bibinfo{person}{Siyeong Lee}, \bibinfo{person}{Namhyun Ahn}, {and} \bibinfo{person}{Suk{-}Ju Kang}.} \bibinfo{year}{2022}\natexlab{}.
\newblock \showarticletitle{Deep Arbitrary {HDRI:} Inverse Tone Mapping With Controllable Exposure Changes}.
\newblock \bibinfo{journal}{\emph{{IEEE} Trans. Multim.}}  \bibinfo{volume}{24} (\bibinfo{year}{2022}), \bibinfo{pages}{2713--2726}.
\newblock
\urldef\tempurl%
\url{https://doi.org/10.1109/TMM.2021.3087034}
\showDOI{\tempurl}


\bibitem[Kirillov et~al\mbox{.}(2023)]%
        {kirillov+23}
\bibfield{author}{\bibinfo{person}{Alexander Kirillov}, \bibinfo{person}{Eric Mintun}, \bibinfo{person}{Nikhila Ravi}, \bibinfo{person}{Hanzi Mao}, \bibinfo{person}{Chloe Rolland}, \bibinfo{person}{Laura Gustafson}, \bibinfo{person}{Tete Xiao}, \bibinfo{person}{Spencer Whitehead}, \bibinfo{person}{Alexander~C Berg}, \bibinfo{person}{Wan-Yen Lo}, {et~al\mbox{.}}} \bibinfo{year}{2023}\natexlab{}.
\newblock \showarticletitle{{Segment Anything}}.
\newblock \bibinfo{journal}{\emph{arXiv preprint arXiv:2304.02643}} (\bibinfo{year}{2023}).
\newblock


\bibitem[Kovaleski and Oliveira(2014)]%
        {Kovaleski+14}
\bibfield{author}{\bibinfo{person}{Rafael~Pacheco Kovaleski} {and} \bibinfo{person}{Manuel~M. Oliveira}.} \bibinfo{year}{2014}\natexlab{}.
\newblock \showarticletitle{High-Quality Reverse Tone Mapping for a Wide Range of Exposures}. In \bibinfo{booktitle}{\emph{27th SIBGRAPI Conference on Graphics, Patterns and Images}}. \bibinfo{publisher}{IEEE Computer Society}, \bibinfo{address}{New York}, \bibinfo{pages}{49--56}.
\newblock


\bibitem[Lee et~al\mbox{.}(2018)]%
        {Lee+18}
\bibfield{author}{\bibinfo{person}{Siyeong Lee}, \bibinfo{person}{Gwon~Hwan An}, {and} \bibinfo{person}{Suk-Ju Kang}.} \bibinfo{year}{2018}\natexlab{}.
\newblock \showarticletitle{Deep recursive hdri: Inverse tone mapping using generative adversarial networks}. In \bibinfo{booktitle}{\emph{proceedings of the European Conference on Computer Vision (ECCV)}}. \bibinfo{pages}{596--611}.
\newblock


\bibitem[Liu et~al\mbox{.}(2020)]%
        {Liu+20}
\bibfield{author}{\bibinfo{person}{Yu-Lun Liu}, \bibinfo{person}{Wei-Sheng Lai}, \bibinfo{person}{Yu-Sheng Chen}, \bibinfo{person}{Yi-Lung Kao}, \bibinfo{person}{Ming-Hsuan Yang}, \bibinfo{person}{Yung-Yu Chuang}, {and} \bibinfo{person}{Jia-Bin Huang}.} \bibinfo{year}{2020}\natexlab{}.
\newblock \showarticletitle{Single-Image HDR Reconstruction by Learning to Reverse the Camera Pipeline}. In \bibinfo{booktitle}{\emph{Proceedings of the IEEE/CVF Conference on Computer Vision and Pattern Recognition (CVPR)}}. \bibinfo{publisher}{IEEE}, \bibinfo{pages}{1648--1657}.
\newblock


\bibitem[Mantiuk et~al\mbox{.}(2023)]%
        {Mantiuk+23}
\bibfield{author}{\bibinfo{person}{Rafal~K. Mantiuk}, \bibinfo{person}{Dounia Hammou}, {and} \bibinfo{person}{Param Hanji}.} \bibinfo{year}{2023}\natexlab{}.
\newblock \bibinfo{title}{HDR-VDP-3: A multi-metric for predicting image differences, quality and contrast distortions in high dynamic range and regular content}.
\newblock
\newblock
\showeprint[arxiv]{2304.13625}~[eess.IV]


\bibitem[Marnerides et~al\mbox{.}(2018)]%
        {Marnerides+18}
\bibfield{author}{\bibinfo{person}{Demetris Marnerides}, \bibinfo{person}{Thomas Bashford{-}Rogers}, \bibinfo{person}{Jonathan Hatchett}, {and} \bibinfo{person}{Kurt Debattista}.} \bibinfo{year}{2018}\natexlab{}.
\newblock \showarticletitle{ExpandNet: {A} Deep Convolutional Neural Network for High Dynamic Range Expansion from Low Dynamic Range Content}.
\newblock \bibinfo{journal}{\emph{Comput. Graph. Forum}} \bibinfo{volume}{37}, \bibinfo{number}{2} (\bibinfo{year}{2018}), \bibinfo{pages}{37--49}.
\newblock
\urldef\tempurl%
\url{https://doi.org/10.1111/cgf.13340}
\showDOI{\tempurl}


\bibitem[Masia et~al\mbox{.}(2009)]%
        {Masia+09}
\bibfield{author}{\bibinfo{person}{Belen Masia}, \bibinfo{person}{Sandra Agustin}, \bibinfo{person}{Roland~W. Fleming}, \bibinfo{person}{Olga Sorkine}, {and} \bibinfo{person}{Diego Gutierrez}.} \bibinfo{year}{2009}\natexlab{}.
\newblock \showarticletitle{Evaluation of reverse tone mapping through varying exposure conditions}. In \bibinfo{booktitle}{\emph{ACM SIGGRAPH Asia 2009 Papers}} (Yokohama, Japan) \emph{(\bibinfo{series}{SIGGRAPH Asia '09})}. \bibinfo{publisher}{Association for Computing Machinery}, \bibinfo{address}{New York, NY, USA}, Article \bibinfo{articleno}{160}, \bibinfo{numpages}{8}~pages.
\newblock
\showISBNx{9781605588582}
\urldef\tempurl%
\url{https://doi.org/10.1145/1661412.1618506}
\showDOI{\tempurl}


\bibitem[Meylan et~al\mbox{.}(2006)]%
        {Meylan+06}
\bibfield{author}{\bibinfo{person}{Laurence Meylan}, \bibinfo{person}{Scott~J. Daly}, {and} \bibinfo{person}{Sabine S{\"{u}}sstrunk}.} \bibinfo{year}{2006}\natexlab{}.
\newblock \showarticletitle{The Reproduction of Specular Highlights on High Dynamic Range Displays}. In \bibinfo{booktitle}{\emph{14th Color and Imaging Conference, {CIC} 2006, Scottsdale, Arizona, USA, November 6-10, 2006}}. \bibinfo{publisher}{Society for Imaging Science and Technology}, \bibinfo{pages}{333--338}.
\newblock
\urldef\tempurl%
\url{https://doi.org/10.2352/CIC.2006.14.1.ART00061}
\showDOI{\tempurl}


\bibitem[Rempel et~al\mbox{.}(2007)]%
        {Rempel+07}
\bibfield{author}{\bibinfo{person}{Allan~G. Rempel}, \bibinfo{person}{Matthew Trentacoste}, \bibinfo{person}{Helge Seetzen}, \bibinfo{person}{H.~David Young}, \bibinfo{person}{Wolfgang Heidrich}, \bibinfo{person}{Lorne Whitehead}, {and} \bibinfo{person}{Greg Ward}.} \bibinfo{year}{2007}\natexlab{}.
\newblock \showarticletitle{LDR2HDR: On-the-Fly Reverse Tone Mapping of Legacy Video and Photographs}.
\newblock \bibinfo{journal}{\emph{ACM Trans. Graph.}} \bibinfo{volume}{26}, \bibinfo{number}{3} (\bibinfo{year}{2007}), \bibinfo{pages}{39}.
\newblock
\showISSN{0730-0301}
\urldef\tempurl%
\url{https://doi.org/10.1145/1276377.1276426}
\showDOI{\tempurl}


\bibitem[Rombach et~al\mbox{.}(2022)]%
        {stableDiffusion}
\bibfield{author}{\bibinfo{person}{Robin Rombach}, \bibinfo{person}{Andreas Blattmann}, \bibinfo{person}{Dominik Lorenz}, \bibinfo{person}{Patrick Esser}, {and} \bibinfo{person}{Bj{\"o}rn Ommer}.} \bibinfo{year}{2022}\natexlab{}.
\newblock \showarticletitle{High-resolution image synthesis with latent diffusion models}. In \bibinfo{booktitle}{\emph{Proceedings of the IEEE/CVF conference on computer vision and pattern recognition}}. \bibinfo{pages}{10684--10695}.
\newblock


\bibitem[Ronneberger et~al\mbox{.}(2015)]%
        {Ronneberger+15}
\bibfield{author}{\bibinfo{person}{Olaf Ronneberger}, \bibinfo{person}{Philipp Fischer}, {and} \bibinfo{person}{Thomas Brox}.} \bibinfo{year}{2015}\natexlab{}.
\newblock \showarticletitle{U-Net: Convolutional Networks for Biomedical Image Segmentation}. In \bibinfo{booktitle}{\emph{Medical Image Computing and Computer-Assisted Intervention - {MICCAI} 2015 - 18th International Conference Munich, Germany, October 5 - 9, 2015, Proceedings, Part {III}}} \emph{(\bibinfo{series}{Lecture Notes in Computer Science}, Vol.~\bibinfo{volume}{9351})}, \bibfield{editor}{\bibinfo{person}{Nassir Navab}, \bibinfo{person}{Joachim Hornegger}, \bibinfo{person}{William M.~Wells III}, {and} \bibinfo{person}{Alejandro~F. Frangi}} (Eds.). \bibinfo{publisher}{Springer}, \bibinfo{pages}{234--241}.
\newblock
\urldef\tempurl%
\url{https://doi.org/10.1007/978-3-319-24574-4\_28}
\showDOI{\tempurl}


\bibitem[Saharia et~al\mbox{.}(2022)]%
        {Imagen}
\bibfield{author}{\bibinfo{person}{Chitwan Saharia}, \bibinfo{person}{William Chan}, \bibinfo{person}{Saurabh Saxena}, \bibinfo{person}{Lala Li}, \bibinfo{person}{Jay Whang}, \bibinfo{person}{Emily~L Denton}, \bibinfo{person}{Kamyar Ghasemipour}, \bibinfo{person}{Raphael Gontijo~Lopes}, \bibinfo{person}{Burcu Karagol~Ayan}, \bibinfo{person}{Tim Salimans}, {et~al\mbox{.}}} \bibinfo{year}{2022}\natexlab{}.
\newblock \showarticletitle{Photorealistic text-to-image diffusion models with deep language understanding}.
\newblock \bibinfo{journal}{\emph{Advances in Neural Information Processing Systems}}  \bibinfo{volume}{35} (\bibinfo{year}{2022}), \bibinfo{pages}{36479--36494}.
\newblock


\bibitem[Santos et~al\mbox{.}(2020)]%
        {Santos+20}
\bibfield{author}{\bibinfo{person}{Marcel~Santana Santos}, \bibinfo{person}{Tsang~Ing Ren}, {and} \bibinfo{person}{Nima~Khademi Kalantari}.} \bibinfo{year}{2020}\natexlab{}.
\newblock \showarticletitle{Single Image HDR Reconstruction Using a CNN with Masked Features and Perceptual Loss}.
\newblock \bibinfo{journal}{\emph{ACM Trans. Graph.}} \bibinfo{volume}{39}, \bibinfo{number}{4}, Article \bibinfo{articleno}{80} (\bibinfo{year}{2020}), \bibinfo{numpages}{10}~pages.
\newblock
\showISSN{0730-0301}
\urldef\tempurl%
\url{https://doi.org/10.1145/3386569.3392403}
\showDOI{\tempurl}


\bibitem[Seetzen et~al\mbox{.}(2004)]%
        {Seetzen+04}
\bibfield{author}{\bibinfo{person}{Helge Seetzen}, \bibinfo{person}{Wolfgang Heidrich}, \bibinfo{person}{Wolfgang Stuerzlinger}, \bibinfo{person}{Greg Ward}, \bibinfo{person}{Lorne Whitehead}, \bibinfo{person}{Matthew Trentacoste}, \bibinfo{person}{Abhijeet Ghosh}, {and} \bibinfo{person}{Andrejs Vorozcovs}.} \bibinfo{year}{2004}\natexlab{}.
\newblock \showarticletitle{High dynamic range display systems}.
\newblock \bibinfo{journal}{\emph{ACM Trans. Graph.}} \bibinfo{volume}{23}, \bibinfo{number}{3} (\bibinfo{date}{aug} \bibinfo{year}{2004}), \bibinfo{pages}{760–768}.
\newblock
\showISSN{0730-0301}
\urldef\tempurl%
\url{https://doi.org/10.1145/1015706.1015797}
\showDOI{\tempurl}


\bibitem[Venkatanath et~al\mbox{.}(2015)]%
        {Venkatanath+15}
\bibfield{author}{\bibinfo{person}{N. Venkatanath}, \bibinfo{person}{Praneeth D.}, \bibinfo{person}{Maruthi~Chandrasekhar Bh.}, \bibinfo{person}{Sumohana~S. Channappayya}, {and} \bibinfo{person}{Swarup~S. Medasani}.} \bibinfo{year}{2015}\natexlab{}.
\newblock \showarticletitle{Blind image quality evaluation using perception based features}. In \bibinfo{booktitle}{\emph{Twenty First National Conference on Communications, {NCC} 2015, Mumbai, India, February 27 - March 1, 2015}}. \bibinfo{publisher}{{IEEE}}, \bibinfo{pages}{1--6}.
\newblock
\urldef\tempurl%
\url{https://doi.org/10.1109/NCC.2015.7084843}
\showDOI{\tempurl}


\bibitem[Wang et~al\mbox{.}(2023)]%
        {Wang+23}
\bibfield{author}{\bibinfo{person}{Chao Wang}, \bibinfo{person}{Ana Serrano}, \bibinfo{person}{Xingang Pan}, \bibinfo{person}{Bin Chen}, \bibinfo{person}{Karol Myszkowski}, \bibinfo{person}{Hans-Peter Seidel}, \bibinfo{person}{Christian Theobalt}, {and} \bibinfo{person}{Thomas Leimk{\"u}hler}.} \bibinfo{year}{2023}\natexlab{}.
\newblock \showarticletitle{GlowGAN: Unsupervised Learning of HDR Images from LDR Images in the Wild}. In \bibinfo{booktitle}{\emph{Proceedings of the IEEE/CVF International Conference on Computer Vision}}. \bibinfo{pages}{10509--10519}.
\newblock


\bibitem[Wang et~al\mbox{.}(2007)]%
        {Wang+07}
\bibfield{author}{\bibinfo{person}{Lvdi Wang}, \bibinfo{person}{Li-Yi Wei}, \bibinfo{person}{Kun Zhou}, \bibinfo{person}{Baining Guo}, {and} \bibinfo{person}{Heung-Yeung Shum}.} \bibinfo{year}{2007}\natexlab{}.
\newblock \showarticletitle{High Dynamic Range Image Hallucination}. In \bibinfo{booktitle}{\emph{SIGGRAPH '07: ACM SIGGRAPH 2007 Sketches}} (San Diego, California). \bibinfo{publisher}{ACM}, \bibinfo{address}{New York, NY, USA}, \bibinfo{pages}{72}.
\newblock
\urldef\tempurl%
\url{https://doi.org/10.1145/1278780.1278867}
\showDOI{\tempurl}


\bibitem[Wu et~al\mbox{.}(2019)]%
        {fastfcn+19}
\bibfield{author}{\bibinfo{person}{Huikai Wu}, \bibinfo{person}{Junge Zhang}, \bibinfo{person}{Kaiqi Huang}, \bibinfo{person}{Kongming Liang}, {and} \bibinfo{person}{Yizhou Yu}.} \bibinfo{year}{2019}\natexlab{}.
\newblock \showarticletitle{{Fastfcn: Rethinking Dilated Convolution in the Backbone for Semantic Segmentation}}.
\newblock \bibinfo{journal}{\emph{arXiv preprint arXiv:1903.11816}} (\bibinfo{year}{2019}).
\newblock


\bibitem[Xie et~al\mbox{.}(2021)]%
        {xie+21}
\bibfield{author}{\bibinfo{person}{Enze Xie}, \bibinfo{person}{Wenhai Wang}, \bibinfo{person}{Zhiding Yu}, \bibinfo{person}{Anima Anandkumar}, \bibinfo{person}{Jose~M Alvarez}, {and} \bibinfo{person}{Ping Luo}.} \bibinfo{year}{2021}\natexlab{}.
\newblock \showarticletitle{{SegFormer: Simple and efficient design for semantic segmentation with transformers}}.
\newblock \bibinfo{journal}{\emph{Advances in Neural Information Processing Systems}}  \bibinfo{volume}{34} (\bibinfo{year}{2021}), \bibinfo{pages}{12077--12090}.
\newblock


\bibitem[Yu et~al\mbox{.}(2021)]%
        {Yu+21}
\bibfield{author}{\bibinfo{person}{Hanning Yu}, \bibinfo{person}{Wentao Liu}, \bibinfo{person}{Chengjiang Long}, \bibinfo{person}{Bo Dong}, \bibinfo{person}{Qin Zou}, {and} \bibinfo{person}{Chunxia Xiao}.} \bibinfo{year}{2021}\natexlab{}.
\newblock \showarticletitle{Luminance Attentive Networks for HDR Image and Panorama Reconstruction}.
\newblock \bibinfo{journal}{\emph{Computer Graphics Forum}} \bibinfo{volume}{40}, \bibinfo{number}{7} (\bibinfo{year}{2021}), \bibinfo{pages}{181--192}.
\newblock


\bibitem[Yu and Smith(2019)]%
        {yu2019inverserendernet}
\bibfield{author}{\bibinfo{person}{Ye Yu} {and} \bibinfo{person}{William~AP Smith}.} \bibinfo{year}{2019}\natexlab{}.
\newblock \showarticletitle{Inverserendernet: Learning single image inverse rendering}. In \bibinfo{booktitle}{\emph{Proceedings of the IEEE/CVF Conference on Computer Vision and Pattern Recognition}}. \bibinfo{pages}{3155--3164}.
\newblock


\bibitem[Yurtkulu et~al\mbox{.}(2019)]%
        {yurtkulu+19}
\bibfield{author}{\bibinfo{person}{Salih~Can Yurtkulu}, \bibinfo{person}{Yusuf~H{\"u}seyin {\c{S}}ahin}, {and} \bibinfo{person}{Gozde Unal}.} \bibinfo{year}{2019}\natexlab{}.
\newblock \showarticletitle{{Semantic Segmentation with Extended DeepLabv3 Architecture}}. In \bibinfo{booktitle}{\emph{2019 27th Signal Processing and Communications Applications Conference (SIU)}}. IEEE, \bibinfo{pages}{1--4}.
\newblock


\bibitem[Zhang et~al\mbox{.}(2023)]%
        {zhang2023adding}
\bibfield{author}{\bibinfo{person}{Lvmin Zhang}, \bibinfo{person}{Anyi Rao}, {and} \bibinfo{person}{Maneesh Agrawala}.} \bibinfo{year}{2023}\natexlab{}.
\newblock \showarticletitle{Adding conditional control to text-to-image diffusion models}. In \bibinfo{booktitle}{\emph{Proceedings of the IEEE/CVF International Conference on Computer Vision}}. \bibinfo{pages}{3836--3847}.
\newblock


\bibitem[Zhou et~al\mbox{.}(2017)]%
        {zhou+17}
\bibfield{author}{\bibinfo{person}{Bolei Zhou}, \bibinfo{person}{Hang Zhao}, \bibinfo{person}{Xavier Puig}, \bibinfo{person}{Sanja Fidler}, \bibinfo{person}{Adela Barriuso}, {and} \bibinfo{person}{Antonio Torralba}.} \bibinfo{year}{2017}\natexlab{}.
\newblock \showarticletitle{{Scene Parsing through ADE20k Dataset}}. In \bibinfo{booktitle}{\emph{Proceedings of the IEEE conference on computer vision and pattern recognition}}. \bibinfo{pages}{633--641}.
\newblock


\bibitem[Zou et~al\mbox{.}(2023)]%
        {Zou+23}
\bibfield{author}{\bibinfo{person}{Yunhao Zou}, \bibinfo{person}{Chenggang Yan}, {and} \bibinfo{person}{Ying Fu}.} \bibinfo{year}{2023}\natexlab{}.
\newblock \showarticletitle{RawHDR: High Dynamic Range Image Reconstruction from a Single Raw Image}. In \bibinfo{booktitle}{\emph{Proceedings of the IEEE/CVF International Conference on Computer Vision (ICCV)}}. \bibinfo{publisher}{IEEE}, \bibinfo{pages}{12334--12344}.
\newblock


\end{thebibliography}

\clearpage
\newpage

\noindent\begin{minipage}{\textwidth}
    \centering
    \includegraphics[width=1\linewidth]{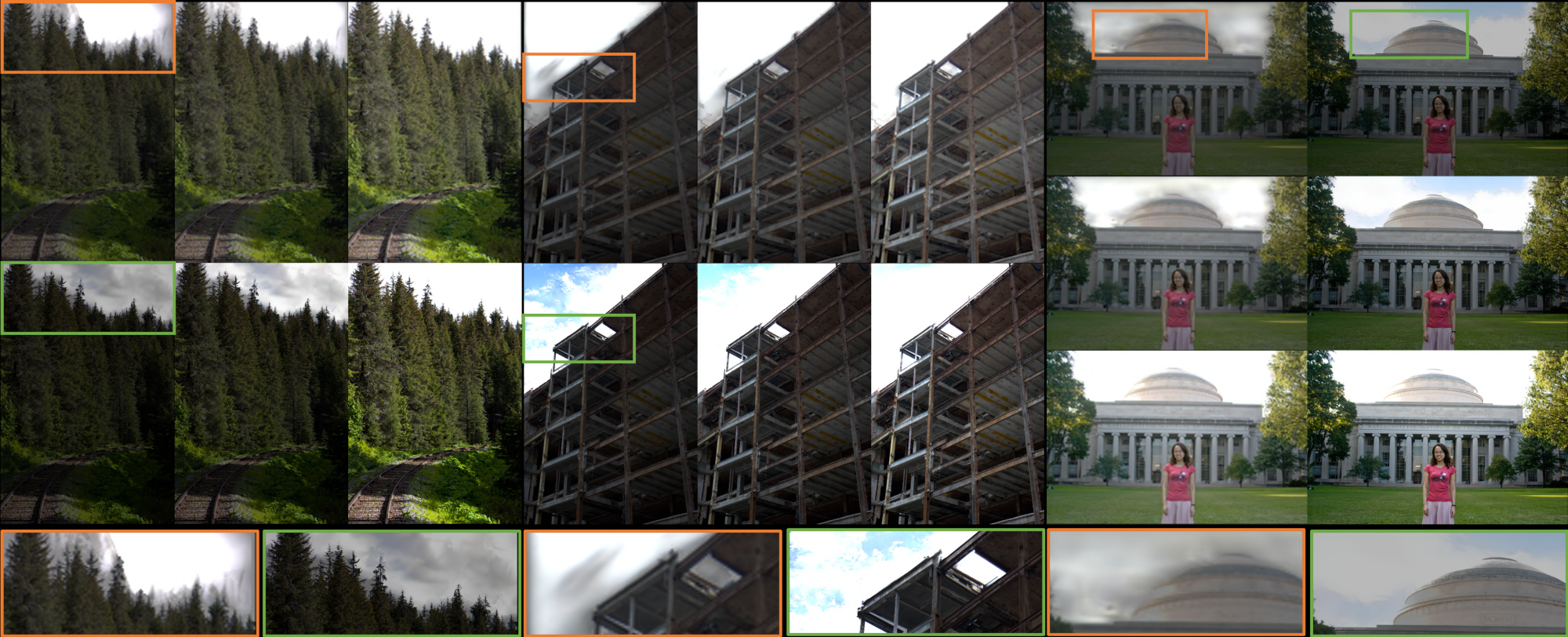}
    \captionof{figure}{GlowGAN Comparisons. GlowGAN is on the top in the landscape images, and the left in the portrait images. This shows multiple exposures of the same scene and the insets highlight the differences between the methods.}
    \label{fig:Glow}
% \end{figure*}
\end{minipage}

\newpage

\end{document}